 \documentclass[final,5p,times,twocolumn]{elsarticle}

\usepackage{amssymb}
\usepackage{amsthm}

 \theoremstyle{definition}

\usepackage[keeplastbox]{flushend}

\usepackage{natbib}        
\usepackage{framed}
\usepackage{subfig}
\usepackage{indentfirst}
\usepackage{amsmath}
\usepackage{cleveref}
\crefname{figure}{Fig.}{Fig.}
\Crefname{figure}{Figure}{Fig.}
\usepackage{bm}

\usepackage{pgfplots}
\pgfplotsset{compat=newest}
\usetikzlibrary{calc}
\usetikzlibrary{plotmarks}
\usetikzlibrary{arrows.meta}
\usepgfplotslibrary{patchplots}
\usepackage{grffile}

\usepackage{siunitx}
\definecolor{mycolor1}{rgb}{0.00000,0.44700,0.74100}%
\definecolor{mycolor2}{rgb}{0.85000,0.32500,0.09800}%
\definecolor{mycolor3}{rgb}{0.92900,0.69400,0.12500}%
\definecolor{mycolor4}{rgb}{0.49400,0.18400,0.55600}%
\definecolor{mycolor5}{rgb}{0.46600,0.67400,0.18800}%
\definecolor{mycolor6}{rgb}{0.30100,0.74500,0.93300}%

\usepackage{enumitem}
\newenvironment{descriere}%
  {\begin{description}%
    \setlength{\itemsep}{0pt}%
    \setlength{\parskip}{1pt}%
    }
  {\end{description}}

\journal{*}

\begin{document}

\begin{frontmatter}



\title{SERoCS: Safe and Efficient Robot Collaborative Systems \\for Next Generation Intelligent Industrial Co-Robots\tnoteref{footnoteinfo}} 

\tnotetext[footnoteinfo]{The work is supported by National Science Foundation Award \#1734109.}

\author[First]{Changliu Liu\fnref{equalcontrobution}}
\ead{changliuliu@berkeley.edu}
\author[First]{Te Tang\fnref{equalcontrobution}}
\ead{tetang@berkeley.edu}
\author[First]{Hsien-Chung Lin\fnref{equalcontrobution}}
\ead{hclin@berkeley.edu}
\author[First]{Yujiao Cheng\fnref{equalcontrobution}}
\ead{yujiaocheng@berkeley.edu}
\author[First]{Masayoshi Tomizuka}
\ead{tomizuka@berkeley.edu}

\address[First]{Department of Mechanical Engineering, University of California at Berkeley, Berkeley CA 94720}
\fntext[equalcontrobution]{The authors contributed equally to this paper.}

\begin{abstract}
Human-robot collaborations have been recognized as an essential component for future factories. It remains challenging to properly design the behavior of those co-robots. Those robots operate in dynamic uncertain environment with limited computation capacity. The design objective is to maximize their task efficiency while guaranteeing safety. This paper discusses a set of design principles of a safe and efficient robot collaboration system (SERoCS) for the next generation co-robots, which consists of robust cognition algorithms for environment monitoring, efficient task planning algorithms for reference generations, and safe motion planning and control algorithms for safe human-robot interactions.  The proposed SERoCS will address the design challenges and significantly expand the skill sets of the co-robots to allow them to work safely and efficiently with their human counterparts. The development of SERoCS will create a significant advancement toward adoption of co-robots in various industries. The experiments validate the effectiveness of SERoCS.
\end{abstract}


\end{frontmatter}



\section{Introduction}

\subsection{Human-Robot Collaboration in Manufacturing\label{sec: motivation}}
\begin{figure}
  \begin{center}
    \includegraphics[width=7.9cm]{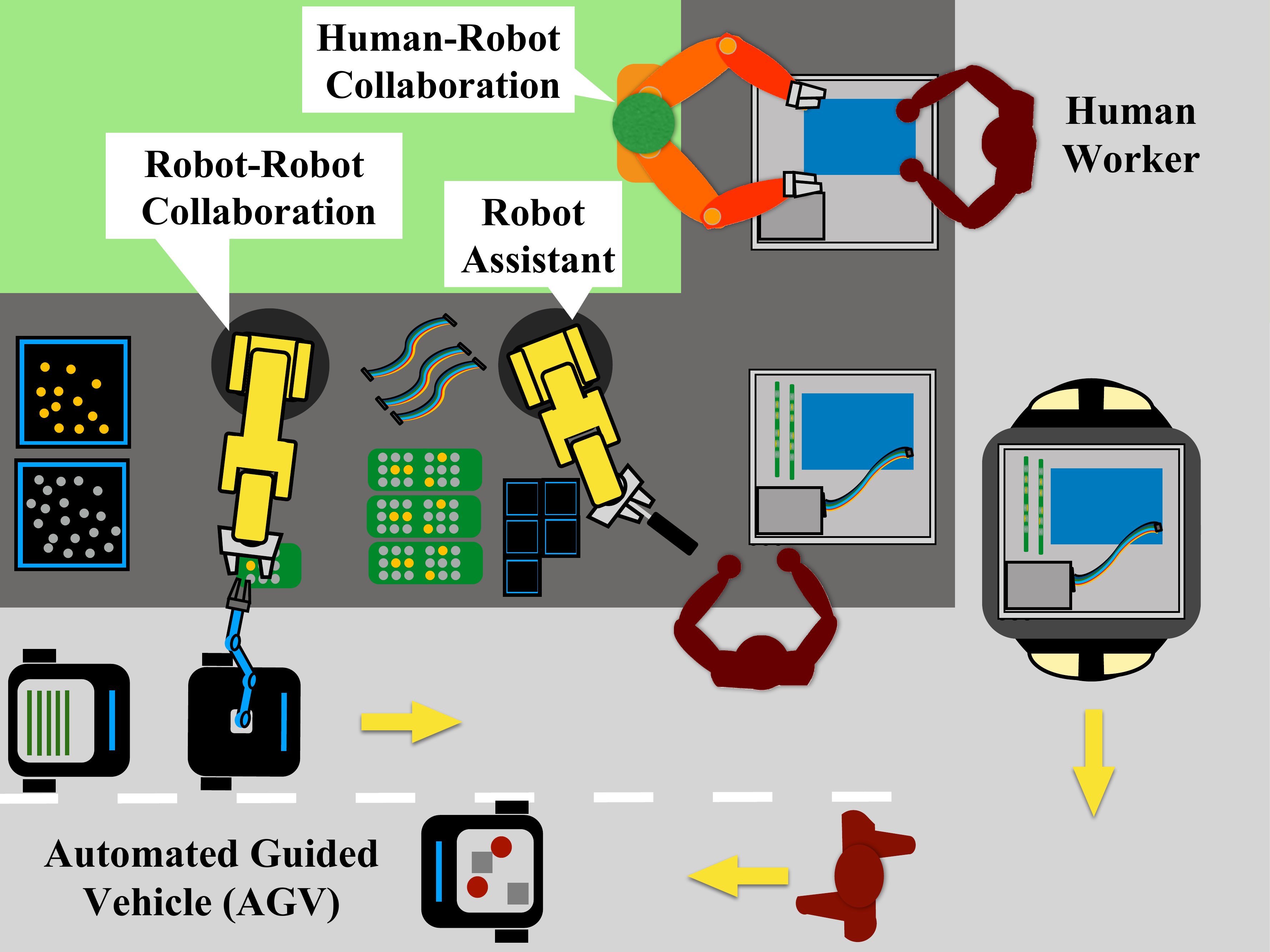}
  \end{center}
  \caption{Flexible production lines in the future, which involve human-robot co-operation and co-inhabitance.\label{fig: future factory}}
\end{figure}

In modern factories, human workers and robots are two major workforces.  For safety concerns, the two are normally separated with robots confined in metal cages, which limits the productivity as well as the flexibility of production lines. In recent years, attention has been directed to remove the cages so that human workers and robots may collaborate to create a human-robot co-existing factory \cite{charalambous2013human, koeppe2005robot} as illustrated in \cref{fig: future factory}.

The potential benefits of uncaged robots are huge and extensive. For example, they may be placed in human-robot teams in flexible production lines \cite{kruger2009cooperation,schmidt2008kobot}.  It is observed that the emphasis in manufacturing will shift from mass production to mass customization, as consumers' interest in personalized products keeps increasing \cite{pine1999mass}. In response to such shifts, many research and development efforts have been directed to flexible automation \cite{hutchinson1982economic,jovane2003present}. However, it is difficult to make the current production lines with robots truly flexible, due to the rigidity of the current generation of industrial robots. On the other hand, including human workers in the human-robot teams will bring flexibility, intelligence and versatility to automation.

Automotive manufacturers such as Volkswagen \cite{Volkswagen-Cooperative} and BMW \cite{Econ-Our-Friends-Electric} introduced human-robot cooperation in final assembly lines in 2013. In BMW's factory in Spartanburg, South Carolina, robot arms cooperate with human workers to insulate and water-seal automobile doors in final door assembly.  The robot spreads out and glues down material that is held in place by the human worker's more agile hands. Before the introduction of these robots, workers had to be rotated off this uncomfortable and physically straining task after one or two hours to prevent elbow strain.
In addition to cooperative robot arms, other types of cooperation are attractive \cite{rey2013cooperation}. For example, cooperation among automated guided vehicles (AGVs) and human workers \cite{ulusoy1997genetic} in factory logistics. 
Such cooperation will be key to making the factories of the future productive and competitive, which will revitalize the production system and enhance the world's economy.  

\subsection{Co-Robot: State of the Art}
There are several commercialized safe cooperative robots or co-robots on the market, such as UR5 from Universal Robots (Denmark) \cite{UR5}, Collaborative Robots \textit{CR} family from FANUC (Japan) \cite{fanucCorobot, fanucGreenRobot}, Baxter from Rethink Robotics (US) \cite{rethinkrobotics}, NextAge from Kawada (Japan) \cite{kawada} and WorkerBot from Pi4\_Robotics GmbH (Germany) \cite{workerbot}. 
Most of these robots ensure safety using two protection mechanisms: 1) motion shutdown when a potential collision is predicted; and 2) impact reduction if a collision takes place. These mechanisms are very primitive. The current design approach for co-robots 1) sacrifices efficiency for safety and 2) require extensive programming in order to adapt these robots to different tasks, which is not desirable from the viewpoint of productivity.  

In literature, several successful implementations of non-industrial co-robots have been reported, e.g., home assist robots \cite{yamazaki2012home}, museum tour guide robots \cite{burgard1999experiences,thrun1999minerva} and nursing robots \cite{pineau2003towards}. To enable interactive behaviors, complex software architectures are developed to equip the robots with various cognition, learning, and motion planning abilities. 
However, these robots are mostly of human-size or smaller with slow motion, which may not be cost-efficient for industrial applications as discussed in \cref{sec: motivation}. 
To fully realize a human-robot co-existing factory, the software design methodology for high performance industrial co-robots, especially those that are large in size, with multiple links and complicated dynamics, needs to be explored. 

On the other hand, safety of industrial robots during physical human-robot interactions (pHRI) also attracts attention from standardization bodies \cite{harper2010towards}, research communities \cite{PHRIENDS,ROSETTA,SAPHARI}, as well as major robot manufacturers 
\cite{anandan2014major,Tadele2014safety}. 
However, existing researches mainly focus on intrinsic safety, i.e., safety in mechanical design \cite{hirzinger2001new}, actuation \cite{zinn2004new,jafari2010novel,english1999implementation} and low level motion control \cite{albu2007unified,luo2011adaptive,hogan1984impedance}. 
Behavioral safety during collaborations and interactions, which depends on a clear understanding of the environment and the ability to generate responsive motions, still needs to be explored.

In addition to safety, the task efficiency of industrial robots is also important. As production lines become more flexible, robots need to be able to adapt to various tasks. The emphasis of motion planning is shifting from rigid methods such as hard coding to flexible skill-based methods \cite{tan2009human}, e.g. robots should understand certain generalized skills to perform various tasks and be able to generate motion in different environments.


\subsection{The Design Challenges}
\begin{figure}[t]
\begin{center}
\includegraphics[width=7cm]{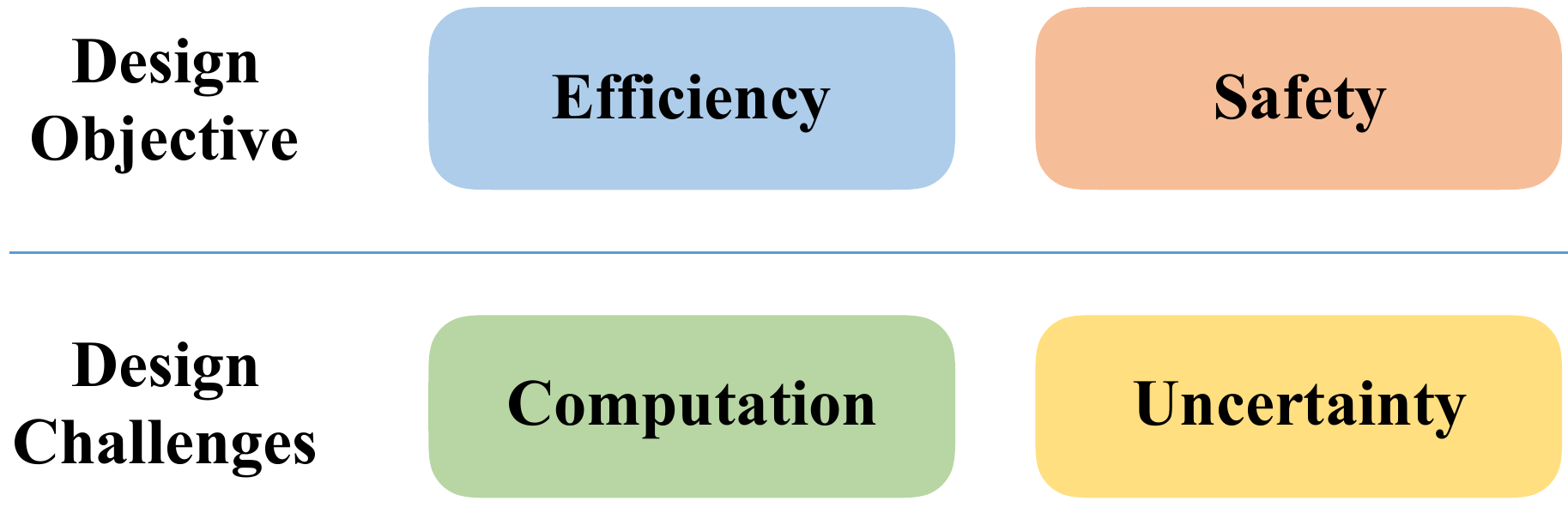}
\caption{The challenges in designing the behavior of the co-robots.}
\label{fig: challenges}
\end{center}
\end{figure}

Moving robots from cages poses new challenges in robotics as robots and human workers now directly interact with each other. A prerequisite for successful collaboration between humans and robots is to guarantee the safety of the humans.  At the same time, it is important to ensure that robots collaborate with humans with the best performance possible, i.e., the robot motion should be both safe and efficient. In well-defined and deterministic environments, safety and efficiency can be achieved by the state of art. However, interactions with human workers bring a lot of uncertainties to the system. Moreover, the onboard computation power is limited to allow the robot to account for all possible scenarios during real time interactions. These represent major challenges faced by the co-robots as summarized in \cref{fig: challenges}. This paper discusses methods to design the behavior of those co-robots in dynamic uncertain environment with limited computation capacity in order to maximize task efficiency while guaranteeing safety. 

\subsection{Contributions of the paper}
This paper discusses a set of design principles of a safe and efficient robot collaboration system (SERoCS) for the next generation co-robots, which consists of robust cognition algorithms for environment monitoring, efficient task planning algorithms for safe human-robot collaboration, and safe motion planning and control algorithms for safe human-robot interactions (HRI).  The proposed SERoCS will address the design challenges and significantly expand the skill sets of the co-robots to allow them to work safely and efficiently with their human counterparts. The development of SERoCS will create a significant advancement toward adoption of co-robots in various industries.

The remainder of this paper is organized as follows. \Cref{sec: overview} provides an overview of SERoCS. \Cref{sec: T1,sec: T2,sec: T3} discuss each module in details. \Cref{sec: T4} provides both theoretical and experimental evaluations of SERoCS. \Cref{sec: conclusion} concludes the paper.

\subsection{Nomenclature}
\begin{descriere}
\item[$x_R$] Robot state ($\mathbf{x}_R$ for corresponding trajectory)
\item[$u_R$] Robot control input ($\mathbf{u}_R$ for corresponding trajectory)
\item[$x_H$] Human state ($\mathbf{x}_H$ for corresponding trajectory)
\item[$x_e$] Environment state ($\mathbf{x}_e$ for corresponding trajectory)
\item[$\pi_R$] Sensory information
\item[$t$] Time (\si{\second})
\item[$k$] Time step
\item[$T$] Planning horizon (\si{\second})
\item[$t_s$] Sampling time (\si{\second})
\item[$N$] Planning steps $=T/t_s+1$
\item[$\Gamma$] Constraints on the robot trajectory
\item[$\Omega$] Constraints on the robot input
\item[$J$] Cost function
\item[$X_S$] The safe set
\item[$p$] Human plan
\item[$\mathbf{P}^s$] Source point cloud set
\item[$\mathbf{P}^t$] Target point cloud set
\item[$\mathbf{g}$] Grasp pose
\item[$\bm{t}$] The center of grasp
\item[$\bm{R}$] The orientation of grasp
\item[$\mathcal{F}$] The convex feasible set
\item[$\phi$] The safety index
\item[$R_S$] Safety constraint on the robot state space
\item[$U_S$] Safety constraint on the robot control space
\end{descriere}

\section{Overview of SERoCS\label{sec: overview}}
This section provides an overview of SERoCS. The behavior design problem during human-robot collaboration is described mathematically, followed by the introduction of the SERoCS architecture that solves the problem. An example is provided to illustrate the desired performance.

\subsection{The Mathematical Problem}
For simplicity, this paper focuses on the scenario with one robot and one human. The methodology extends to scenarios with multiple robots and multiple humans. 
Denote the robot trajectory from current time $t$ to time $t+T$ as $\mathbf{x}_R:=x_R(t:t+T)$. The planning horizon $T$ can either be chosen as a fixed number or as a decision variable that should be optimized up to the accomplishment of the task. Similarly, the trajectories of the human and the environment from $t$ to $t+T$ are $\mathbf{x}_H$ and $\mathbf{x}_e$. The sensory information $\pi_R$ contains information up to current time $t$. To obtain desired motion trajectory for the robot during human-robot collaborations, the following optimization problem is considered,
\begin{subequations}\label{eq: chap2 design of knowledge}
\begin{align}
\min_{\mathbf{x}_R} ~&  E\left[J(\mathbf{x}_R,\mathbf{x}_H,\mathbf{x}_e)\mid\pi_R\right]\label{eq: chap2 behavior design cost},\\
s.t. ~& \mathbf{x}_R \in \Gamma\label{eq: chap2 behavior design self constraint},\\
& P\left(\left\{(x_R(t), x_H(t), x_e(t))\in {X}_S\right\}\mid\pi_R\right)=1,\forall t\label{eq: chap2 behavior design interaction constraint},
\end{align}
\end{subequations}
where \eqref{eq: chap2 behavior design cost} is the expected cost for task performance. The cost function $J$ evaluates the trajectories of the robot, the human, and the environment. 
Equation \eqref{eq: chap2 behavior design self constraint} represents the feasibility and dynamic constraint on the robot trajectory. The planned trajectory should be executable by the robot hardware, considering the robot dynamics 
\begin{equation}
\dot x_R(\tau) = f(x_R(\tau))+h(x_R(\tau))u_R(\tau), \label{eq: robot dynamics}
\end{equation}
which is assumed to be affine in the control input. The functions $f$ and $h$ are assumed to be smooth. 
Then $\Gamma=\{\mathbf{x}_R:\exists \ u_R(\tau) \in \Omega \text{, s.t. }\dot x_R(\tau) = f(x_R(\tau))+h(x_R(\tau))u_R(\tau), \ \forall \ \tau\in[t,t+T]\}$. The set $\Omega$ is assumed to be convex. 
Equation \eqref{eq: chap2 behavior design interaction constraint} is a chance constraint for safety. The safe set ${X}_S$ is a subset of the system's state space. The system state should belong to the safe set with absolute certainty.  

There are two steps in generating a desired robot motion trajectory: 
\begin{itemize}
\item Formulation of the problem \eqref{eq: chap2 design of knowledge};
\item Solving the problem \eqref{eq: chap2 design of knowledge} for $\mathbf{x}_R$.
\end{itemize}
During problem formulation, the trajectories $\mathbf{x}_H$ and $\mathbf{x}_e$ need to be predicted, which will be handled by an environment monitoring module. The cost function $J$ needs to be constructed given current task progress, which will be handled by a task planning module. Finally, the optimal trajectory will be obtained by solving problem \eqref{eq: chap2 design of knowledge} in a motion planning module. The three modules are the main components in SERoCS, which will be discussed in detail below.

\subsection{The Architecture of SERoCS}
\begin{figure}
  \begin{center}
    \includegraphics[width=8.5cm]{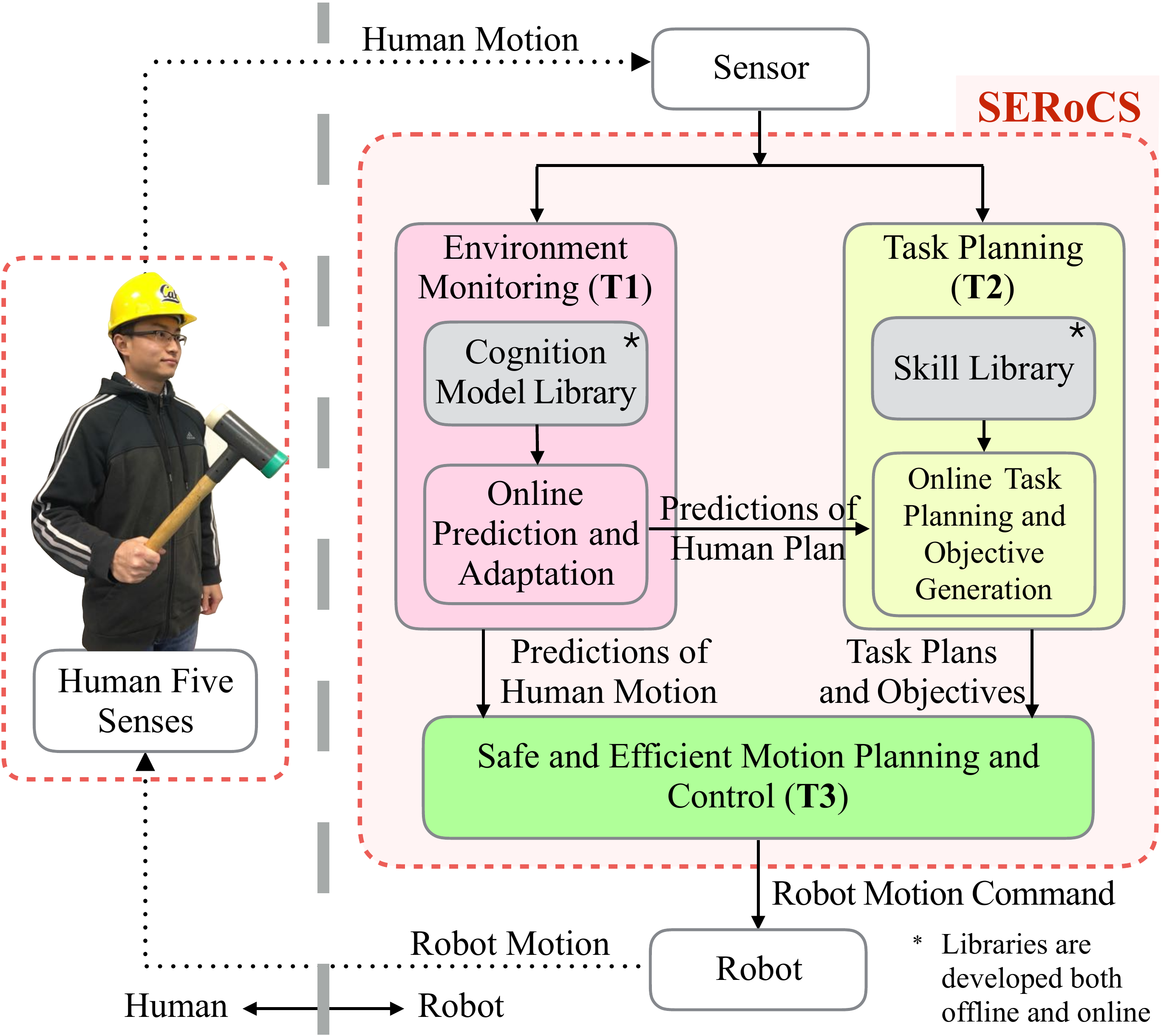}
  \end{center}
  \caption{Safe and efficient robot collaboration system (SERoCS). 
  \label{fig: SERoCS}}
\end{figure}

The architecture of the SERoCS is shown in \cref{fig: SERoCS}, which depicts the tasks in a one human worker and one robot situation, but can also extend to multi-agent situations. 

\subsubsection*{T1. Environment Monitoring with Human Motion Prediction}
The input of this module is the sensory information $\pi_R$. The output consists of the current states $x_H(t)$ and $x_e(t)$ as well as the predicted trajectories $\hat{\mathbf{x}}_H$ and $\hat{\mathbf{x}}_e$. To generate high fidelity prediction, a learning-based method is used. Offline deep learning is employed to construct cognition models for human plan recognition and human motion prediction. Online learning is designed to adapt the models to time-varying behaviors and quantify the uncertainties in the prediction.

\subsubsection*{T2. Task Planning with Skill Library Learned from Human Demonstration} 
The input of this module is the sensory information $\pi_R$ and the predicted trajectories $\hat{\mathbf{x}}_H$ and $\hat{\mathbf{x}}_e$. The output is parameters in the cost function $J$, especially the target pose or trajectory reference that a robot show arrive at or follow. In order to adapt to various tasks and environment, the robot learns offline to perform different tasks from human demonstration and record the knowledge in their motion skill library.
During online execution, the robot adapts to different environments by generating corresponding objectives using the motion skill library.

\begin{figure*}
\begin{center}
\subfloat[Step 1: Learning the human behavior and constructing a cognition model in order to predict the human motion online (T1). (i) The human assembles the two workpieces together. (ii) The human picks the tool. (ii) The human uses the tool to fasten the assembly.]{\includegraphics[width=16cm]{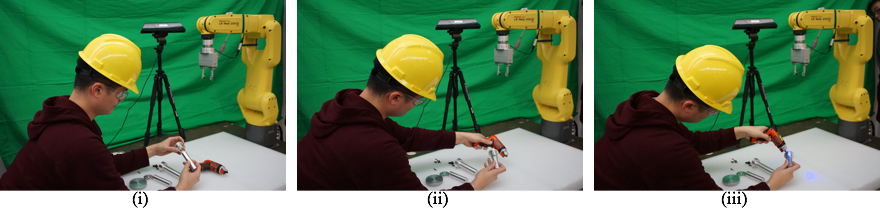}}\\
\subfloat[Step 2: Learning to grasp the tool by lead-through-teaching (T2). (i) The configuration before teaching. (ii) Human drags the robot to the desired grasping point. (iii) Human guides the robot to grasp and lift the tool.]{\includegraphics[width=16cm]{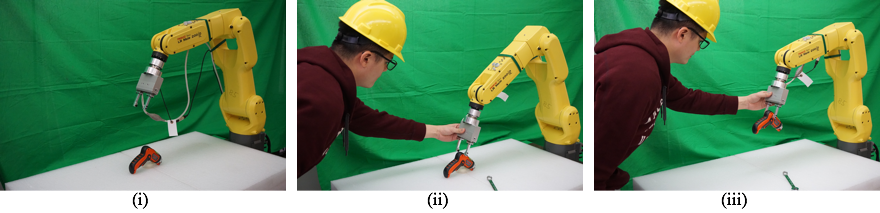}}\\
\subfloat[Step 3: Online human-robot collaboration enabled by human motion prediction (T1) using the cognition model constructed in step 1, task planning (T2) using the learned skill in step 2, and online motion planning (T3). (i) The human assembles the two workpieces together and the robot stays away from the human. (ii) The human finishes the assembly and the robot recognizes that the human needs the tool. (iii) The robot moves toward the tool. (iv) The robot picks up the tool. (v) The robot passes the tool to the human. (vi) The human fastens the assembly using the tool and the robot steps back.]{\includegraphics[width=16cm]{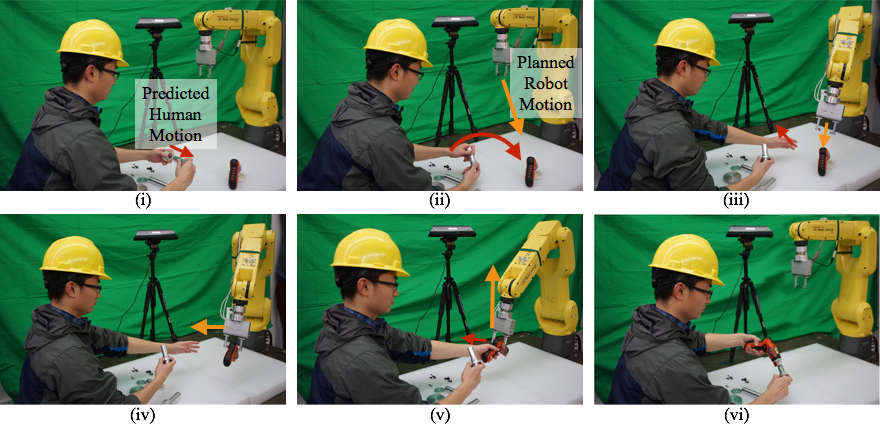}}
\caption{Illustration of the expected performance of the SERoCS in human-robot collaboration. The robot helps the human performing an assembly task by passing a tool. Human subjects and environment configurations vary in different steps.}
\label{fig: performance}
\end{center}
\end{figure*}

\subsubsection*{T3. Safe and Efficient Motion Planning and Control in Real Time} 
The input of this module is the optimization problem \eqref{eq: chap2 design of knowledge}, where the objective function is given by task planning in T2 and the safety constraint depends on the predicted trajectories in T1. The output is the desired motion trajectory $\mathbf{x}_R$. 
To ensure real time computation of a feasible and safe trajectory, a parallel planning and control architecture is developed, which consists of a long term efficiency-oriented planner and a short term safety-oriented controller. Real time algorithms are developed to solve the problems efficiently and make the SERoCS scalable.

\subsection{Example}
The expected performance of SERoCS is illustrated through an example of human-robot collaborative assembly in \cref{fig: performance}. There are three steps. In the first step, the robot learns the human behavior (in the example, the procedure for assembling the workpieces). In the second step, the robot learns to grasp the tool by human lead-through teaching. In the third step, the robot helps the human in finishing the assembly task which it learned in the first step by passing the tool to the human. 
The first two steps are called offline learning, while the last step is called online execution. The learned human behavior in the first step is recorded in the cognition model library, while the learned skill in the second step is recorded in the skill library. During online execution, the motion planning problem is formulated according to the outputs of the two libraries. The motion trajectory is then computed in real time.

\section{T1: Environment Monitoring\label{sec: T1}}
Environment monitoring aims to detect both workpieces (as static objects) and humans (as moving objects). As factory environment is highly structured, detailed CAD models of the static objects are usually available, which simplifies the recognition and detection of the workpieces. It is more challenging to monitor moving objects, i.e., human. This section discusses methods to track and predict the human plan $p$, and the human trajectory $\mathbf{x}_H$. 

The human plan $p$ corresponds to different ways to complete a task. Assuming there are $K$ different plans, then $p\in\{1,2,...,K\}$. The robot needs to know what plan the human is executing for smooth collaboration. For example, in \cref{fig: performance}, once the robot realizes that the human's plan is to assemble the two workpieces, it then passes the right tool to the human to let him fasten the assembly. By predicting the human plan, the robot can make corresponding long-term plans in advance, which improves task efficiency. 
Human trajectory consists of a sequence of the human's joints positions. Prediction of the human trajectory helps the robot determine the safety constraint \eqref{eq: chap2 behavior design interaction constraint}. Accurate prediction improves safety as well as task efficiency. In this paper, we use Kinect to detect human's joint positions. 

\begin{figure}[tb]
\begin{center}
\includegraphics[width=1\linewidth]{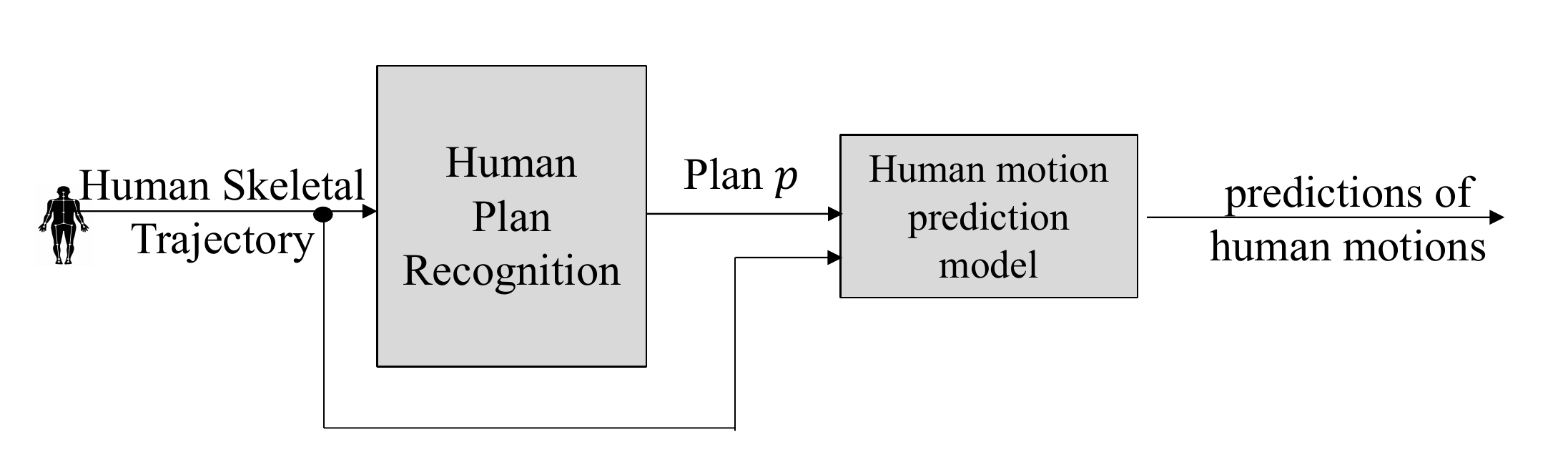}
\end{center}
\caption{Environment monitoring diagram. }
\label{fig: diagramT1}
\end{figure}

To predict the human's plan and trajectory, learning-based methods are used. A cognition model library for human plan recognition and human motion prediction is built offline by learning different human behaviors. Online algorithm is developed to make predictions of time-varying human behaviors using the trained models in the library. The process is shown in \cref{fig: diagramT1}. The human plan recognition model takes the previous human trajectory and predicts the plan $p$. The human motion prediction model takes the previous human trajectory and the predicted plan to predict the future human trajectory. 
 


\subsection{Trajectory Based Plan Recognition}
A plan is a sequence of actions, e.g.,  picking, screwing the hinge, etc. There may be many plans for a human to complete a task. For instance, if the human needs to go to two places, A and B, he or she can either go to place A first, or go to place B first, which correspond to two different plans. Plan recognition is to predict which plan the human is conducting. It is better to have accurate recognition as early in time as possible.  

The human trajectory provides information on the plan that he or she is conducting. In a highly structured factory environment, different plans lead to different motions, especially when the task is location-sensitive. In this sense, we can recognize plans robustly and accurately using only trajectories as inputs.

\begin{figure}[t]
\centering
\subfloat[The trajectory.]{
%
%
\definecolor{mycolor1}{rgb}{0.00000,0.44700,0.74100}%
\begin{tikzpicture}

\begin{axis}[%
width=1.6in,
height=1.3in,
at = {(0cm, 0cm)},
scale only axis,
xmin=0,
xmax=0.6,
tick align=outside,
xlabel={$x$ [\si{\meter}]},
ymin=0,
ymax=0.6,
ylabel={$y$ [\si{\meter}]},
zmin=0.16,
zmax=0.3,
zlabel={$z$ [\si{\meter}]},
view={-66}{16},
font = \footnotesize,
axis background/.style={fill=white}
]
\addplot3 [color=mycolor1, mark=asterisk, mark options={solid, mycolor1}]
 table[row sep=crcr] {%
0.0330431460979933	0.0229107892247837	0.182040786274848\\
0.0300428482622516	0.0209210079473722	0.181269806123256\\
0.0277084866028828	0.0192275857180553	0.180625393554566\\
0.027895200404857	0.0190749824162615	0.180718655100845\\
0.0260221717461278	0.0176865820350056	0.183122753910205\\
0.0310964441188445	0.0180408949302622	0.190792152054536\\
0.0579388380304499	0.0290486796314302	0.2061265928322\\
0.0786901132219744	0.038722883864093	0.216677627863274\\
0.0843761610092865	0.0410191068935235	0.22047311671525\\
0.10045362505978	0.0491740394013751	0.228538189768821\\
0.116650134784655	0.0570843712240711	0.236752845992039\\
0.12483855603951	0.0606710757367668	0.242237398543328\\
0.133376950235068	0.0648351635363502	0.248514357111573\\
0.17354555409587	0.0861154162117889	0.252638039721423\\
0.194459554820057	0.0993339457976327	0.267597737386549\\
0.180972090898252	0.0943535359437605	0.281963468281165\\
0.214511743724397	0.114088867896579	0.272006357570236\\
0.256503502566221	0.140053073891006	0.257471361580639\\
0.236196823254372	0.134024463152852	0.269715322824547\\
0.2491370773264	0.145504080472167	0.267480731366101\\
0.296313200366213	0.170713218556216	0.252400288439094\\
0.314899219034582	0.178256907847066	0.246643260524828\\
0.332641213585175	0.191129890192366	0.238596081151908\\
0.341366825245034	0.20193841534537	0.241469170733456\\
0.360189924616853	0.224000600721597	0.228093963629923\\
0.374640632043	0.240771008290077	0.216752729729854\\
0.366269661251683	0.239188300686043	0.238856279030183\\
0.375628275006473	0.254180217983237	0.237442015682188\\
0.39505685337078	0.270594676732676	0.217204386738832\\
0.398424554578187	0.270924353968714	0.214626090821106\\
0.401920243619319	0.272829623108945	0.211320957382135\\
0.406119800551123	0.278992902614861	0.208402269855688\\
0.409693970679517	0.287921620335723	0.204435860733347\\
0.412916352081846	0.298113168986126	0.201953705370393\\
0.418948998229427	0.306647744791856	0.201904222996929\\
0.424699246206916	0.313787693755106	0.200718653237709\\
0.42744615944923	0.323462009783	0.199978337502261\\
0.430641550636782	0.331019608443406	0.199495486606549\\
0.424083309612754	0.348407739986109	0.208526037996003\\
0.414961980737084	0.363802782887704	0.21989110648292\\
0.424704027096494	0.372385187411888	0.210583448885255\\
0.425907065859	0.38234423163541	0.20744634036882\\
0.417517599581507	0.388655884557872	0.209924654760731\\
0.416801989582141	0.397094846114866	0.205010707430437\\
0.413671760301886	0.405254653767591	0.20002394140329\\
0.397055980687735	0.414198738607481	0.200490810088401\\
0.386736347039229	0.423747010238086	0.200918400423853\\
0.374287276115442	0.432837184725137	0.194062016991616\\
0.358765902439245	0.436923527295539	0.192669617901586\\
0.358554462787394	0.437894481384895	0.193550933399016\\
0.355975959021945	0.440909632676381	0.19004154678703\\
0.356658027250316	0.445126866332466	0.186527940984861\\
0.358584046517161	0.449092328014982	0.18023570372995\\
0.360550191914466	0.454341307949396	0.180951154068025\\
0.354374811428754	0.461385509273486	0.192510773926166\\
0.342823680034084	0.466355471419899	0.195055570286562\\
0.337513851370628	0.469255172866512	0.193448174293495\\
0.33370653010338	0.471435024702117	0.193484227691765\\
0.331311888130394	0.472927061766092	0.191555810708508\\
0.33007456959725	0.475934767964161	0.19042695568383\\
0.323697033917521	0.482525832975408	0.191881858282039\\
0.340655245680415	0.501124169010698	0.184436192232707\\
0.361577017101313	0.511167226768407	0.172470114483446\\
0.359334389145678	0.507614970797643	0.169680817625434\\
0.358649160138666	0.507998005658268	0.169745730696445\\
0.348440041975917	0.503623576903354	0.16871450229029\\
0.337494180206183	0.499760810001027	0.167299011374942\\
0.335333847439411	0.500094898595937	0.166801362114738\\
0.332382004180754	0.499587345461588	0.166731704865909\\
0.330216069185548	0.499717652913217	0.166775653168142\\
0.328998854989267	0.50075095699337	0.167293714196042\\
0.327920159153053	0.501528661652136	0.167339362488139\\
0.327687523079997	0.502517301269325	0.165520877795909\\
0.327328152168444	0.502976518498314	0.163994691446061\\
0.327181266576785	0.500550666571993	0.164480225696603\\
0.327681330343417	0.500574308456483	0.166448099060425\\
0.327190328001364	0.502335878202692	0.167479372071435\\
0.323740238504965	0.500703329786363	0.16619893413817\\
0.320625806836161	0.497520924574211	0.165252906673205\\
0.319309295811887	0.49501277334084	0.165834615905031\\
0.31391460439136	0.490012917549056	0.168896172565139\\
0.310436940581552	0.486261159879486	0.173966595183075\\
0.31208980911643	0.491958578291585	0.174742203795629\\
0.312882166269725	0.498556512744015	0.173445338334591\\
0.313243178251076	0.500641773133017	0.174295454490961\\
0.311525984493054	0.497531214969268	0.172401251396519\\
0.309747184519444	0.494187409533074	0.170414125380447\\
0.308823969826144	0.504352343722708	0.16563014085864\\
0.310924408641423	0.505386537453576	0.170282820082045\\
0.311442547525279	0.494566909381891	0.178205958762649\\
0.306078586407536	0.491918214880471	0.177915305767822\\
0.305328553660161	0.492514561341021	0.18294484632419\\
0.306311093654275	0.493114698091515	0.18598204232277\\
0.306525744508689	0.492133657432043	0.185341720821257\\
0.306308925866336	0.491341377297383	0.18452840901219\\
0.303452595103514	0.491979865423209	0.180461180303893\\
0.302389049521444	0.49326635225861	0.176588409966632\\
0.302596074326847	0.493592553280771	0.175862480391244\\
};
 \end{axis}
\end{tikzpicture}%
}\\
\subfloat[Three view of the trajectory.]{
\input{src/traj_3.tex}}\\
\subfloat[The RGB image by combining the three view images.]{\makebox{\hspace{1.5cm}
%
%
\begin{tikzpicture}

\begin{axis}[%
width=1.6in,
height=1.3in,
at = {(1 cm, 1cm)},
scale only axis,
axis on top,
xmin=0.5,
xmax=224.5,
tick align=outside,
y dir=reverse,
ymin=0.5,
ymax=224.5,
axis line style={draw=none},
ticks=none
]
\addplot [forget plot] graphics [xmin=0.5, xmax=224.5, ymin=0.5, ymax=224.5] {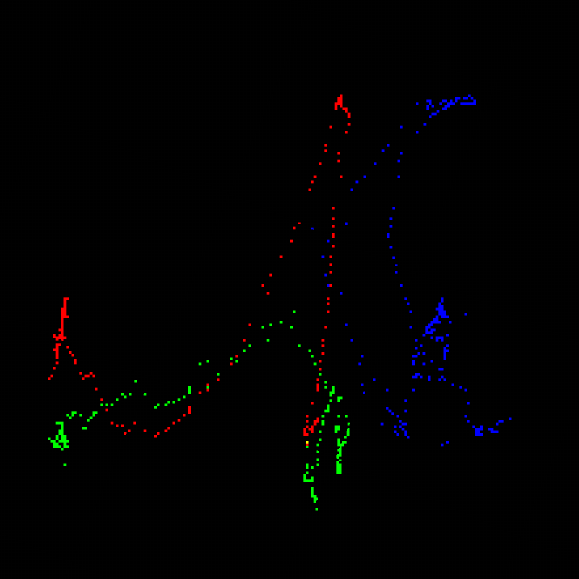};
\end{axis}
\end{tikzpicture}%
\hspace{1.5cm}}}
\caption{Transforming a trajectory into an image.\label{fig: inference2}}
\end{figure}

The state-of-the-art deep learning method provides highly accurate image classification. Taking the advantage of the deep learning method, we transform the 3D trajectories into colored images, then train a convolutional neural network model for plan recognition.
The way to transform the trajectories is shown in \cref{fig: inference2}. The trajectories of human for each time step are projected to the XY plane, the YZ plane, and the ZX plane. These three images are then put into RGB channels respectively to get the training images. The training images together with manually assigned labels are then used to train an Alexnet \cite{krizhevsky2012imagenet}. 

The trained model will be used for online prediction of the human plan $p$. The identified task plan will be sent to T2 for robot task planning, and be used in human motion prediction.

\subsection{Motion Prediction}

\subsubsection{State Transition Model}
Sample the human trajectory by rate $t_s$. At current time step $k$, $\mathbf{x}_H^*(k)$ denotes human's past trajectory at time steps $k, k-1, \ldots, k-N+1$. $\mathbf{x}_H(k+1)$ denotes human's future trajectory at time steps $k+1, k+2, \ldots, k+N$. Note that $\mathbf{x}_H(k) = \mathbf{x}_H^*(k+N-1)$. 
The dynamics of human motion are described by the following equation
\begin{align}
{\mathbf{x}_H(k+1)}  = f^*(\mathbf{x}_H^*(k), p) + w_k,\label{eq: T1 transition model}
\end{align}
where $f^*(\mathbf{x}_H^*(k), p): \mathbb{R}^{3N} \times \mathbb{N}^1 \to \mathbb{R}^{3N}$ is assumed to be an analytical function, representing the transition of the human motion.
The noise $w_k \in \mathbb{R}^{3N}$ is zero-mean Gaussian and white. The nonlinear function $f^*(\mathbf{x}_H^*(k), p)$ is modeled using a two-layer Neural Network 
\begin{align}
f^*(\mathbf{x}_H^*(k), p)  = W^T \max(0, U^Ts_k) + \epsilon(s_k),\label{eq: T1 NN}
\end{align}
where $s_k = [\mathbf{x}_H^*(k)^T, p^T,1]^T \in \mathbb{R}^{3N+2}$ is the input vector to the Neural Network. $\max(0, U^Ts_k)$ is the activation function. $\epsilon(s_k) \in \mathbb{R}^{3N}$ is the function reconstruction error that goes to zero when the neural network is fully trained. $W\in\mathbb{R}^{n_h\times 3N}$, and $U\in\mathbb{R}^{(3N+2)\times n_h}$ where $n_h \in \mathbb{N} $ is the number of neurons in the hidden layer of the neural network \cite{ravichandar2017human}.  
\begin{figure}[tb]
	\begin{center}
		\includegraphics[width=.6\linewidth]{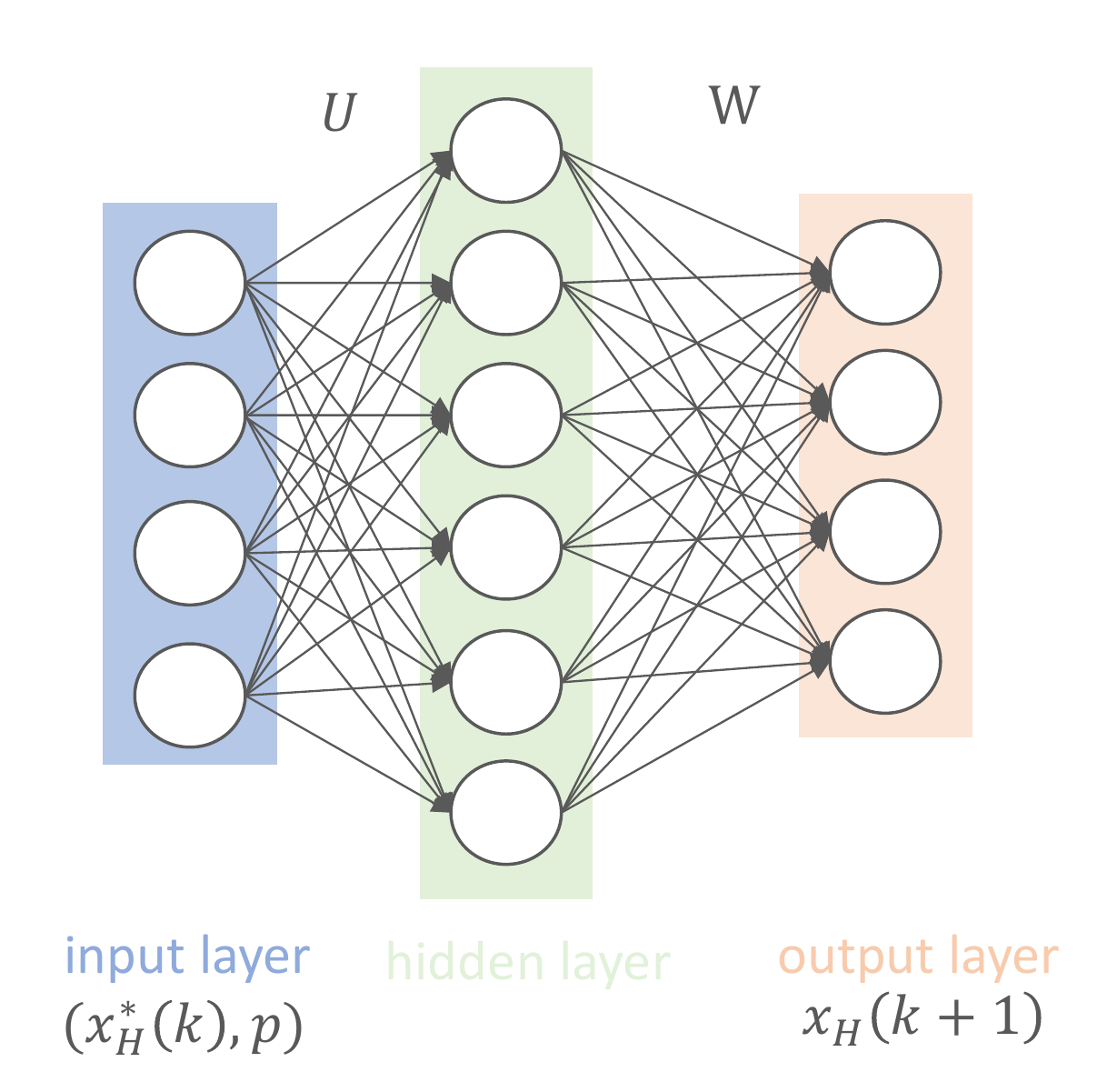}
	\end{center}
	\caption{Two layer neural network for human motion transition model. The activation function for hidden layer is $x\mapsto \max(0,x)$.}
	\label{fig:NN.png}
\end{figure}

\subsubsection{Quantifying Uncertainty in Prediction}
Due to humans' time varying behaviors and individual differences among difference people, the state transition model is also adapted online. For simplicity, we only adapts $W$ and keeps $U$ constant. To provide safety guarantees, we also quantify the uncertainty of $W$ during online adaptation \cite{liu2015safe}. Reshape the matrix $W$ to a vector $\theta\in\mathbb{R}^{3Nn_h}$ by stacking all the column vectors of $W$. To account for time varying behaviors, $\theta$ is considered a time varying parameter, where $\theta_k$ denotes its value at time step $k$. Define a new data matrix $\Phi_k \in\mathbb{R}^{3N\times 3Nn_h}$ as a diagonal concatenation of $N$ pieces of $\max(0, U^Ts_k)^T$. 
Using $\Phi_k$ and $\theta_k$, \eqref{eq: T1 transition model} and \eqref{eq: T1 NN} can be written as
\begin{equation}
\mathbf{x}_H(k+1) =\Phi_k \theta_k+w_k,\label{eq:Transformed LTV}
\end{equation}

Let $\hat{\theta}_k$ be the estimate of ${\theta_k}$ at time step $k$ and $\tilde{\theta}_k=\theta_k-\hat{\theta}_k$
be the estimation error at step $k$.

\paragraph{State estimation}
The \textit{a priori} estimate of the state and the estimation error is
\begin{align}
\hat{\mathbf{x}}_H\left(k+1|k\right) =&\Phi_k\hat{\theta}_k,\label{eq:prediction}\\
\tilde{\mathbf{x}}_H\left(k+1|k\right) =&\Phi_k\tilde{\theta}_k+w_k.
\end{align}

Since $\hat{\theta}_k$ only contains information up to the $\left(k-1\right)$th time step, $\tilde{\theta}_k$
is independent of $w_k$. Thus the \textit{a priori} mean squared estimation error (MSEE) $X_{\tilde{x}\tilde{x}}\left(k+1|k\right)=E\left[\tilde{\mathbf{x}}_H\left(k+1|k\right)\tilde{\mathbf{x}}_H\left(k+1|k\right)^{T}\right]$
is
\begin{equation}
X_{\tilde{x}\tilde{x}}\left(k+1|k\right)=\Phi_kX_{\tilde{\theta}\tilde{\theta}}(k)\Phi^{T}_k+Var(w_k),\label{eq: uncertainty matrix T1}
\end{equation}
where $X_{\tilde{\theta}\tilde{\theta}}(k)=E\left[\tilde{\theta}_k\tilde{\theta}_k^{T}\right]$
is the mean squared error of the parameter estimation.

\paragraph{Parameter estimation}
The parameter is estimated as
\begin{equation}
\hat{\theta}_{k+1}=\hat{\theta}_k+F_k\Phi^{T}_k\tilde{\mathbf{x}}_H\left(k+1|k\right),
\end{equation}
where $F_k$ is the learning gain. Since the system is time varying, $\Delta\theta_k=\theta_{k+1}-\theta_k\neq 0$. The parameter estimation error is
\begin{equation}
\tilde{\theta}_{k+1}=\tilde{\theta}_k-F_k\Phi^{T}_k\tilde{\mathbf{x}}_H\left(k+1|k\right)+\Delta\theta_k\label{eq: belief space paa}.
\end{equation}
The estimated parameter is biased and the expectation of the error can be expressed as
\begin{align}
E\left(\tilde{\theta}_{k+1}\right)= &\left[I-F_k\Phi^{T}_k\Phi_k\right]E\left(\tilde{\theta}_k\right)+\Delta\theta_k\nonumber \\
  = & \sum_{n=0}^{k}\prod_{i=n+1}^{k}\left[I-F_i\Phi^{T}\left(i\right)\Phi\left(i\right)\right]\Delta\theta_n\label{eq:expectation of parameter estimation error}.
\end{align}

The mean squared error of parameter estimation follows from (\ref{eq: belief space paa}) and (\ref{eq:expectation of parameter estimation error}):
\begin{align}
& X_{\tilde{\theta}\tilde{\theta}}\left(k+1\right) \nonumber\\
 = & F_k\Phi^{T}_kX_{\tilde{x}\tilde{x}}\left(k+1|k\right)\Phi_kF_k -X_{\tilde{\theta}\tilde{\theta}}(k)\Phi^{T}_k\Phi_kF_k -F_k\Phi^{T}_k\Phi_kX_{\tilde{\theta}\tilde{\theta}}(k)\nonumber \\
& +E\left[\tilde{\theta}_{k+1}\right]\Delta\theta^{T}_k+\Delta\theta_kE\left[\tilde{\theta}_{k+1}\right]^{T} -\Delta\theta_k\Delta\theta_k^{T}+X_{\tilde{\theta}\tilde{\theta}}(k).\label{eq:parameter-estimation-error-covariance}
\end{align}

Since $\Delta\theta_k$ is unknown in (\ref{eq:expectation of parameter estimation error}) and (\ref{eq:parameter-estimation-error-covariance}), it is set to an average time varying rate $d\theta$ during implementation.

At step $k$, the predicted trajectory $\hat{\mathbf{x}}_H(k+1|k)$ together with the uncertainty matrix $X_{\tilde{x}\tilde{x}}(k+1|k)$ is then sent to T3 to generate the safety constraint \eqref{eq: chap2 design of knowledge}.

\section{T2: Task Planning\label{sec: T2}}

\begin{figure}[t]
	\centering
	\includegraphics[width=1\linewidth]{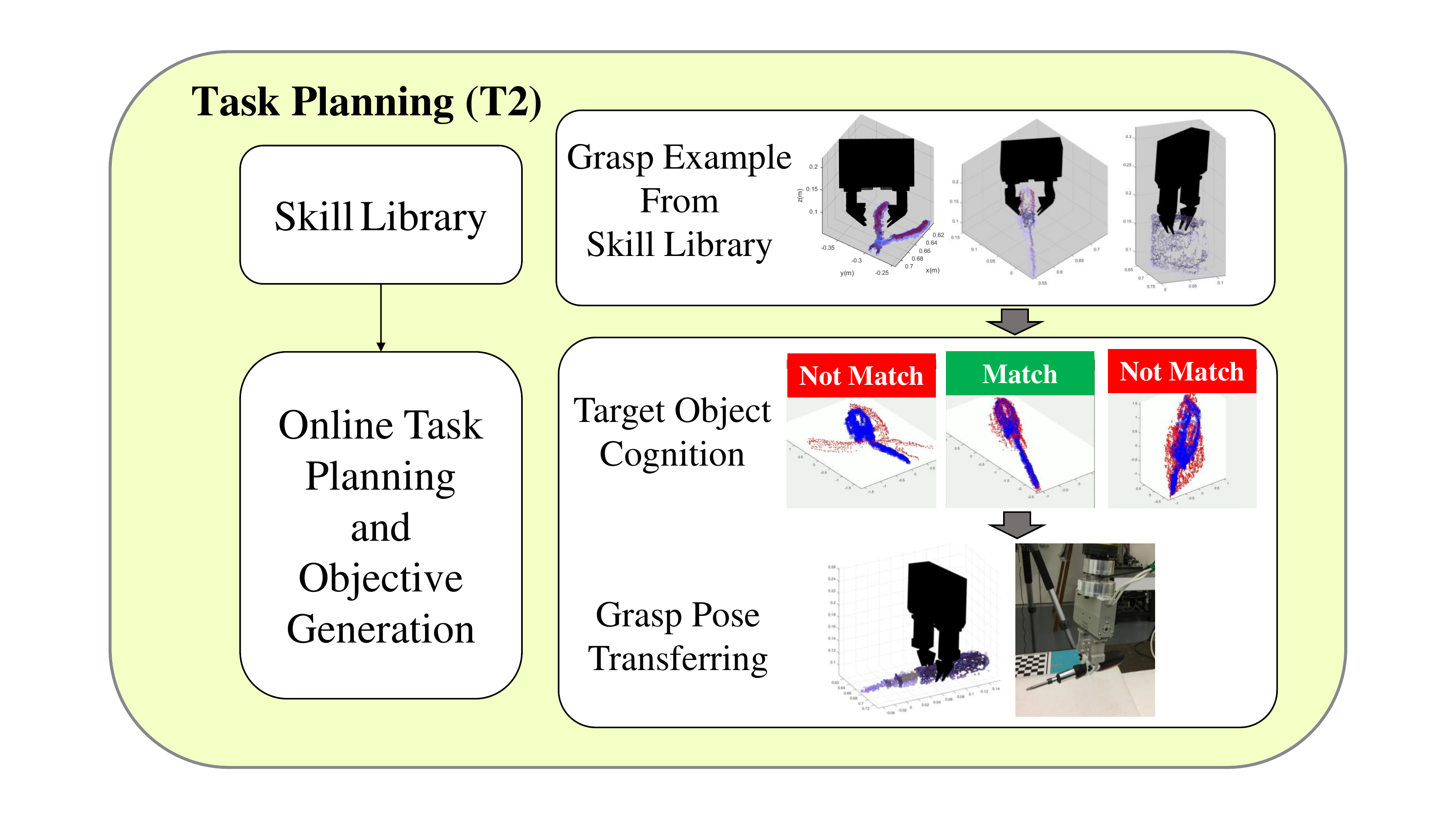}
	\caption{The task planning with object grasping as an example.}
	\label{fig:t2-framework}
\end{figure}

Given the prediction of the human plan as well as the environment information, the objective of task planning is to generate a set of reference actions of robots to assist the human. 
As shown in the aforementioned example in \cref{fig: performance}c, the robot realizes that the human needs a tool to fasten the assembly, then it grasps and delivers the correct tool to the human. In this scenario, the task planning module needs to find the appropriate object from the clustered environment and determines the best pose for grasping. The framework of task planning is shown in detail in \cref{fig:t2-framework}. A skill library for robot grasping is trained offline, which stores grasp examples learned from human demonstration. During online operation, the robot registers the target object with a learned sample, and transforms the grasp pose on the sample to a grasp pose on the target object.


\begin{figure}[t]
	\centering
	\includegraphics[width=1\linewidth]{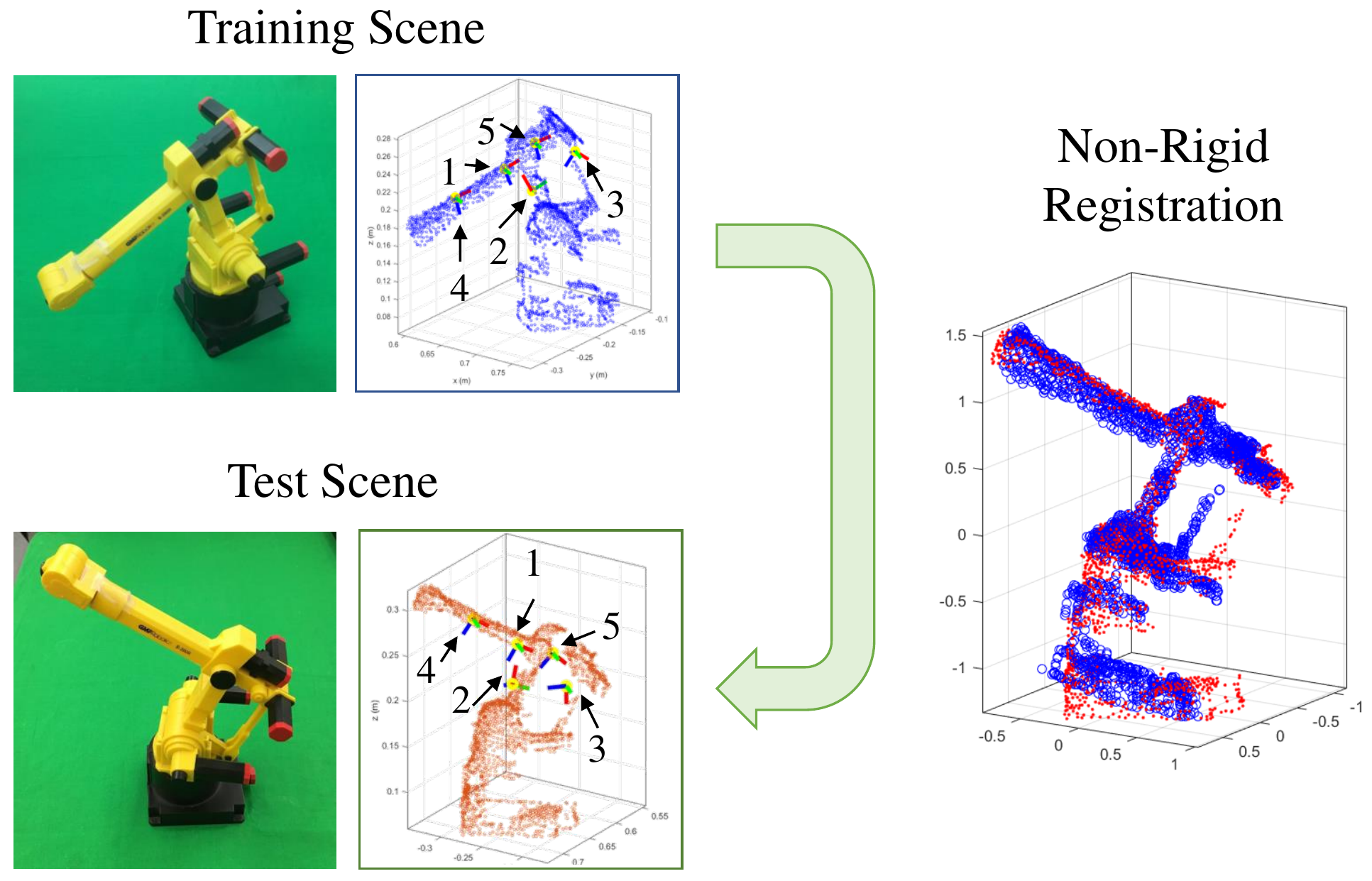}
	\caption{The robot grasp transfer through non-rigid registration.}
	\label{fig:grasp transfer}
\end{figure}

Although the stored grasp examples can provide good grasping points on the source objects among the examples, it is non-trivial and not data-efficient to collect grasp examples for all objects. We classify objects in a typical assembly line into several categories. Objects in each category share similar topological structures but may be different in shape and size. In this paper, the category of the target object will be classified by its similarities towards the source objects. Then the grasp pose is transferred from the identified source object to the target object. Both the similarity measure and the grasp pose transformation require the alignment between the target object and the source object. This task is formulated as a point set registration problem. An example concerning grasping of a toy robot manipulator is given in \cref{fig:grasp transfer}. In the training scene, five good grasp poses are taught on the toy manipulator and labeled in the point cloud. In the test scene, the same toy manipulator with different configuration is given. The point registration aligns the point cloud of the trained object with that of the target object, and the corresponding transformation maps the grasp pose from the source object to the target object. 

Given variation and deformation between the source object and the target object, the mapping should be more flexible than rigid transformation. In the meantime, the topological structure of the point sets must be preserved during the alignment so that the grasp pose can be transferred to a reasonable location. In this work, we use the coherent point drift (CPD) algorithm~\cite{myronenko2010point} to perform a smooth non-rigid registration.

\subsection{Non-Rigid Point Registration by Coherent Point Drift}
Assume the source object and the target object are represented by two point cloud set, $\mathbf{P}^s = (p_1^s, \cdots, p_N^s)\in\mathbb{R}^{N\times D}$ and $\mathbf{P}^t = (p_1^t, \cdots, p_M^t)\in\mathbb{R}^{M\times D}$, where $p_n^s$ and $p_m^t$ are the $n$-th source point and $m$-th target point, respectively.
In order to align the source object toward the target object, CPD considers source points in $\mathbf{P}^s$ as the centroids of Gaussian mixtures, and transforms them to fit the target points in $\mathbf{P}^t$ coherently.
The source points are assumed to deform toward the target points according to a continuous displacement field $v(\cdot)$, and the transformed source point is written as 
\begin{align}
\mathcal{T}(p^s_n) = p^s_n+v(p^s_n), 
\end{align} 
where $\mathcal{T}:\mathbb{R}^D\rightarrow\mathbb{R}^D$ is a non-rigid transformation.
The goal of CPD is to retrieve the displacement field $v$ that maximizes the alignment between the target point set $\mathbf{P}^t$ and the transformed source point set $\mathcal{T}(\mathbf{P}^s)$. The alignment is modeled by the Gaussian mixture model, where each point in $\mathcal{T}(\mathbf{P}^s)$ serves as a Gaussian centroid and likelihood of $\mathbf{P}^t$ sampling from the Gaussian mixture can be quantitatively analyzed. Meanwhile, a smoothness regularization on the transformation $\mathcal{T}$ is imposed, which constraint regularizes the points $\mathbf{P}^s$ to move coherently and have a smooth deformation to its neighbors. The log-likelihood function of the Gaussian mixture model with smoothness regularization can be constructed as,
\begin{align}
\label{eq:logL+regV}
L(v,\sigma^2) = -\sum_{m=1}^{M}\log\sum_{n=1}^{N}\exp \left( \frac{-1}{2\sigma} \Vert p_m^t  - \mathcal{T}(p_n^s) \Vert^2 \right) + \frac{\lambda}{2} \Vert v \Vert_{\mathcal{F} }^2,
\end{align}
where the first term penalizes the deviation between target points and source points after applying transformation, and the second term regularizes the function smoothness by a frequency domain norm, $\lVert v \rVert_{\mathcal{F}}^2 = \int_{\mathbb{R}^D} \frac{|V(s)|^2} {G(s)} ds$~\cite{girosi1995regularization}.  $V(s)$ is a Fourier transform of $v$ and $G(s)$ presents a symmetric filter that approaches to zero as $s \rightarrow \infty$. The overall Fourier domain norm here basically captures the energy of high frequency components of $V(s)$. Intuitively, the larger the norm $||v||_{\mathcal{F}}$, the more `oscillating' $v$ will be, i.e., less smoothness. $\lambda\in\mathbb{R}^+$ is a weighting coefficient that represents the trade off between the fitting of the point sets and the smoothness constraints on the transformation.  

It can be proved by variational calculus that the optimizer of \eqref{eq:logL+regV} has the form of the radial basis function~\cite{myronenko2010point},
\begin{align}
v(z) = \sum_{n=1}^{N}w_n g(z-p^s_n), \label{eq:kernel}
\end{align}
where $g(\cdot)$ is a kernel function retrieved from the inverse Fourier transform of $G(s)$, and $w_n$ is the unknown kernel weights. In general, $g(\cdot)$ can be any formulation with positive definiteness, and $G(s)$ behaves like a low-pass filter. For simplicity, a Gaussian kernel is chosen so that $g(z-p^s_n) =  \exp(-\frac{1}{2\beta^2}||z-p^s_n||^2 )$, where $\beta\in\mathbb{R}^+$ is a parameter that defines the width of smoothing Gaussian filter. Larger $\beta$ corresponds to more rigid transformation, whereas smaller $\beta$ produces more local deformation. 
Substituting \eqref{eq:kernel} to \eqref{eq:logL+regV}, the regularized negative log-likelihood function can be further derived to
\begin{align}
L(\mathbf{W},\sigma^2) &= \frac{-1}{2\sigma^2}\sum_{n=1}^{N}\sum_{m=1}^{M}P(n|p^t_m)\lVert p^t_m - p^s_n - \sum_{k=1}^{N}w_k g(p^s_n-p^s_k)\rVert^2 \nonumber \\
&- \frac{D}{2}\sum_{n=1}^{N}\sum_{m=1}^{M}P(n|p^t_m)\log\sigma^2 - \frac{\lambda}{2}tr(\mathbf{W}^T \mathbf{G}\mathbf{W}), \label{eq:logQ+WGW} 
\end{align}
where $\mathbf{G}\in\mathbb{R}^{N\times N}$ is a Gramian matrix with element $\mathbf{G}_{ij} = g(x_i - x_j)$ and $\mathbf{W} = \left[w_1,\cdots, w_n \right]^T\in\mathbb{R}^{N\times D}$ is the vectorization of kernel weights in \eqref{eq:kernel}. 

Equation~\eqref{eq:logQ+WGW} is now parameterized by $(\mathbf{W},\sigma^2)$, and the EM algorithm can be performed to estimate the parameters by iteratively minimizing the negative log-likelihood function~\cite{dempster1977maximum}. 

\textbf{E-step}: The posterior probability $P(n|p^t_m)$ is calculated by using the previous estimated parameters. To add robustness to outliers, an additional uniform probability distribution is added into the mixture model, and the posterior is given by
\begin{align}
	p(n|p^t_m) = \frac{\exp\left(-\frac{\|p^t_m - p^s_n - v(p^s_n)\|^2}{2\sigma^2} \right)}{\sum_{n=1}^{N}\exp\left(-\frac{\|p^t_m - p^s_n - v(p^s_n) \|^2}{2\sigma^2} \right) + (2\pi\sigma^2)^{D/2}\frac{\mu}{(1-\mu)}\frac{N}{M}}\ ,
\label{eq:estep}
\end{align}
where $\mu\in[0,1]$ reflects the amount of outliers. 

\textbf{M-step}: Take $\partial L/\partial \mathbf{W} = 0$ and $\partial L / \partial \sigma^2 = 0$ to obtain a new estimate of $(\mathbf{W},\sigma^2)$. The closed-form solution for M-step requires further mathematical derivation, more details can be found in~\cite{myronenko2010point, myronenko2007non}. 

After $L$ is converged, the point set of the source object $\mathbf{P}^s$ can be aligned toward the target object by
\begin{align}
\mathcal{T}(\mathbf{P}^s) = \mathbf{P}^s + \mathbf{G}\mathbf{W}.
\end{align}
The transformation $\mathcal{T}$ is further used in measuring the object similarity as well as the grasp pose transferring.

\subsection{Target Object Cognition by Similarity Measure}

Given a desired source object category to grasp, the robot needs to find the target among all the object candidates placed in the workspace. 
By measuring the similarity between the source object $\mathbf{P}^s$ and each target object candidate $\mathbf{P}^t$, the most similar pair will be selected to determine target object to grasp. In our work, since CPD can be applied to warp the source points $\mathbf{P}^s$ to $\mathcal{T}(\mathbf{P}^s)$ which is aligned with $\mathbf{P}^t$, the residual similarity between $\mathcal{T}(\mathbf{P}^s)$ and $\mathbf{P}^t$ instead of the similarity between $\mathbf{P}^s$ and $\mathbf{P}^t$ will be checked to provide a more robust category classification.

The average minimum distance between the two point sets can be designed as:
\begin{align}
d(\mathcal{T}(\mathbf{P}^s),\mathbf{P}^t) = \frac{1}{N}\sum_{n=1}^{N} \min_{m\in[1,M]} ||\mathcal{T}(p^s_n) - p^t_m||, \label{eq:dissilmilar}
\end{align}
where $|| \mathcal{T}(p^s_n)  - p^t_m ||$ is the Euclidean distance between point $\mathcal{T}(p^s_n)$ and $p^t_m$. Equation~\eqref{eq:dissilmilar} is an error function that is commonly used for point cloud alignment. However, \eqref{eq:dissilmilar} is asymmetric. The similarity between a source object and a target object can be formulated as
\begin{align}
D(\mathbf{P}^{s\prime},\mathbf{P}^t) = d(\mathbf{P}^{s\prime},\mathbf{P}^t) + d(\mathbf{P}^t, \mathbf{P}^{s\prime}), \label{eq:D}
\end{align}
where $\mathbf{P}^{s\prime} = \mathcal{T}(\mathbf{P}^s)$ is the source points warped toward $\mathbf{P}^t$ by CPD. The function $D(\cdot,\cdot)$ sums the two asymmetric similarity measurements together so that $D$ is symmetric to its input arguments, i.e. $D(\mathbf{P}^{s\prime},\mathbf{P}^t) = D(\mathbf{P}^t,\mathbf{P}^{s\prime})$. 

Suppose there are $K$ object candidates, the most possible that the target object to grasp is determined by
\begin{align}
	\mathbf{k}^* = \arg\min_{k\in[1,K]} \  D(\mathbf{P}^{s\prime},\mathbf{P}^t_k ).
\end{align} 
\subsection{Grasp Pose Transferring}
After finding the target object to grasp, the mapping from $\mathbf{P}^s$ to the $\mathbf{P}^t$ is also calculated through CPD. As shown in Fig.~\ref{fig:grasp transfer}, the demonstrated grasp poses on $\mathbf{P}^s$ will also be transferred to achieve new grasp poses that are suitable for object $\mathbf{P}^t$. 

Denote the grasp poses as $\bm{g} = (\bm{t},\bm{R}) \in \mathbb{R}^D\otimes \mathbf{SO}(D)$, where $\bm{t}\in\mathbb{R}^D$ is the center of the grasping point, $\bm{R}\in \mathbf{SO}(D)$ represents the grasping orientation.

The grasp pose transformation can be decomposed to two parts: the position transformation and the orientation transformation.
Regards to the position transformation, the non-rigid transformation $\mathcal{T}(\cdot)$ can directly map the center of grasp from grasp example to the target object by
\begin{align}
\bm{t} \leftarrow \mathcal{T}(\bm{t}^s),
\end{align}
where the superscript $s$ denotes as the grasp on the source object. 
As for the orientation, it can be considered as transferring $x,y$, and $z$ axes of the original grasp orientation to the new object space. One natural way to transform a vector $\mathbf{v}$ at a point $\bm{t}$ through a function is to multiply the vector with the gradient of $\mathcal{T}(\bm{t})$~\cite{abraham1978foundations}, i.e. $\nabla \mathcal{T} (\bm{t}) \bm{v}$. Considering the properties of the special orthogonal group, the new orientation of the grasp is constructed by the singular value decomposition (SVD),
\begin{align}
\bm{R} \leftarrow \mathbf{U}\mathbf{V}^T,
\end{align}
where $\mathbf{U}\Sigma\mathbf{V}^T = svd(\nabla\mathcal{T}(\bm{t}^s)\bm{R})$,  $\mathbf{U}, \mathbf{V}$ are the orthonormal basis of the matrix, and $\Sigma$ is a diagonal matrix that consists of the singular values of the matrix.  

Hence, the new grasp pose on the target can be transferred by
\begin{align}
\bm{g} = (\bm{t}, \bm{R}) \leftarrow (\mathcal{T}(\bm{t}),\mathbf{U}\mathbf{V}^T ).
\end{align}
The transferred grasp pose is then sent to T3 for motion planning. For example, given a desired grasp pose, the objective function $J$ in \eqref{eq: chap2 design of knowledge} is designed to be
\begin{equation}
J(\mathbf{x}_R,\mathbf{x}_H,\mathbf{x}_e) = \int_t^{t+T} \|x_R(\tau) - \bm{g}\|^2 d\tau + \int_t^{t+T} \|\dot x_R(\tau)\|^2 d\tau.
\end{equation}

\section{T3: Motion Planning\label{sec: T3}}
Given the information from T1 and the task plan from T2, the objective of motion planning is to generate safe and efficient motions to realize the task plan in order to assist human. 
As it is computationally expensive to obtain the optimal solution of the motion planning problem (\ref{eq: chap2 design of knowledge}) for all scenarios offline, the optimization problem is computed online given information obtained in real time. However, there are two major challenges in real time motion planning. The first challenge is the difficulty to plan a safe and efficient trajectory when there are large uncertainties, especially in humans' behaviors. As the uncertainty accumulates, solving the problem (\ref{eq: chap2 design of knowledge}) in the long term might make the robot's motion very conservative. The second challenge is the difficulty to compute the trajectory in real time with limited computation power since the problem (\ref{eq: chap2 design of knowledge}) is highly non-convex. We design a unique parallel planning and control architecture \cite{liu2017real} to address the first challenge and develop fast online optimization solvers to address the second challenge.

\subsection{The Parallel Planning and Control Architecture}
There are two planning themes to generate robot motion, long term planning and short term planning. 
In the long term planning, accumulation of uncertainty will make the robot motion very conservative. On the other hand, the uncertainty will not accumulate too much for a short term planner. However, using a short term planner alone is also problematic. The robot can easily get stuck in local optima, due to lack of a global perspective. Although it is possible to construct a globally-converging local policy for robots with simple dynamics in specific environments \cite{savkin2015safe}, it is in general hard to obtain a globally converging local policy for robots with complicated dynamics in complicated environments. 

\begin{figure}[t]
\begin{center}
\includegraphics[width=5cm]{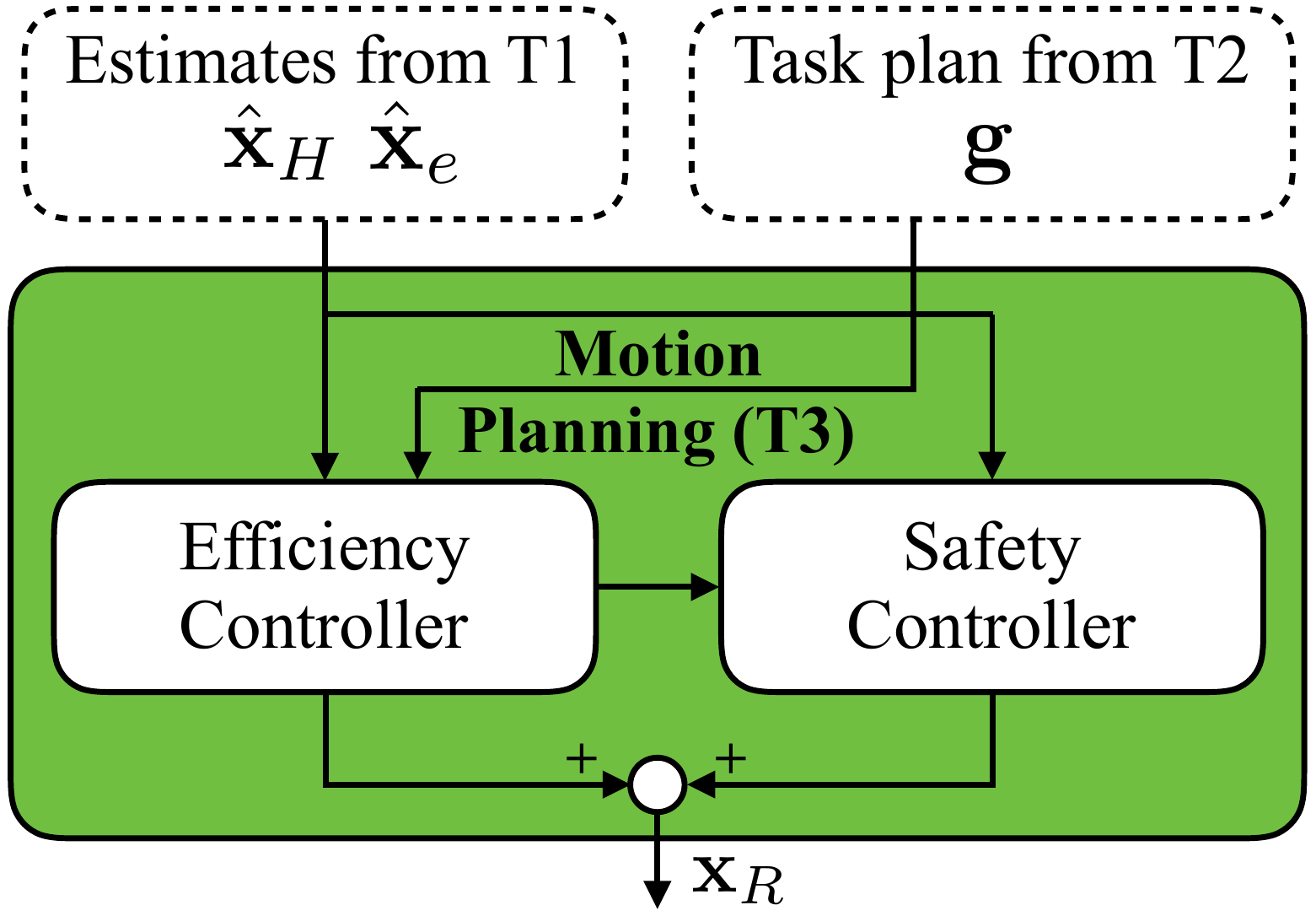}
\caption{The parallel planning structure with an efficiency controller for long term planning and a safety controller for short term planning and control.}
\label{fig: parallel diagram}
\end{center}
\end{figure}

This paper adopts a parallel planner which consists of a long term (global) planner as well as a short term (local) planner to leverage the benefits of the two planners. The idea is to have the long term planner solving (\ref{eq: chap2 design of knowledge}) without considering uncertainties, and have the short term planner addressing uncertainties. The long term planning is efficiency-oriented and is called the efficiency controller, while the short term planning is safety-oriented and is called the safety controller. The two controllers run in parallel as shown in the block diagram \cref{fig: parallel diagram}.  

\begin{figure}[t]
\begin{center}
\includegraphics[width=8.5cm]{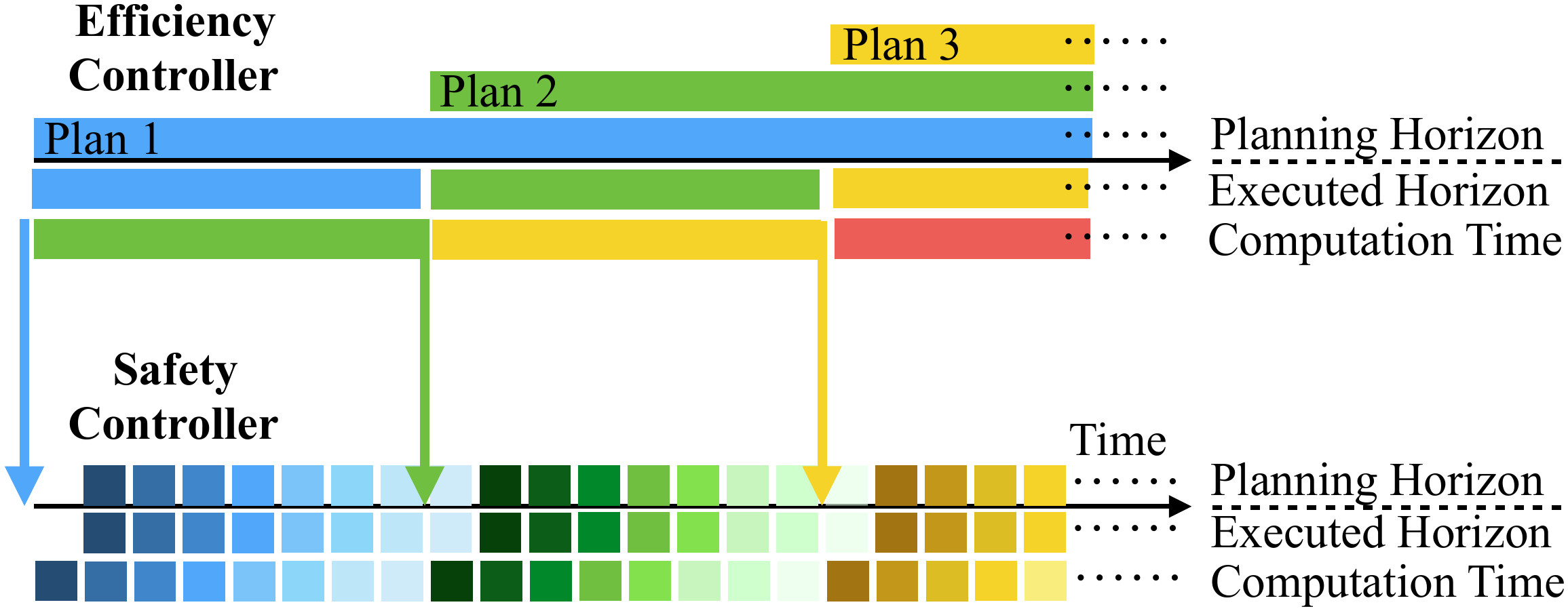}
\caption{The time flow in the parallel planners.}
\label{fig: parallel time flow}
\end{center}
\end{figure}

The computation time flow for the parallel planners is shown in \cref{fig: parallel time flow}, together with the planning horizon and the execution horizon. Three long term plans are shown, each with one distinct color. The upper part of the time axis shows the planning horizon. The middle layer is the execution horizon. Only a portion of the planned trajectory is executed. The bottom layer shows the computation time. The computation is done before the execution of the plan. Once computed, a long term plan is sent to the safety controller for monitoring. The mechanism in the safety controller is similar to that in the efficiency controller. The planning horizon, the executed horizon and the computation time for the same short term plan are shown in the same color. A short term plan can be computed with shorter time. The sampling rate in the safety controller is much higher than that in the efficiency controller. Though the execution horizon in the safety controller is one time step, the planning horizon is not necessarily one time step. 

This approach can be regarded as a two-layer model predictive control (MPC) approach. Coordination between the two layers is important. To avoid instability, a margin is needed for the safety constraint in the efficiency controller so that the long term plan will not be revoked by the safety controller if the long term prediction of the human motion is correct. 
Nonetheless, the successful implementation of the parallel control architecture highly depends on computation. It is important that the optimization algorithm finds a feasible and safe trajectory within the sampling time. The algorithms for real time non-convex optimization will be discussed in \Cref{sec: cfs,sec: ssa}.

\subsection{Efficiency-Oriented Long Term Planning\label{sec: cfs}}

\begin{figure}[t]
\begin{center}
\subfloat[Iteration 1.\label{fig: cfs illustration}]{
\includegraphics[width=4cm]{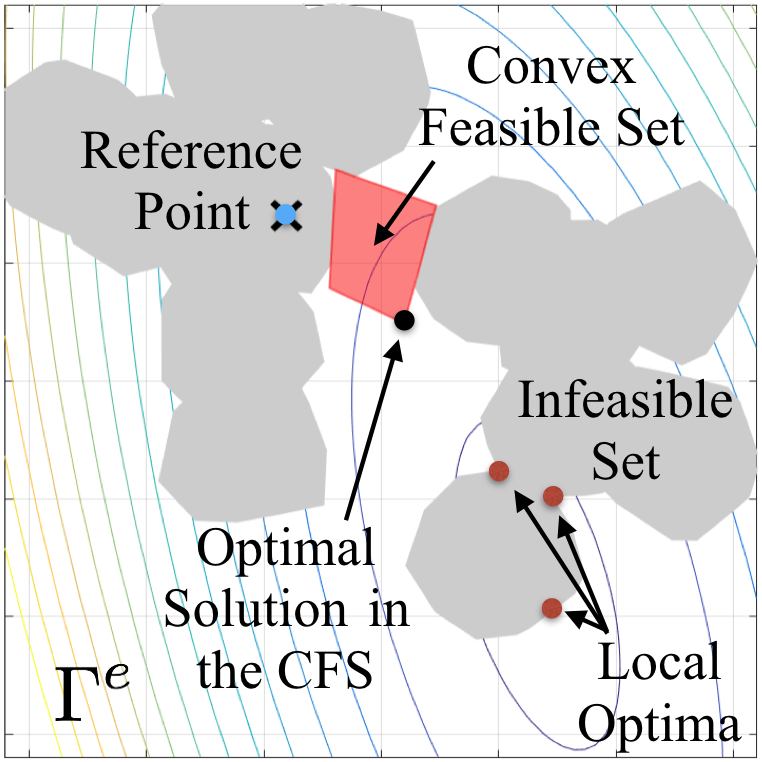}}
\subfloat[Iteration 2.\label{fig: cfs illustration 2}]{
\includegraphics[width=4cm]{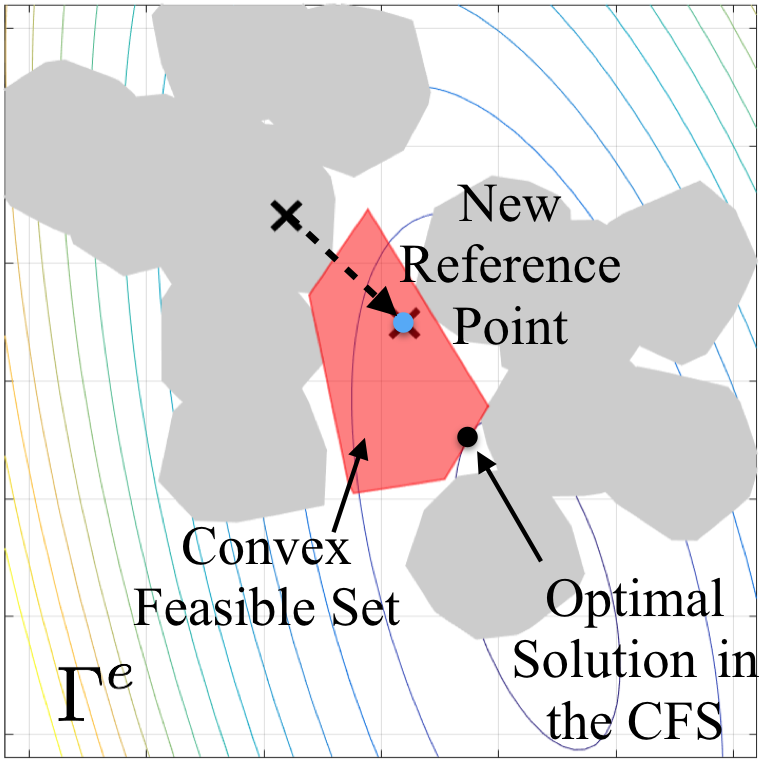}}\\
\subfloat[Iteration 3.\label{fig: cfs illustration 3}]{
\includegraphics[width=4cm]{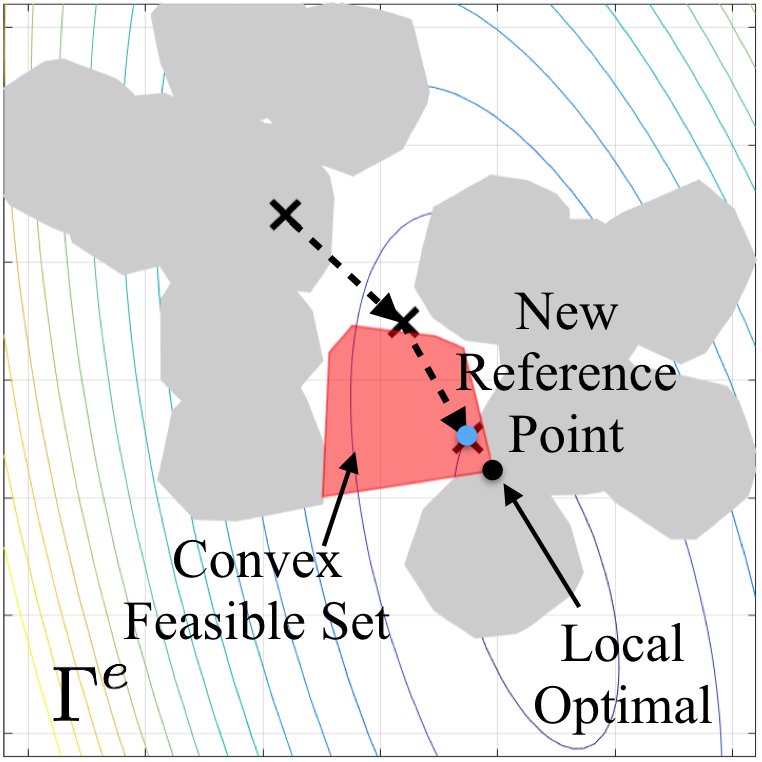}}
\subfloat[Converge to a local optimum.\label{fig: cfs illustration 4}]{
\includegraphics[width=4cm]{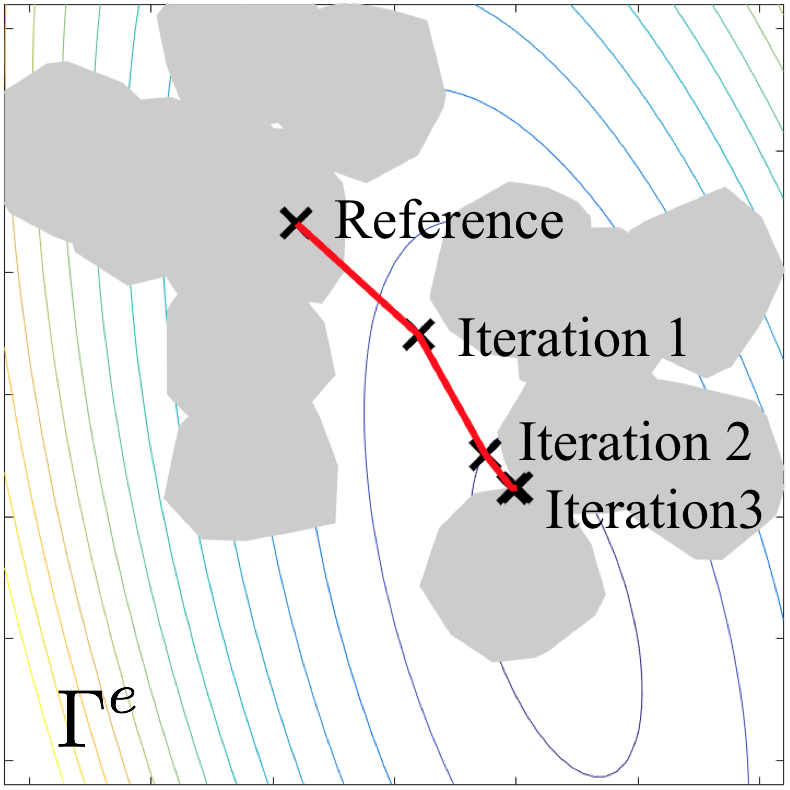}}
\caption{Illustration of the convex feasible set algorithm.}
\label{fig: cfs}
\end{center}
\end{figure}

The optimization problem (\ref{eq: chap2 design of knowledge}) in a clustered environment is highly nonlinear and non-convex, which is hard to solve in real time even without consideration of the uncertainties. Generic non-convex optimization solvers such as sequential quadratic programming (SQP) \cite{boggs1995sequential} may not meet the real time requirement as they neglect the unique geometric features of the problem. A convex feasible set (CFS) algorithm \cite{liu2017sicon} is proposed to convexify the problem considering the geometric features. For simplicity, this sub-section assumes that the cost function is convex with respect to the robot state and control, and the system dynamics \eqref{eq: robot dynamics} are linear. The method to convexify a problem with nonlinear affine dynamics is discussed in \cite{liu2017real}. 

\subsubsection{Convexification of the Motion Planning Problem\label{sec: traj smoothing problem formulation}}
Sample the continuous robot trajectory $\mathbf{x}_R = x_R(t:t+T)$ by rate $t_s$. For simplicity, set current time $t=0$. Denote the variables at time step $k$ (or time $kt_s$) as $x_R(k)$, $u_R(k)$, $x_H(k)$, and $x_e(k)$. Let $\mathbf{x}_R^d$ and $\mathbf{u}_R^d$ denote the discrete trajectories for robot state and robot control at time step $0,1,\ldots,N-1$. The predicted trajectories from T1 are $\hat{\mathbf{x}}_H$ and $\hat{\mathbf{x}}_e$ which contains predictions at step $1,2,\ldots,N$. As the system dynamics are linear and observable, $\mathbf{u}_R^d$ can be computed from $\mathbf{x}_R^d$, i.e., $\mathbf{u}_R^d = \mathcal{L}(\mathbf{x}_R^d)$ for some linear mapping $\mathcal{L}$.   
Rewriting (\ref{eq: chap2 design of knowledge}) in the discrete time as
\begin{equation}\label{eq: benchmark smoothing chp 5}
\min_{\mathbf{x}_R^d\in \Gamma^e}~J^d(\mathbf{x}_R^d),
\end{equation}
where $J^d(\mathbf{x}_R^d)$ is the discretized cost function. When the sampling time $t_s$ goes to zero, $J^d(\mathbf{x}_R^d) = J(\mathbf{x}_R, \hat{\mathbf{x}}_H,\hat{\mathbf{x}}_e)$. The constraint $\Gamma^e : = \{\mathbf{x}_R^d: \mathbf{u}_R^d = \mathcal{L}(\mathbf{x}_R^d)\text{, and } u_R(k)\in\Omega\text{, } (x_R(k),\hat{x}_H(k),\hat{x}_e(k))\in {X}_S\text{, } \forall k\}$, which corresponds to constraints \eqref{eq: chap2 behavior design self constraint} and \eqref{eq: chap2 behavior design interaction constraint}.  Since $J$ is convex, $J^d$ is also convex. The non-convexity mainly comes from the constraint $\Gamma^e$. The geometry of the problem is illustrated in \cref{fig: cfs}. The contour represents the cost function $J^d$, while the gray parts represent the complement of $\Gamma^e$. The goal is to find a local optimum (hopefully global optimum) starting from the initial reference point (blue dot). To make the computation more efficient, we transform the problem into a sequence of convex optimizations by obtaining a sequence of convex feasible sets inside the non-convex domain $\Gamma^e$. As shown in \cref{fig: cfs}, the idea is implemented iteratively. At current iteration, a convex feasible set for the current reference point (blue dot) is obtained. The optimal solution in the convex feasible set (black dot) is set as the reference point for the next iteration.

\subsubsection{Convex Feasible Set Algorithm\label{sec: application}}
The general method in constructing convex feasible set is discussed in \cite{liu2017sicon}. As $\mathcal{L}$ is linear and $\Omega$ is convex, we only need to convexify the safety constraint $(x_R(k),\hat{x}_H(k),\hat{x}_e(k))\in {X}_S$. 
For each time step $k$, the infeasible set in the robot's state space is $\mathcal{O}_k:=\{x_R(k):(x_R(k),\hat{x}_H(k),\hat{x}_e(k))\notin {X}_S\}$. 
Then the safety constraint in (\ref{eq: benchmark smoothing chp 5}) is equivalent to $d^*(x_R(k),\mathcal{O}_k)\geq 0$ where $d^*(x_R(k),\mathcal{O}_k)$ is the signed distance function to $\mathcal{O}_k$ such that
\begin{equation}
d^*(x_R(k),\mathcal{O}_k):=\left\{\begin{array}{cc}
\min_{z\in\partial\mathcal{O}_k}\|x_R(k)-z\| & x_R(k)\notin\mathcal{O}_k\\
-\min_{z\in\partial\mathcal{O}_k}\|x_R(k)-z\| & x_R(k)\in\mathcal{O}_k
\end{array}\right..
\end{equation}
The symbol $\partial\mathcal{O}_k$ denotes the boundary of the obstacle $\mathcal{O}_k$. 

Note that if $\mathcal{O}_k$ is convex, then the function $d^*(\cdot,\mathcal{O}_k)$ is also convex. Hence $d^*(x_R(k),\mathcal{O}_k)\geq d^*(r_k,\mathcal{O}_k)+\nabla d^*(r_k,\mathcal{O}_k)(x_R(k)-r_k)$ for any reference point $r_k$. Then $d^*(r_k,\mathcal{O}_k)+\nabla d^*(r_k,\mathcal{O}_k)(x_R(k)-r_k)\geq 0$ implies that $x_R(k)\notin \mathcal{O}_k$. If the obstacle $\mathcal{O}_k$ is not convex, we then break it into several simple convex objects $\mathcal{O}^j_k$ such as circles or spheres, polygons or polytopes. The $\mathcal{O}^j_k$'s need not be disjoint. Then $d^*(\cdot,\mathcal{O}^j_k)$ is the convex cone of the convex set $\mathcal{O}_k^j$. Suppose a reference trajectory is $\mathbf{r}:=[r_0;r_1;\ldots;r_{N-1}]$, the convex feasible set $\mathcal{F}(\mathbf{r})$ for $\Gamma^e$ in (\ref{eq: benchmark smoothing chp 5}) is defined as
\begin{subequations}
\begin{align}
& \mathcal{F}(\mathbf{r}):=\{\mathbf{x}_R^d:  \mathbf{u}_R^d = \mathcal{L}(\mathbf{x}_R^d)\text{, and } u_R(k)\in\Omega\text{, }\\
&~~~~ d^*(r_k,\mathcal{O}_k^j)+\nabla d^*(r_k,\mathcal{O}_k^j)(x_R(k)-r_k)\geq 0, \forall k,j\},\label{eq: convex distance}
\end{align}
\end{subequations}
which is a convex subset of $\Gamma^e$.

Starting from an initial reference trajectory $\mathbf{x}_R^{(0)}$, the  convex optimization (\ref{eq: cfs}) needs to be solved iteratively until either the solution converges or the decrease in cost is small.
\begin{equation}
\mathbf{x}_R^{(i+1)} = \arg\min_{\mathbf{x}_R^d\in\mathcal{F}(\mathbf{x}_R^{(i)})} J^d(\mathbf{x}_R^d).\label{eq: cfs}
\end{equation}
It has been proved in \cite{liu2017sicon} that the sequence $\{\mathbf{x}_R^{(i)}\}$ converges to a local optimum of problem (\ref{eq: benchmark smoothing chp 5}). The computation time can be greatly reduced using the convex feasible set algorithm. This is due to the fact that we directly search for solutions in the feasible area. Hence 1) the computation time per iteration is smaller than existing methods as no linear search is needed, and 2) the number of iterations is reduced as the step size (change of the trajectories between two consecutive steps) is unconstrained.
Applications of the CFS algorithm can be found in \cite{liu2017convex}.

\subsection{The Safety-Oriented Short Term Planning\label{sec: ssa}}
Suppose a reference trajectory $\mathbf{x}_R^d$ is received from the efficiency controller. $u_R^o$ is the control input to execute the trajectory. The safety controller needs to ensure that the safety constraint \eqref{eq: chap2 behavior design interaction constraint} will be satisfied after applying this input. Hence the short-term planning problem can be formulated as the following optimization,
\begin{subequations}\label{eq: chap4 optimization for local planning}
\begin{align}
u_R^* = ~& \arg\min_{u_R} \|u_R-u_R^o\|^2_Q,\\
s.t.~& \mathbf{x}_R\in\Gamma\label{eq: chap4 optimization for local planning constraint},\\
& P\left(\left\{(x_R(t), x_H(t), x_e(t))\in {X}_S\right\}|\pi_R\right)=1,\forall t,
\end{align}
\end{subequations}
where $\|u_R-u_R^o\|_Q=(u_R-u_R^o)^T Q(u_R-u_R^o)$ penalizes the deviation from the reference input, where $Q$ should be designed as a second order approximation of the cost function $J$, e.g. $Q\approx d^2 J/d(u_R)^2$. The constraints are the same as the constraints in (\ref{eq: chap2 design of knowledge}). The safe set and the robot dynamics impose nonlinear and non-convex constraints which make the problem hard to solve. We propose to transform the non-convex state space constraint into convex control space constraint using invariant set. 

\subsubsection{The Safety Principle}
\begin{figure}[t]
\begin{center}
\subfloat[\label{fig: constraints ssa}]{
\includegraphics[width=3.3cm]{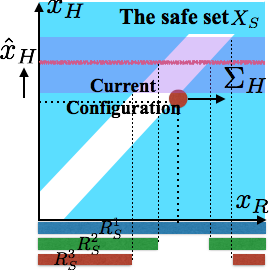}}
\subfloat[\label{fig: safe set}]{
\includegraphics[width=5.0cm]{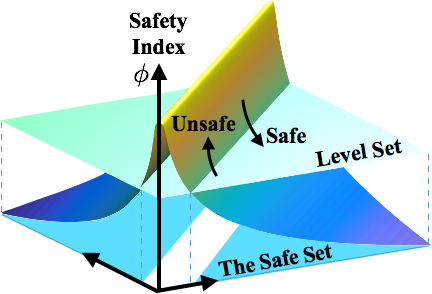}}
\caption{Illustration of the state space safety constraints $X_S$, $R_S^1$, $R_S^2$ and $R_S^3$.} \end{center}
\end{figure}

According to the safe set $X_{S}$, define the state space constraint $R_S$ for the robot as $R_S(x_H,x_e)=\{x_R:(x_R,x_H,x_e)\in X_S\}$, which depends on the human state and the environment state. Without loss of generality, we ignore $x_e$ in the following discussion for simplicity. Suppose the estimation of the human state is $\hat x_H$ and the uncertainty range is $\Gamma_H$ from T1, then the constraint on the robot state can be posed differently,
\begin{subequations}
\begin{align}
R_S^1= & \{x_R: x_R\in R_S(x_H) \text{ for some }x_H\},\label{eq: human take care}\\
R_S^2= & \{x_R: x_R\in R_S(\hat x_H)\},\label{eq: robot no uncertainty}\\
R_S^3= & \{x_R: x_R\in R_S(x_H),\forall x_H\in \Gamma_H\}.\label{eq: robot uncertainty}
\end{align}
\end{subequations}
In (\ref{eq: human take care}), it is assumed that the human will take care of the safety issue by choosing $x_H$ to satisfy the safety constraint $X_S$ given the robot state $x_R$. The robot only needs to make sure that the human always has such a choice. 
However, to make the system reliable, the safety problem should be taken care of by the robot as shown in (\ref{eq: robot no uncertainty}) given the estimate $\hat x_H$. To account for uncertainties, the robot state should be constrained in a smaller set (\ref{eq: robot uncertainty}). The set $\Gamma_H$ is computed from the MSEE $X_{\tilde{x}\tilde{x}}$ in \eqref{eq: uncertainty matrix T1}. In practice, we choose the $3\sigma$ set to bound the uncertainty \cite{liu2015safe}. \Cref{fig: constraints ssa} illustrates the safe set $X_S$ and the state space constraints $R_S^1$, $R_S^2$ and $R_S^3$. It is clear that $R_S^3\subset R_S^2\subset R_S^1$.

\textit{The safety principle} \cite{liu2015safe} requires that the robot control input $u_R(t)$ should be chosen such that $X_S$ is invariant, i.e., $x(t)\in X_S$ for all $t$. Given the uncertainty $\Gamma_H(t)$, we need to ensure that $x_R(t)\in R_S^3(t)$ for robust safety. 

\subsubsection{The Safety Index\label{sec: safety index}}

In addition to constraining the state in the safe region $R_S^3$, the robot should also be able to cope with any unsafe human movement. Given the current configuration in \cref{fig: constraints ssa}, if the human is anticipated to move upward, the robot should go right in order for the combined trajectory to stay in the safe set. To cope with the safety issue dynamically, a safety index is introduced as shown in \cref{fig: safe set}. The safety index $\phi:X\rightarrow\mathbb{R}$ is a function on the system state space such that
1) $\phi$ is differentiable with respect to $t$, i.e. $\dot\phi=(\partial\phi/\partial x )\dot x$ exists everywhere;
2) $\partial{\dot\phi}/\partial{u_R}\neq 0$;
3) The unsafe set $X\setminus X_S$ is not reachable given the control law $\dot\phi<0\left. when\right. \phi\geq 0$  and the initial condition $x(t_0)\in X_S$. 

The first condition is to ensure that $\phi$ is smooth. The second condition is to ensure that the robot input can always affect the safety index. The third condition provides a criterion to determine whether a control input is safe or not, e.g. all the control inputs that drive the state below the level set $0$ are safe and unsafe otherwise. The existence of such an index is proved in~\cite{liu2014control}.


\subsubsection{The Set of Safe Control}
To ensure safety, the robot's control must be chosen from the set of safe control
$U_S(t)=\{u_R(t):\dot\phi\leq-\eta_R\left. when\right. \phi\geq 0\}$
where $\eta_R\in \mathbb{R}^+$ is a safety margin. By the dynamic equation in \eqref{eq: robot dynamics}, the derivative of the safety index can be written as $\dot\phi=\frac{\partial \phi}{\partial x_R}hu_R+\frac{\partial \phi}{\partial x_R}f+\frac{\partial\phi}{\partial x_{H}}\dot{x}_{H}$. 
Then the set of safe control is
\begin{equation}
U_{S}\left(t\right)=\left\{ u_{R}\left(t\right):L\left(t\right)u_{R}\left(t\right)\leq S\left(t,\dot x_H\right)\right\} \label{eq:simplified safe control},
\end{equation}
where 
\begin{subequations}
\begin{align}
L\left(t\right) =& \frac{\partial\phi}{\partial x_{R}}h\label{eq: l(t)},\\
S\left(t,\dot x_H\right) =& 
\begin{cases}
\begin{array}{c}
-\eta_{R}-\frac{\partial\phi}{\partial x_{H}}\dot{x}_{H}-\frac{\partial\phi}{\partial x_{R}}f\\
\infty\\
\end{array}
\begin{array}{c}
\phi\geq 0\\
\phi<0\\
\end{array}
\end{cases}.\label{eq: s(t)}
\end{align}
\end{subequations}

The vector $L(t)$ points to the ``safe'' direction, while the scalar $S(t,\dot x_H)$ indicates the allowed range of safe control input. The scalar $S(t,\dot x_H)$ consists of three parts: a margin $-\eta_R$, a term to compensate human motion $-\frac{\partial\phi}{\partial x_{H}}\dot{x}_{H}$ and a term to compensate the inertia of the robot itself $-\frac{\partial\phi}{\partial x_{R}}f$. In the following discussion when there is no ambiguity, $S(t,\dot x_H)$ denotes the value in the case $\phi\geq 0$ only. Under different assumptions of the human behavior, $S(t)$ varies. The sets of safe control correspond to $R_S^i$ are $U_S^i = \{u_R(t):L(t)u_R(t)\leq S^i(t)\}$ where
\begin{subequations}
\begin{align}
S^1\left(t\right) =& \max_{\dot x_H} S(t,\dot x_H) \label{eq: us1},\\
S^2\left(t\right) =& S\left(t,\hat {\dot x}_H\right), \label{eq: us2}\\
S^3\left(t\right) =& \min_{\dot x_H\in \dot\Gamma_H} S\left(t,\dot x_H\right), \label{eq: us3}
\end{align}
\end{subequations}
where $\hat{\dot x}_H$ is the velocity vector that moves the current configuration $x_H$ of human to $\hat x_H$ and $\dot\Gamma_H$ is the set of velocity vectors that move $x_H$ to $\Gamma_H$. Computationally, $\hat{\dot x}_H(k)=\frac{\hat x_H(k+1)-\hat x_H(k)}{t_s}$. 
Obviously $U_S^3\subset U_S^2\subset U_S^1$. When the uncertainties in the estimation of $\hat{\dot x}_H$ reduce, $U_S^3$ converges to $U_S^2$. For robust safety, $U_S^3$ is the best design.

The difference between $R_S$ and $U_S$ is that $R_S$ is static as it is on the state space, while $U_S$ is dynamic as it concerns with the ``movements''.
Due to introduction of the safety index, the non-convex state space constraint $R_S$ is transformed to a convex state space constraint $U_S$. Since $\Omega$ is convex, the problem (\ref{eq: chap4 optimization for local planning}) is transformed to a convex optimization,
\begin{subequations}\label{eq: convex optimization for local planning}
\begin{align}
u_R^* = ~& \arg\min_{u_R} \|u_R-u_R^o\|^2_Q,\\
s.t.~& u_R\in\Omega\cap U_S^3\label{eq: convex optimization for local planning constraint}.
\end{align}
\end{subequations}
The robot can either sent the modified trajectory or the control input to the robot hardware for execution. Applications of the method can be found in \cite{liu2014control,liu2015safe}.

\section{Integration and Evaluation\label{sec: T4}}


The proposed SERoCS is evaluated in a human-robot collaborative desktop assembly task as illustrated in \cref{fig: future factory}. \Cref{sec: validation of T1} shows the experiment result of human motion prediction in T1. \Cref{sec: validation of T2} validates the grasping skills learned in T2. \Cref{sec: sc experiment,sec: validation all} present integrated experiments. In \Cref{sec: sc experiment}, the robot is in idle, while it collaborates with the human worker in \cref{sec: validation all}.

\subsection{Experiment Setup}
\begin{figure}[t]
\begin{center}
\includegraphics[width = 0.8\linewidth]{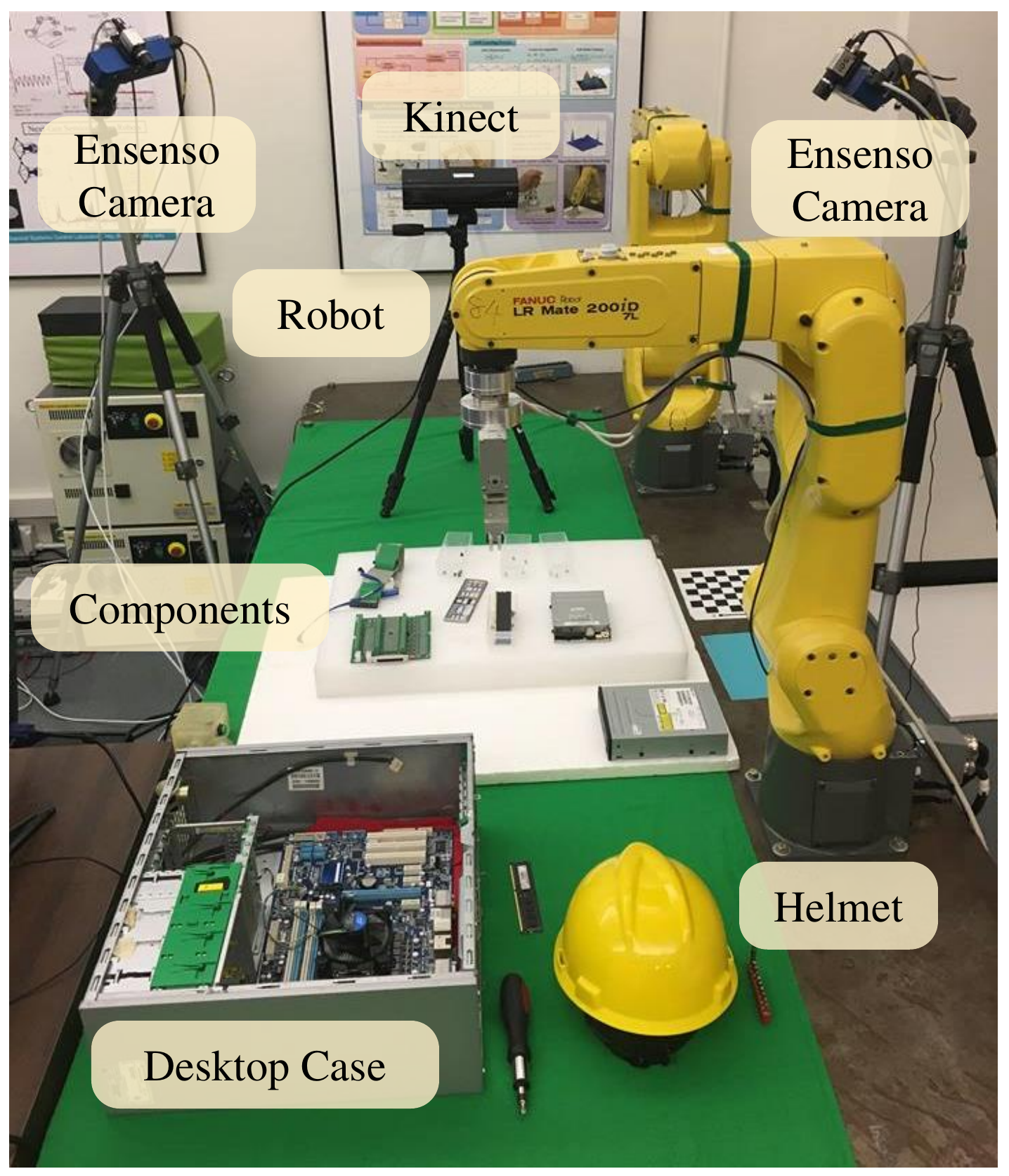}
\caption{Experiment Setup.}
\label{fig: setup}
\end{center}
\end{figure}

The experiment platform is shown in \cref{fig: setup}. The robot manipulator is FANUC LR Mate 200iD/7L. There are one Kinect sensor to monitor the dynamic environment and two Ensenso cameras to capture the static components placed in the workspace. For simplicity, the desktop case and the helmet are attached markers so that Kinect can directly retrieve their location in real time. All the algorithms are implemented in MATLAB on a Windows desktop with an Intel Core i5 CPU and 16GB RAM. The robot
controller is deployed on a Simulink RealTime target.

\subsection{Validation of the Environment Monitoring\label{sec: validation of T1}}
In order to verify the proposed environment monitoring approach in \cref{sec: T1}, a series of experiments are conducted to complete a task with different plans. Human and robot collaborate to assemble a desktop. Human has two plans in mind: inserting the RAMs in the motherboard first and then assembling the disk to the desktop case, or assembling the disk to the desktop case first and then inserting the RAM to the motherboard. The robot may collaborate with the human by handing the other RAM to the human if the human is assembling a RAM, or bringing the screwdriver to the human if the human is assembling the disk. To simply test the performance of T1, the robot is not plugged in in this experiment.

Human worker's trajectories of right wrist joint are retrieved automatically from the Kinect sensor, the rate of which is about 15 frames per second. By applying a simple averaging filter, the smoothed trajectories are generated, which are further used to train the plan recognition classifier and the motion prediction models, the number of neurons of the hidden layer is set to be 40. 50 trajectories for each plan is collected, among which 5 are randomly chosen to be in the test set. 



\begin{figure}[t]
{
%
%
%
\begin{tikzpicture}

\begin{axis}[%
name = ax1,
width=1.2in,
height=1.3in,
at={(0cm,0cm)},
scale only axis,
xmin=0,
xmax=100,
xlabel={Iterations},
ymin=-350,
ymax=6000,
ylabel={Loss},
yticklabel style={rotate=90},
axis background/.style={fill=white},
font = \footnotesize,
]
\addplot [color=mycolor2, line width=1.5pt, forget plot] 
 table[row sep=crcr]{%
1	614.610534667969\\
2	5632.59619140625\\
3	4496.53076171875\\
4	2578.416015625\\
5	305.584899902344\\
6	2.03061413764954\\
7	606.508544921875\\
8	382.819671630859\\
9	50.3890724182129\\
10	11.7716245651245\\
11	2.39829540252686\\
12	0.760988473892212\\
13	0.539470434188843\\
14	8.10189247131348\\
15	12.0173826217651\\
16	20.3759613037109\\
17	15.8303775787354\\
18	15.813193321228\\
19	22.0257797241211\\
20	21.9405746459961\\
21	17.829418182373\\
22	14.6533880233765\\
23	12.5280885696411\\
24	21.0777454376221\\
25	18.4825992584229\\
26	18.5369892120361\\
27	9.98285865783691\\
28	6.02986335754395\\
29	7.91874885559082\\
30	2.12479400634766\\
31	1.47274076938629\\
32	1.69227004051209\\
33	0.0249317940324545\\
34	0.977720379829407\\
35	0.597698390483856\\
36	0.723062932491302\\
37	0.424267798662186\\
38	0.193426907062531\\
39	0.661928474903107\\
40	0.403519302606583\\
41	0.88819819688797\\
42	1.734330534935\\
43	0.15135945379734\\
44	1.40489459037781\\
45	1.04557573795319\\
46	0.241058245301247\\
47	1.18875634670258\\
48	0.980545818805695\\
49	1.10321366786957\\
50	0.646167159080505\\
51	1.17019295692444\\
52	1.79356575012207\\
53	0.115876704454422\\
54	0.853349030017853\\
55	0.672412812709808\\
56	0.138056263327599\\
57	0.52553403377533\\
58	0.0808399543166161\\
59	0.53196257352829\\
60	0.207322001457214\\
61	0.197327971458435\\
62	0.6676384806633\\
63	0\\
64	0.104774862527847\\
65	0.251513928174973\\
66	0.17339463531971\\
67	0.151372045278549\\
68	0\\
69	0.234845951199532\\
70	0.184589982032776\\
71	0.00315754022449255\\
72	0.215696409344673\\
73	0\\
74	0.00311286048963666\\
75	0.111306570470333\\
76	0.238734856247902\\
77	0.0793190747499466\\
78	0.0137044806033373\\
79	0.0650107860565186\\
80	0.232582435011864\\
81	0\\
82	0.0881247594952583\\
83	0\\
84	0\\
85	0.0571066774427891\\
86	0.281005680561066\\
87	0.0682730749249458\\
88	0.0325116440653801\\
89	0.030212352052331\\
90	0.238193243741989\\
91	0\\
92	0.0547516718506813\\
93	0\\
94	0\\
95	0.0380343720316887\\
96	0.252332508563995\\
97	0.058990441262722\\
98	0.026467414572835\\
99	0.0170135106891394\\
100	0.221244513988495\\
};
\coordinate (c1) at (axis cs:30,-100);
  \coordinate (c2) at (axis cs:30,500);
  \draw [mycolor1, thick](c1) rectangle (axis cs:100,500);
\end{axis}
\begin{axis}[
	name = ax2,
	width=1.2in,
	height=1.3in,
	scale only axis,
	xmin = 30, xmax = 100,
	ymin = -.1, ymax = 2.5,
	xlabel = {Iterations},
	ylabel = {Loss},
	yticklabel style={rotate=90},
	at={($(ax1.south east) + (1cm, 0)$)},
	font = \footnotesize,
]
	\addplot [color=mycolor2, line width=1.5pt, forget plot] 
 table[row sep=crcr] {%
30	2.12479400634766\\
31	1.47274076938629\\
32	1.69227004051209\\
33	0.0249317940324545\\
34	0.977720379829407\\
35	0.597698390483856\\
36	0.723062932491302\\
37	0.424267798662186\\
38	0.193426907062531\\
39	0.661928474903107\\
40	0.403519302606583\\
41	0.88819819688797\\
42	1.734330534935\\
43	0.15135945379734\\
44	1.40489459037781\\
45	1.04557573795319\\
46	0.241058245301247\\
47	1.18875634670258\\
48	0.980545818805695\\
49	1.10321366786957\\
50	0.646167159080505\\
51	1.17019295692444\\
52	1.79356575012207\\
53	0.115876704454422\\
54	0.853349030017853\\
55	0.672412812709808\\
56	0.138056263327599\\
57	0.52553403377533\\
58	0.0808399543166161\\
59	0.53196257352829\\
60	0.207322001457214\\
61	0.197327971458435\\
62	0.6676384806633\\
63	0\\
64	0.104774862527847\\
65	0.251513928174973\\
66	0.17339463531971\\
67	0.151372045278549\\
68	0\\
69	0.234845951199532\\
70	0.184589982032776\\
71	0.00315754022449255\\
72	0.215696409344673\\
73	0\\
74	0.00311286048963666\\
75	0.111306570470333\\
76	0.238734856247902\\
77	0.0793190747499466\\
78	0.0137044806033373\\
79	0.0650107860565186\\
80	0.232582435011864\\
81	0\\
82	0.0881247594952583\\
83	0\\
84	0\\
85	0.0571066774427891\\
86	0.281005680561066\\
87	0.0682730749249458\\
88	0.0325116440653801\\
89	0.030212352052331\\
90	0.238193243741989\\
91	0\\
92	0.0547516718506813\\
93	0\\
94	0\\
95	0.0380343720316887\\
96	0.252332508563995\\
97	0.058990441262722\\
98	0.026467414572835\\
99	0.0170135106891394\\
100	0.221244513988495\\
};
\end{axis}

\draw [dashed] (c1) -- (ax2.south west);
\draw [dashed] (c2) -- (ax2.north west);

\end{tikzpicture}%
}
\caption{Learning curve for the trajectory based plan recognition classifier.}
\label{fig: learning_plan}
\end{figure}
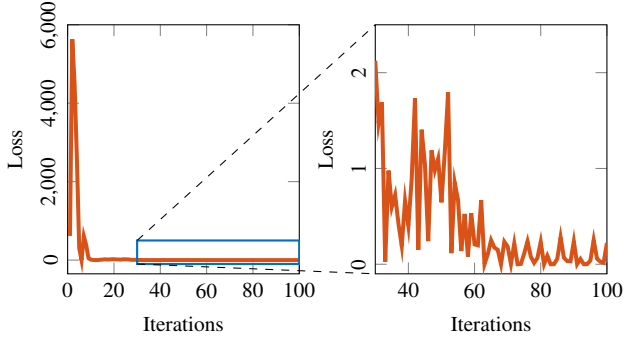

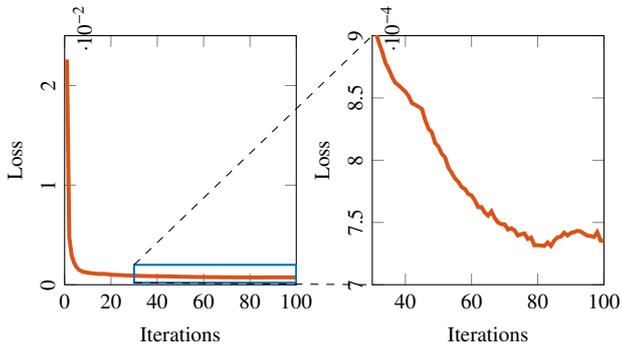
\begin{figure}[t]
{
%
%
%
\begin{tikzpicture}

\begin{axis}[%
name = ax1,
width=1.2in,
height=1.3in,
at={(0cm,0cm)},
scale only axis,
xmin=0,
xmax=100,
xlabel={Iterations},
ymin=0,
ymax=0.025,
ylabel={Loss},
yticklabel style={rotate=90},
axis background/.style={fill=white},
font = \footnotesize,
]
\addplot [color=mycolor2, line width=1.5pt, forget plot]
  table[row sep=crcr]{%
1	0.0226084537031477\\
2	0.00476348158965809\\
3	0.00298748551111934\\
4	0.00224470982547814\\
5	0.00181363802355722\\
6	0.00154205973775541\\
7	0.00139060283623613\\
8	0.0013029489633456\\
9	0.0012492554004978\\
10	0.00120900259397148\\
11	0.00117483430374534\\
12	0.00114596913737203\\
13	0.00111911983524365\\
14	0.00110103092645786\\
15	0.00108380524954899\\
16	0.00109296223515914\\
17	0.00109335579547873\\
18	0.00106404882735684\\
19	0.00103967226315879\\
20	0.00101764124623262\\
21	0.00100043190140132\\
22	0.000988828487886832\\
23	0.000980729319770676\\
24	0.000965419017740778\\
25	0.000952810339601869\\
26	0.000941988259173224\\
27	0.000933774019672183\\
28	0.000925775828650908\\
29	0.000916531485900394\\
30	0.000911548306746233\\
31	0.000903246426290076\\
32	0.000894586510630363\\
33	0.000887163672927305\\
34	0.000878115736507538\\
35	0.000873084799494302\\
36	0.000866828853952837\\
37	0.000862022115519591\\
38	0.000860142313917375\\
39	0.000857495349201519\\
40	0.000855037233811775\\
41	0.00085145107778346\\
42	0.000845775444616217\\
43	0.00084446585199413\\
44	0.000842943927957296\\
45	0.000841166998324694\\
46	0.000832093885260651\\
47	0.000825296667507702\\
48	0.000822471032137631\\
49	0.00081371127681616\\
50	0.000811009885444167\\
51	0.000805190633111884\\
52	0.000802250560864581\\
53	0.000794010389410957\\
54	0.000789903714560039\\
55	0.000785428144170723\\
56	0.000782666183737344\\
57	0.000778469980183667\\
58	0.000776877532065422\\
59	0.000772980500907481\\
60	0.000771722487210838\\
61	0.000767646744706373\\
62	0.000762308421465186\\
63	0.000762384106650599\\
64	0.000758048632606972\\
65	0.000755493972126144\\
66	0.000758925594932418\\
67	0.000753636121408638\\
68	0.000750048691059581\\
69	0.000748644690961447\\
70	0.000748522083185206\\
71	0.000744357445487621\\
72	0.000745378169783327\\
73	0.000743668084360806\\
74	0.000739554379820877\\
75	0.000740604836444342\\
76	0.000741183410546888\\
77	0.000736789507653978\\
78	0.000737905972028336\\
79	0.000731680084664743\\
80	0.000731728466105247\\
81	0.000731522335533437\\
82	0.000731112821747632\\
83	0.000733933019585746\\
84	0.000731171243825994\\
85	0.00073557831508823\\
86	0.000737991533813521\\
87	0.000738693401399255\\
88	0.000741784949818126\\
89	0.000739551132561318\\
90	0.000741544536503407\\
91	0.000742262084407125\\
92	0.000743311374161583\\
93	0.00074278683081107\\
94	0.000741008501007636\\
95	0.000739465340012012\\
96	0.000739157598958114\\
97	0.000738333330482097\\
98	0.000742208669482425\\
99	0.000735462990582198\\
100	0.000735115923298782\\
};
\coordinate (c1) at (axis cs:30,.0002);
  \coordinate (c2) at (axis cs:30,.002);
  \draw [mycolor1, thick](c1) rectangle (axis cs:100,.002);
\end{axis}

\begin{axis}[
	name = ax2,
	width=1.2in,
	height=1.3in,
	scale only axis,
	xmin = 30, xmax = 100,
	ymin = 0.0007, ymax = .0009,
	xlabel = {Iterations},
	ylabel = {Loss},
	yticklabel style={rotate=90},
	at={($(ax1.south east) + (1cm, 0)$)},
	font = \footnotesize,
]
\addplot [color=mycolor2, line width=1.5pt, forget plot] 
 table[row sep=crcr] {%
 30	0.000911548306746233\\
31	0.000903246426290076\\
32	0.000894586510630363\\
33	0.000887163672927305\\
34	0.000878115736507538\\
35	0.000873084799494302\\
36	0.000866828853952837\\
37	0.000862022115519591\\
38	0.000860142313917375\\
39	0.000857495349201519\\
40	0.000855037233811775\\
41	0.00085145107778346\\
42	0.000845775444616217\\
43	0.00084446585199413\\
44	0.000842943927957296\\
45	0.000841166998324694\\
46	0.000832093885260651\\
47	0.000825296667507702\\
48	0.000822471032137631\\
49	0.00081371127681616\\
50	0.000811009885444167\\
51	0.000805190633111884\\
52	0.000802250560864581\\
53	0.000794010389410957\\
54	0.000789903714560039\\
55	0.000785428144170723\\
56	0.000782666183737344\\
57	0.000778469980183667\\
58	0.000776877532065422\\
59	0.000772980500907481\\
60	0.000771722487210838\\
61	0.000767646744706373\\
62	0.000762308421465186\\
63	0.000762384106650599\\
64	0.000758048632606972\\
65	0.000755493972126144\\
66	0.000758925594932418\\
67	0.000753636121408638\\
68	0.000750048691059581\\
69	0.000748644690961447\\
70	0.000748522083185206\\
71	0.000744357445487621\\
72	0.000745378169783327\\
73	0.000743668084360806\\
74	0.000739554379820877\\
75	0.000740604836444342\\
76	0.000741183410546888\\
77	0.000736789507653978\\
78	0.000737905972028336\\
79	0.000731680084664743\\
80	0.000731728466105247\\
81	0.000731522335533437\\
82	0.000731112821747632\\
83	0.000733933019585746\\
84	0.000731171243825994\\
85	0.00073557831508823\\
86	0.000737991533813521\\
87	0.000738693401399255\\
88	0.000741784949818126\\
89	0.000739551132561318\\
90	0.000741544536503407\\
91	0.000742262084407125\\
92	0.000743311374161583\\
93	0.00074278683081107\\
94	0.000741008501007636\\
95	0.000739465340012012\\
96	0.000739157598958114\\
97	0.000738333330482097\\
98	0.000742208669482425\\
99	0.000735462990582198\\
100	0.000735115923298782\\
};
\end{axis}
\draw [dashed] (c1) -- (ax2.south west);
\draw [dashed] (c2) -- (ax2.north west);
\end{tikzpicture}%
}
\caption{Learning curve for the neural network of the motion prediction.}
\label{fig: motion_pred}
\end{figure}



\begin{figure}[!t]
        \centering
        \subfloat[]{\includegraphics[width=0.24\textwidth]{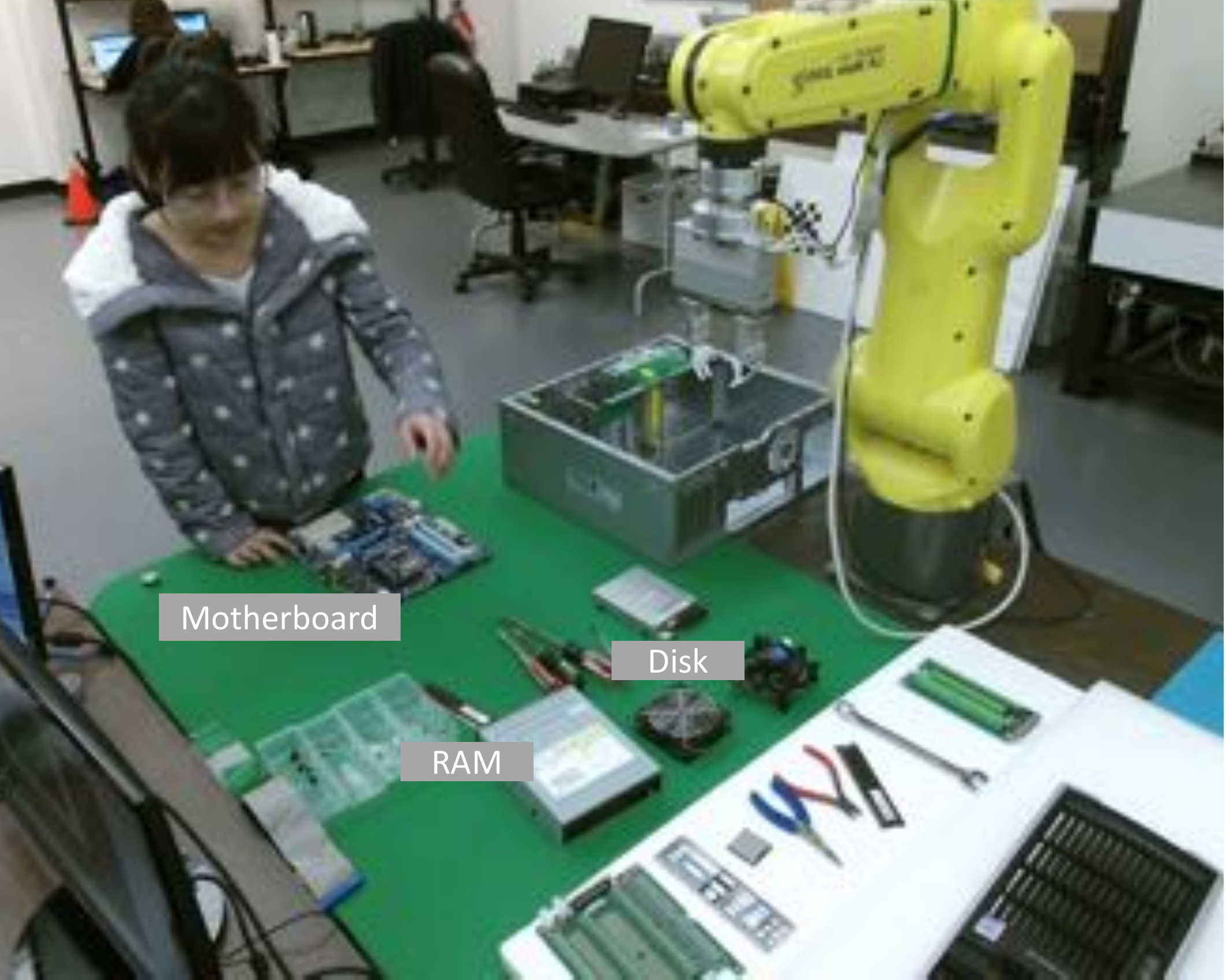}%
        		  }
        \hfil
        \subfloat[]{\includegraphics[width=0.24\textwidth]{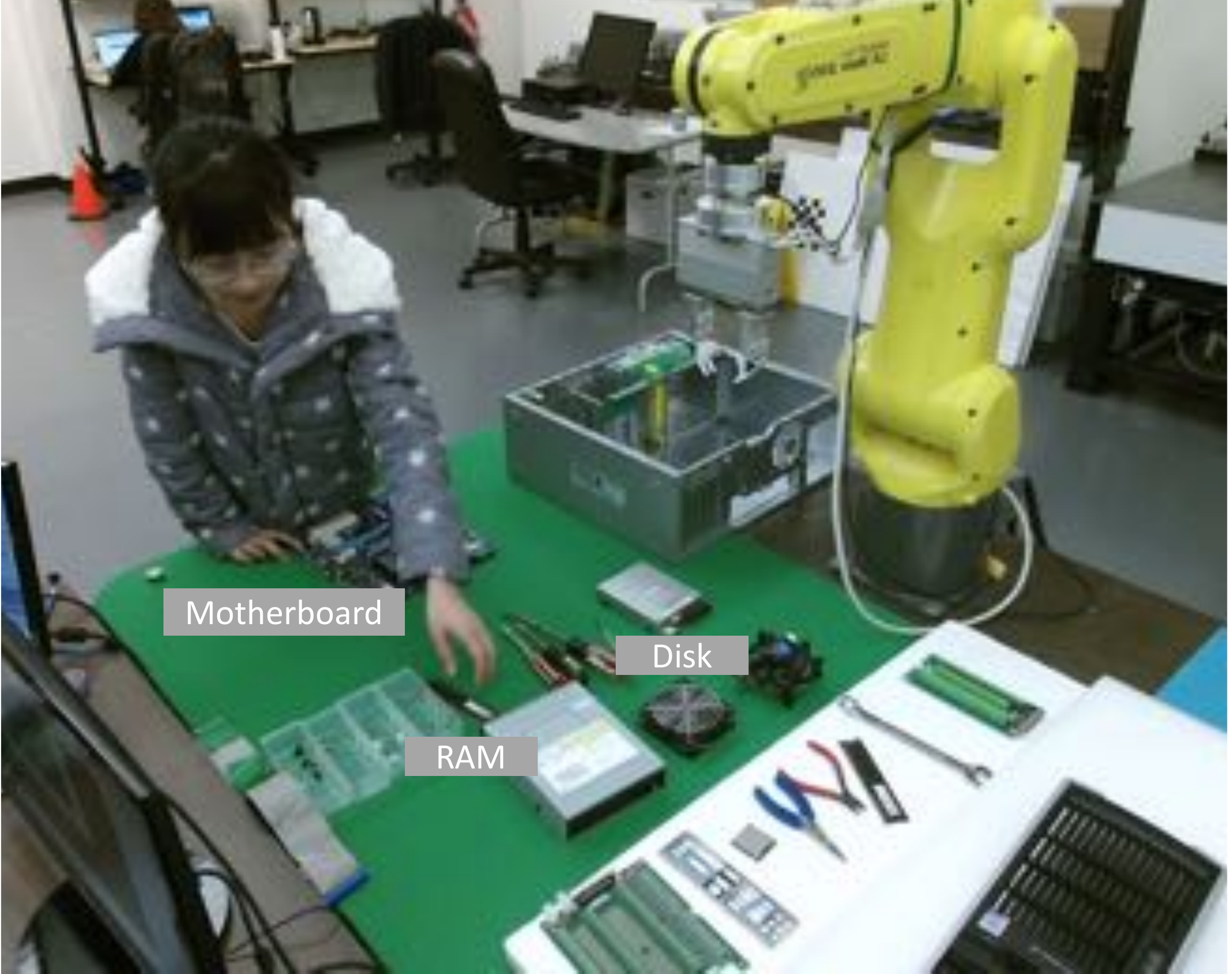}%
        		  }
        \\
        \subfloat[]{\includegraphics[width=0.24\textwidth]{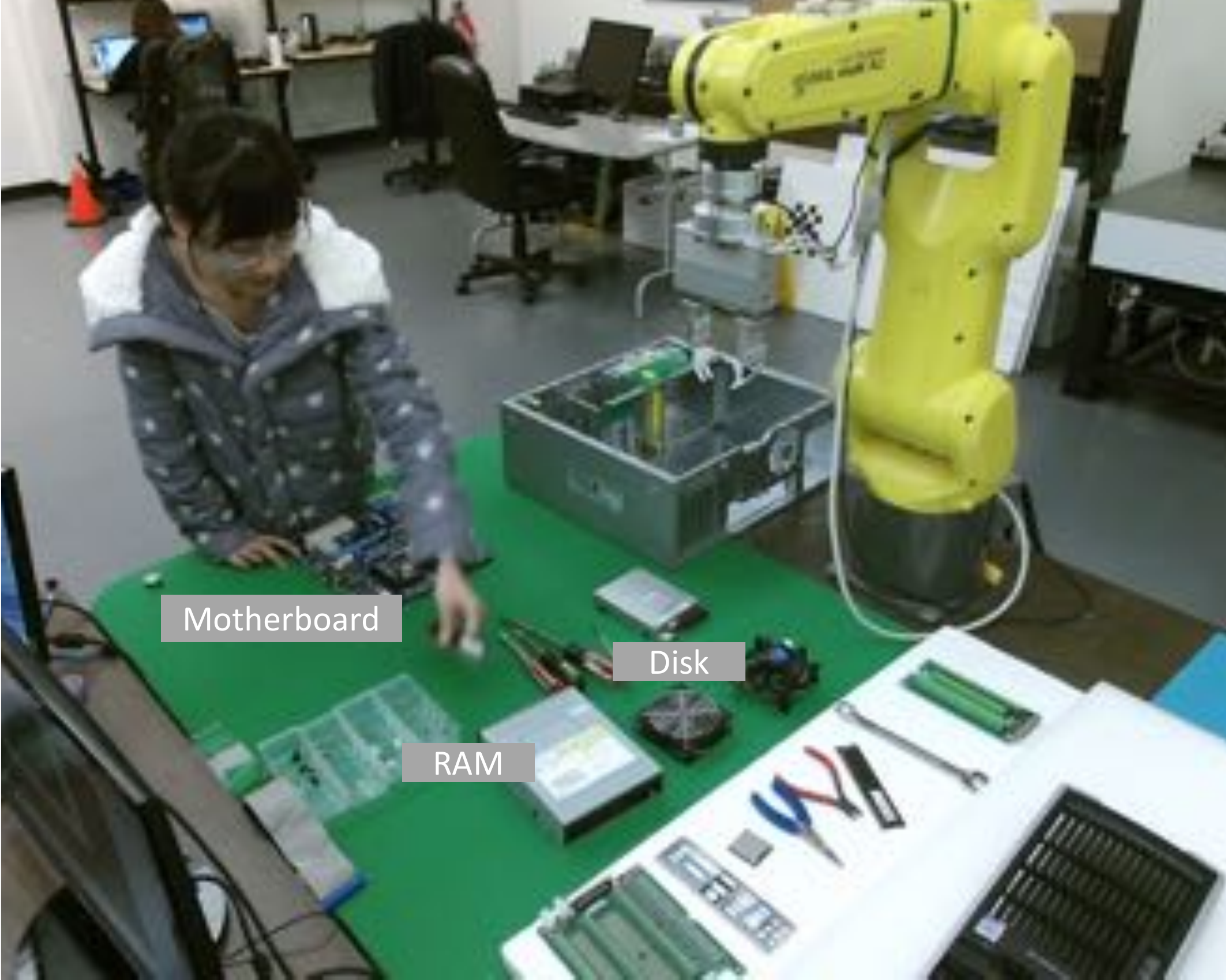}%
               }
        \hfil
        \subfloat[]{\includegraphics[width=0.24\textwidth]{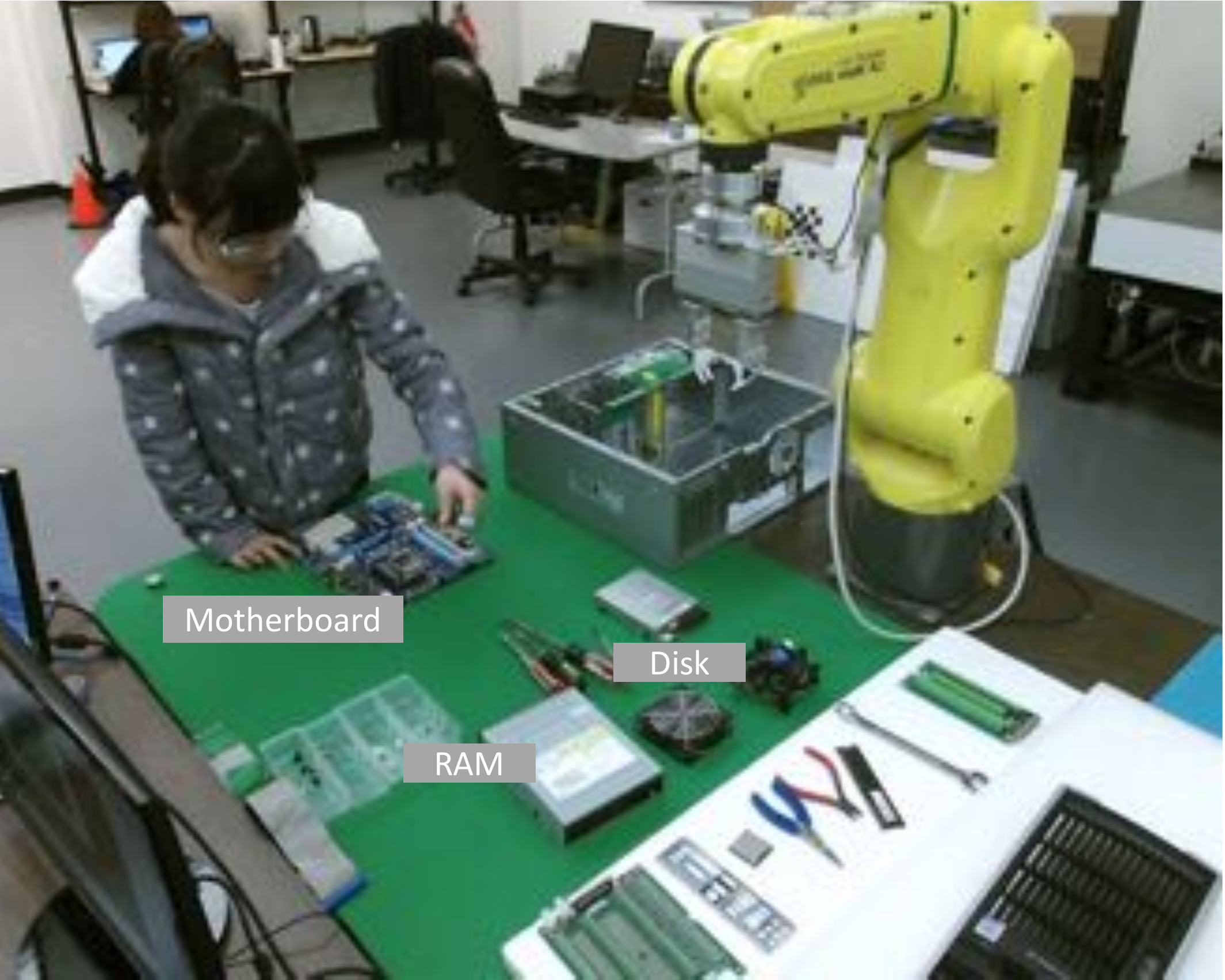}%
                }
        \caption{The human subject conducts the first plan. (a) Reaching for the RAM. (b) Getting closer to the RAM. (c) Holding the RAM. (d) Assembling the RAM.}
        \label{fig: feature_compare 1}
\end{figure}

\begin{figure}[!t]
        \centering
        \subfloat[]{\includegraphics[width=0.24\textwidth]{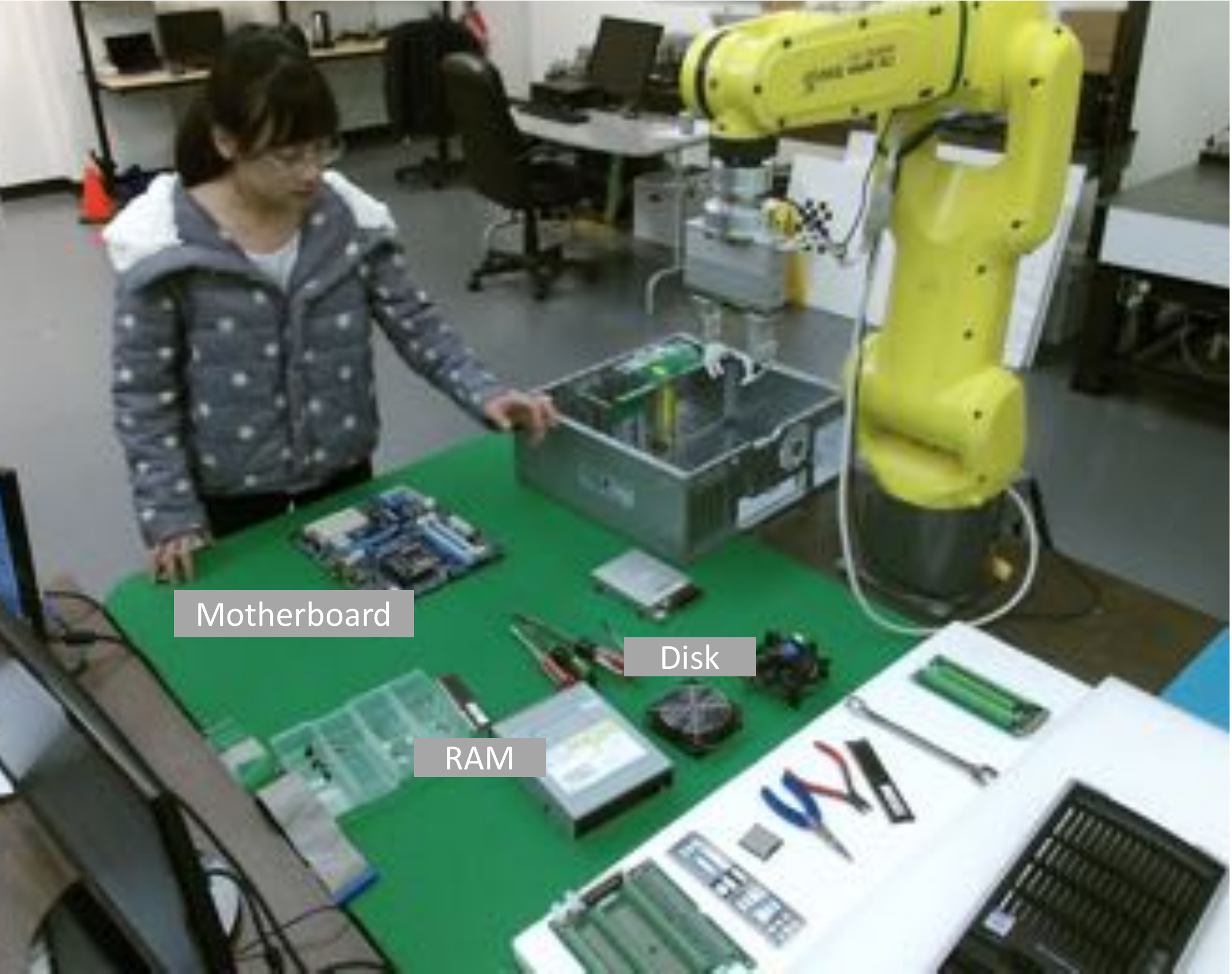}%
        		  }
        \hfil
        \subfloat[]{\includegraphics[width=0.24\textwidth]{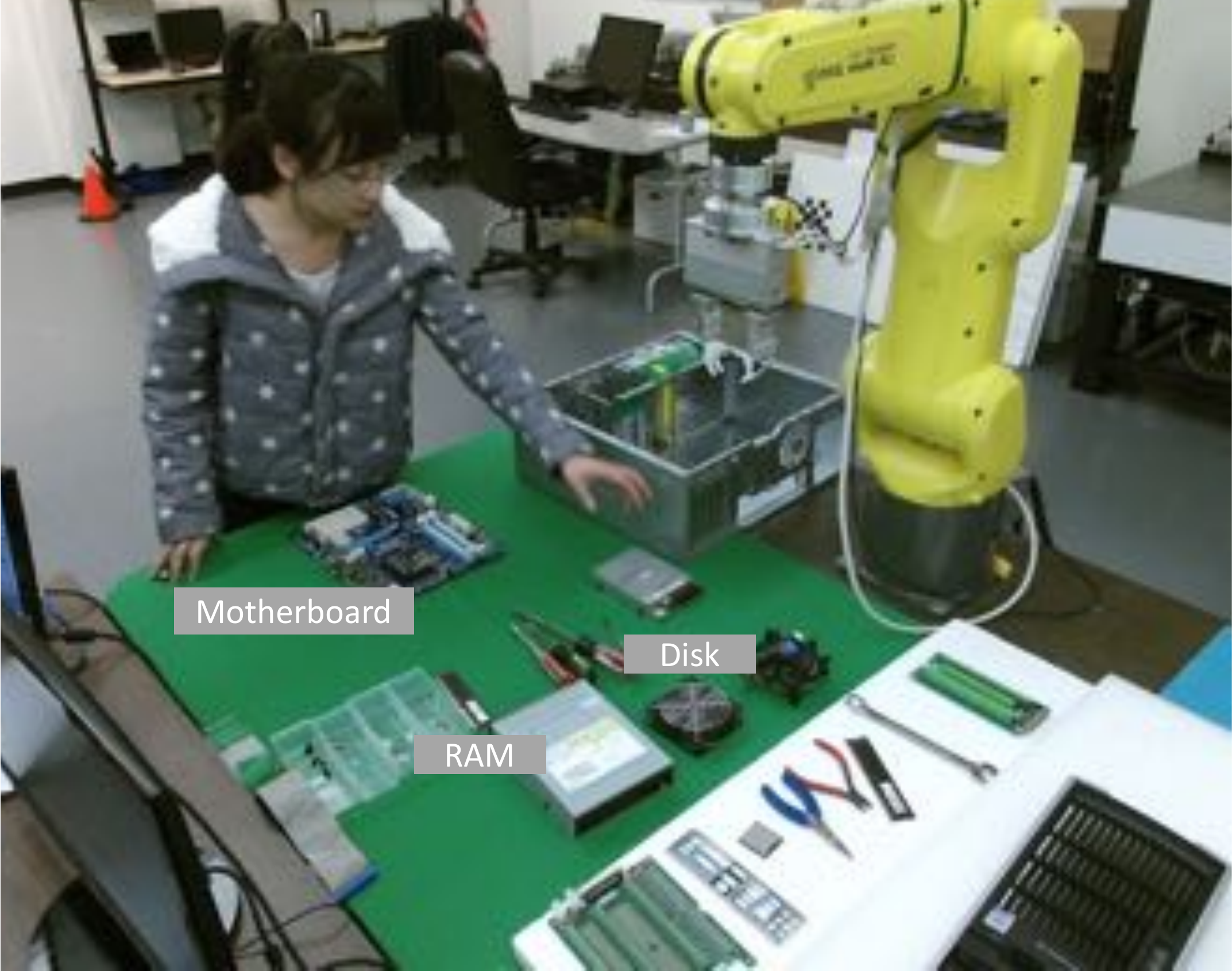}%
        		  }
        \\
        \subfloat[]{\includegraphics[width=0.24\textwidth]{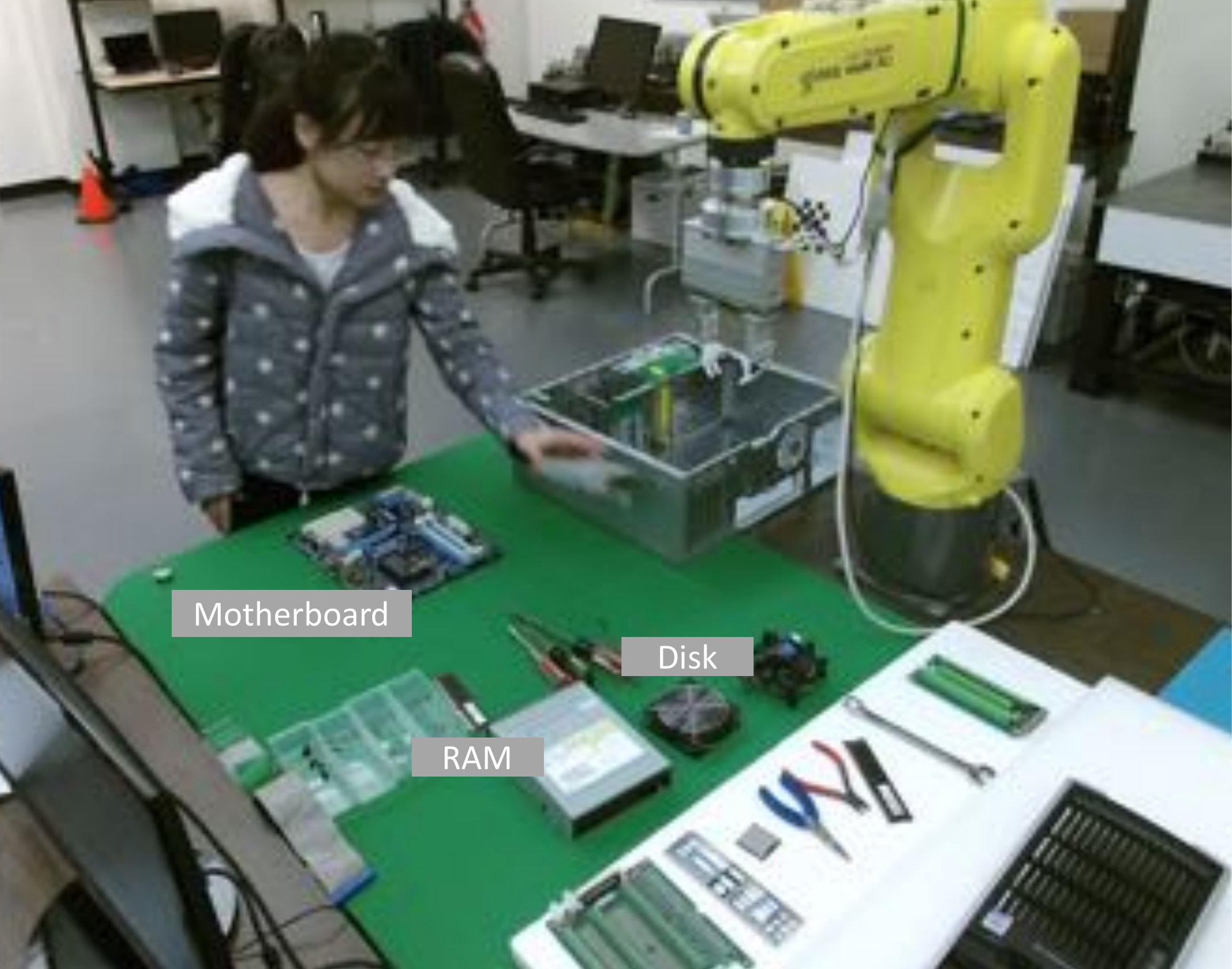}%
               }
        \hfil
        \subfloat[]{\includegraphics[width=0.24\textwidth]{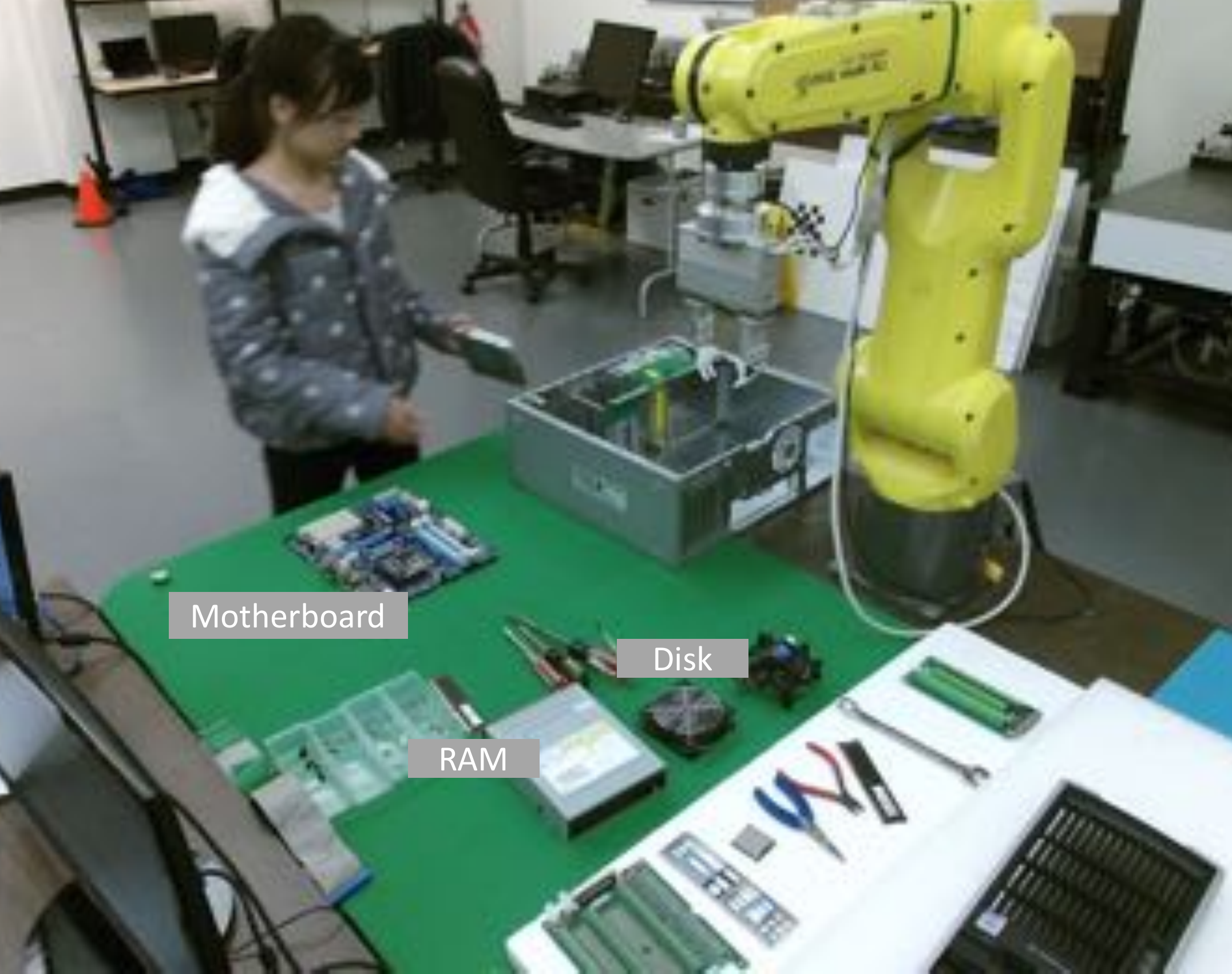}%
                }
        \caption{The human subject conducts the second plan. (a) Reaching for the disk. (b) Getting closer to the disk. (c) Holding the disk. (d) Assembling the disk.}
        \label{fig: feature_compare 2}
\end{figure}

The learning curve for the trajectory based plan recognition is shown in \cref{fig: learning_plan}, and the learning curve for the motion prediction model is shown in \cref{fig: motion_pred}, which indicate a quick convergence in both training process. The trajectory-based plan recognition classifier can get right predictions when the test trajectory is about $10\%$ of the whole process. From the model trained by the neural networks, we get \SI{0.015}{\meter} mean squared error, which is satisfactory.

\begin{figure}[t]
{
%
%
%
\begin{tikzpicture}

\begin{axis}[%
width=2.8in,
height=2.25in,
at={(0cm,0cm)},
scale only axis,
plot box ratio=3.41 3.273 1,
xmin=-0.00665089633468346,
xmax=0.492181543135919,
tick align=outside,
xlabel={$x$ [\si{\meter}]},
ymin=-0.00058216536372957,
ymax=0.478176281360589,
ylabel={$y$ [\si{\meter}]},
zmin=0.135415218997953,
zmax=0.281690507916215,
zlabel={$z$ [\si{\meter}]},
view={45.7}{44.4},
axis background/.style={fill=white},
axis x line*=bottom,
axis y line*=left,
axis z line*=left,
font = \footnotesize,
legend style={at={(0,0.8)}, anchor=south west, legend cell align=left, align=left, draw=white!15!black}
]
\addplot3 [color=mycolor3, line width=1.1pt, mark=asterisk, mark options={solid, mycolor3}]
 table[row sep=crcr] {%
0.374313749253298	0.0495463186937328	0.135415218997953\\
0.364152112815107	0.0568524270161124	0.156864094602857\\
0.344759569646596	0.0622453862818293	0.169211185605004\\
0.320181505246383	0.0629776712814236	0.174727286775321\\
0.301340637204834	0.0675693812091734	0.19380894266587\\
0.278247974425706	0.0639162636218588	0.203745479577002\\
0.249100571842269	0.0530929799226166	0.215533944655196\\
0.23117293549372	0.0475981698645996	0.227038594721607\\
0.199111144391008	0.0515330667769294	0.230064536900004\\
0.169400474929292	0.0581063538566542	0.23844777998988\\
0.154713420579545	0.059132085668556	0.250640883564026\\
0.117861310748681	0.05017271709032	0.246874684304345\\
0.0742579701585863	0.0358725184832551	0.237216169091551\\
0.037731868150222	0.0204152184974612	0.220216383825177\\
0.00867517364706116	0.00954911601466069	0.200119305003029\\
-0.00052383491537894	0.00600119634131113	0.189021896999276\\
-0.00392622018296174	0.00223168455301337	0.176020347878223\\
-0.00665089633468346	-0.00058216536372957	0.162864069682253\\
0.00307755171999569	0.00245114618423781	0.160817340502066\\
0.0140225174060882	0.00651248501987511	0.161964609607233\\
0.0154533285232064	0.00797355031595515	0.161169958432193\\
0.0158962058680444	0.00919853329181075	0.160737400774861\\
0.0169030175411173	0.0106575040613573	0.160051360179089\\
0.0198896219023081	0.0131785538204517	0.159976755138562\\
0.0238630585361385	0.0154062995785027	0.160096703451718\\
0.0346520778158683	0.0173439365170998	0.16245889976782\\
0.0578599306662941	0.0241687454191832	0.177821282864064\\
0.0845923774122737	0.036056044190661	0.19791283329702\\
0.108198958535389	0.0477058030745174	0.21290678300401\\
0.134185651935782	0.0580828467221405	0.235118182408078\\
0.161347046742992	0.067477946972033	0.258380408236473\\
0.221726601308793	0.0904321466831424	0.264959456456951\\
0.304419380266029	0.121936812281106	0.2537908341905\\
0.377233789453985	0.145442094601359	0.265654064552067\\
0.427005643366868	0.165933787715704	0.281690507916215\\
0.449586550300474	0.187698338080659	0.273930927313096\\
0.475151644092896	0.216551939804281	0.266163926149717\\
0.491733009320918	0.246273104596748	0.252069092776592\\
0.492181543135919	0.294637940300042	0.234599927138153\\
0.463187180734581	0.356991023816939	0.226052429415791\\
0.442540798634769	0.386378791965133	0.228793392365467\\
0.406263481035078	0.416112033239761	0.208897320715798\\
0.369763460434734	0.446204525130772	0.185921116709041\\
0.374314242237095	0.454532368242316	0.187343638293859\\
0.363729449236077	0.461270769324157	0.185004954543103\\
0.353077990981497	0.463832789162728	0.180859982646777\\
0.337348004250994	0.465537623639048	0.17771766211143\\
0.327857833245783	0.467883370221488	0.178557294401513\\
0.334767770315313	0.470996048899561	0.183372429263067\\
0.338893078525531	0.472862064365585	0.183732235880205\\
0.342284557390049	0.474084927979927	0.18322575287017\\
0.343410966997249	0.47748236090451	0.185262813368752\\
0.333299319960378	0.477120803176134	0.183559483224986\\
0.319484457466029	0.472528995726928	0.176853323163117\\
0.314615019320071	0.470705503956003	0.16695962616102\\
0.294126122142803	0.472697502614135	0.169755334766423\\
0.303282582912554	0.476845016504129	0.183988920739867\\
0.327987442899	0.478176281360589	0.188103757712232\\
0.327222884041548	0.476891884116134	0.187825064413226\\
0.326602395002254	0.474782045998451	0.187687994579885\\
0.317416603374967	0.468973493346518	0.177932547589077\\
0.29715018502932	0.4649270518584	0.172441761767826\\
0.286788338495592	0.465426288152032	0.174369235171308\\
0.294451845112727	0.467071177656414	0.17664580806615\\
0.295091623608001	0.467825814175336	0.17725850157558\\
0.288780967661733	0.468472711199938	0.174489601100184\\
0.288971261612273	0.467649133922084	0.171014842706736\\
0.291103913530958	0.465033841414331	0.168708696602011\\
0.269961186058609	0.458835778722225	0.164122203548892\\
0.275571036163368	0.457177869491036	0.16141286502626\\
0.293556291322032	0.460756335622081	0.167525874038669\\
0.287213664953805	0.461415787786915	0.170629330991852\\
};
 \addlegendentry{actual trajectory}

\addplot3 [color=mycolor5, line width=1.1pt, mark=o, mark options={solid, mycolor5}]
 table[row sep=crcr] {%
0.199111148715019	0.0515330657362938	0.230064541101456\\
0.182662025094032	0.0528463497757912	0.22579887509346\\
0.171258181333542	0.0476835779845715	0.225725412368774\\
0.163138955831528	0.0460424534976482	0.223589420318604\\
};
 \addlegendentry{predicted trajectory}

\addplot3 [color=mycolor5, line width=1.1pt, mark=o, mark options={solid, mycolor5}]
 table[row sep=crcr] {%
0.0377318672835827	0.0204152185469866	0.220216378569603\\
0.0244324058294296	0.0127568896859884	0.216780483722687\\
0.0210430100560188	0.00663892552256584	0.214604943990707\\
0.0107168927788734	0.00748161599040031	0.20533487200737\\
};
 \addplot3 [color=mycolor5, line width=1.1pt, mark=o, mark options={solid, mycolor5}]
 table[row sep=crcr] {%
0.084592379629612	0.0360560454428196	0.197912827134132\\
0.0976691693067551	0.0511934012174606	0.198826730251312\\
0.110062576830387	0.0570559166371822	0.202987521886826\\
0.120039761066437	0.0641614645719528	0.204068273305893\\
};
 \addplot3 [color=mycolor5, line width=1.1pt, mark=o, mark options={solid, mycolor5}]
 table[row sep=crcr] {%
0.492181539535522	0.294637948274612	0.234599933028221\\
0.482968956232071	0.299464553594589	0.233728319406509\\
0.481824636459351	0.313164830207825	0.235398471355438\\
0.475398719310761	0.327086687088013	0.233842104673386\\
};
 \end{axis}
\end{tikzpicture}%
}
\caption{Motion prediction result for plan two corresponding to \cref{fig: feature_compare 2}.}
\label{fig: corr_motion 2}
\end{figure}
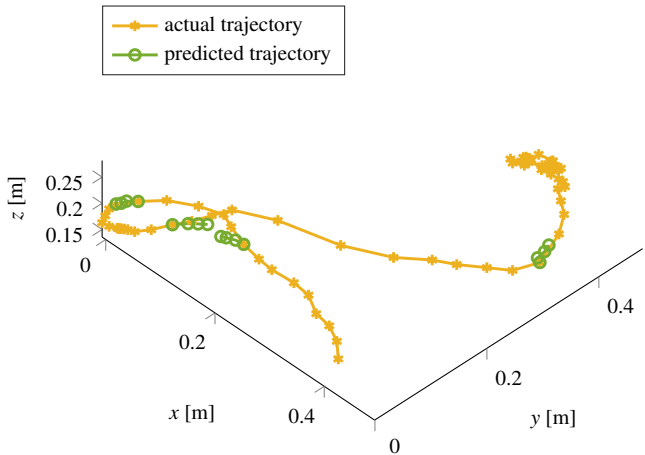

In the test set, we combine the plan recognition and the motion prediction.  The predicted plan of the human worker is an input for the neural network model. \Cref{fig: feature_compare 1} shows how human conducts the first plan.   \Cref{fig: feature_compare 2} shows how human conducts the second plan.  For each moment of the images in \cref{fig: feature_compare 1,fig: feature_compare 2}, the prediction of the motion is shown in \cref{fig: corr_motion 1,fig: corr_motion 2}.

\begin{figure}[t]
{
%
%
%
\begin{tikzpicture}

\begin{axis}[%
width=2.8in,
height=2.25in,
at={(0cm,0cm)},
scale only axis,
plot box ratio=1.864 2.027 1,
xmin=-0.00251062750570163,
xmax=0.322299850452806,
tick align=outside,
xlabel={$x$ [\si{\meter}]},
ymin=-0.323375051416684,
ymax=0.0298694459221789,
ylabel={$y$ [\si{\meter}]},
zmin=0.116199790071178,
zmax=0.290458589792252,
zlabel={$z$ [\si{\meter}]},
view={-166.3}{25.2},
axis background/.style={fill=white},
axis x line*=bottom,
axis y line*=left,
axis z line*=left,
font = \footnotesize,
legend style={at={(0, .8)}, anchor=south west, legend cell align=left, align=left, draw=white!15!black}
]
\addplot3 [color=mycolor3, line width=1.1pt, mark=asterisk, mark options={solid, mycolor3}]
 table[row sep=crcr] {%
0.322299850452806	0.0298694459221789	0.18140452278068\\
0.316648651649731	0.025292174037836	0.188127623608267\\
0.300694808318983	0.0221130550166949	0.2102258807086\\
0.287153765586888	0.0190520417098421	0.236867008458413\\
0.272935510902246	0.00661924191656105	0.254677077351456\\
0.238798033239262	-0.0210466459482117	0.268479179192751\\
0.209964496083573	-0.047287754381237	0.277257768568915\\
0.191073678162618	-0.0656524817928781	0.28157040538792\\
0.160622989304386	-0.0931430286555174	0.286382062345961\\
0.132619983216928	-0.118359652027839	0.288123924285998\\
0.111775502732387	-0.138055819271362	0.285228758478762\\
0.082297151504002	-0.166038943457953	0.272500540857675\\
0.039738247808769	-0.207125588941311	0.247708785467475\\
0.0170272746555526	-0.231714302947771	0.2318537294224\\
0.0169836239332067	-0.236820669619064	0.226296378184735\\
0.0099918068826802	-0.250336439881329	0.213163956102707\\
0.00135007303247586	-0.26927753090114	0.196330865635369\\
-0.00204664110038882	-0.28421812916916	0.183910484917938\\
-0.00251062750570163	-0.294682115702559	0.177552129684712\\
0.00723539113470878	-0.296561993299152	0.184201031220301\\
0.0178571532409139	-0.300613470390307	0.190231497211797\\
0.0192982623914227	-0.309199437285165	0.185250271411471\\
0.0216658118851055	-0.311302549010818	0.184066161260685\\
0.0216587630927632	-0.31583110768555	0.182679416839596\\
0.0185291207041642	-0.321501599173117	0.17978318732996\\
0.017101474909963	-0.323375051416684	0.178937455751202\\
0.01590386478949	-0.323365487450474	0.178073888639119\\
0.0108011115151875	-0.322589431885378	0.171460666805841\\
0.0014633731360445	-0.3173853288498	0.160515763628963\\
0.00687466701816475	-0.30765858248671	0.171250985708217\\
0.0183727654744045	-0.299694027218439	0.186469277182189\\
0.0281499642482079	-0.27986659198342	0.186159722244912\\
0.0255500898256147	-0.252927059594366	0.173737196821261\\
0.0167415601910972	-0.232346852197269	0.164609368793473\\
0.0230149557346366	-0.207763589241138	0.171713853698034\\
0.0448689616921137	-0.170263093918447	0.187437897966457\\
0.0810697544548222	-0.123750622929829	0.205862339967795\\
0.114697202378001	-0.0887432666306294	0.21265267445766\\
0.155846040384198	-0.0532891518725851	0.207357141253084\\
0.194219706090679	-0.0211248786302933	0.200555973569673\\
0.20651149142789	-0.0110208515848749	0.196926831962756\\
0.209660497392976	-0.00655531507214868	0.190411057563987\\
0.214388737410122	-0.00410226280395898	0.181330422194139\\
0.217424558560082	-0.00932333028983934	0.164112043539262\\
0.219782720282855	-0.0156679634081339	0.147055217618873\\
0.222333437844909	-0.0168688169380531	0.140464557397299\\
0.224028020365195	-0.0161516154515493	0.136082150285938\\
0.225374855957233	-0.0150792628521819	0.132734398180188\\
0.227516791525926	-0.0137181946111669	0.131412893546522\\
0.2294361764939	-0.0126909706819132	0.13050185889102\\
0.231114204939698	-0.0130297975752197	0.12658278859485\\
0.232747488236177	-0.0137155389951894	0.122244328114526\\
0.234161663526058	-0.0138926751427586	0.120995494327289\\
0.235211629525466	-0.0143018151157471	0.120136537458921\\
0.235519176471526	-0.0157161367835483	0.117799407894354\\
0.235396440333308	-0.0169935373798397	0.116199790071178\\
0.235493040097934	-0.016997201075908	0.117649691541012\\
0.23611776185487	-0.018374589861245	0.119769959612086\\
0.237282002630587	-0.0203127781672783	0.121150798316919\\
0.237934116918294	-0.0212330388686449	0.121800200394969\\
0.238814026237388	-0.0221329171484602	0.122454985641576\\
0.239335991352082	-0.0233277323180773	0.122813060130026\\
0.239846506478945	-0.0243790354931004	0.123045805983813\\
0.240745603221386	-0.0248867042481828	0.123454876398796\\
0.240882706452217	-0.0251020861074875	0.123706852072703\\
0.240839964256376	-0.0254805565556527	0.123751568249989\\
0.240962524047155	-0.0256515669014581	0.123676807870455\\
0.240806851505536	-0.0259646037296829	0.123477583742725\\
0.238996537424241	-0.0280191997249735	0.122498541443852\\
0.237640446845327	-0.0296055209094082	0.12206704724864\\
0.237727294293122	-0.0296756579354878	0.122575078499333\\
0.237726237107988	-0.0298954864017058	0.122690196540883\\
0.238500803466523	-0.0302962559079072	0.123138312563322\\
0.239391209968472	-0.0310151323482986	0.12438031531007\\
0.240866108060573	-0.0327825484656755	0.126344385177987\\
0.24100847644394	-0.035978360388212	0.126638567969746\\
0.239356933154812	-0.0398153409493194	0.125434972975082\\
0.238954842300505	-0.0440717394463461	0.125999505424862\\
};
 \addlegendentry{actual trajectory}

\addplot3 [color=mycolor5, line width=1.5pt, mark=o, mark options={solid, mycolor5}]
 table[row sep=crcr] {%
0.132619976997375	-0.11835965514183	0.288123935461044\\
0.138774394989014	-0.143515631556511	0.290458589792252\\
0.135267302393913	-0.145870596170425	0.284202665090561\\
0.11555340886116	-0.154574677348137	0.285760283470154\\
};
 \addlegendentry{predicted trajectory}

\addplot3 [color=mycolor5, line width=1.1pt, mark=o, mark options={solid, mycolor5}]
 table[row sep=crcr] {%
-0.00204664119519293	-0.28421813249588	0.183910489082336\\
0.00154546275734901	-0.281344562768936	0.199701651930809\\
0.00687712803483009	-0.280413568019867	0.191875576972961\\
0.00203742459416389	-0.286409437656403	0.195625334978104\\
};
 \addplot3 [color=mycolor5, line width=1.1pt, mark=o, mark options={solid, mycolor5}]
 table[row sep=crcr] {%
0.025550089776516	-0.25292706489563	0.173737198114395\\
0.0296295490115881	-0.249024003744125	0.191475242376328\\
0.0432780385017395	-0.242784693837166	0.189212381839752\\
0.0426618866622448	-0.233259424567223	0.186783701181412\\
};
 \addplot3 [color=mycolor5, line width=1.1pt, mark=o, mark options={solid, mycolor5}]
 table[row sep=crcr] {%
0.081069752573967	-0.123750619590282	0.20586234331131\\
0.100758388638496	-0.093712218105793	0.209479212760925\\
0.132344156503677	-0.0774202644824982	0.203407555818558\\
0.152396023273468	-0.0590489953756332	0.202377736568451\\
};
 \end{axis}
\end{tikzpicture}%
}
\caption{Motion prediction result for plan one corresponding to \cref{fig: feature_compare 1}.}
\label{fig: corr_motion 1}
\end{figure}
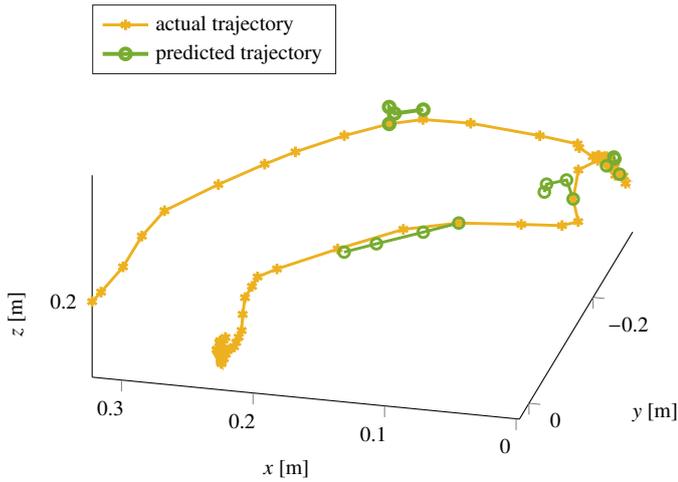
%


\begin{figure*}[t]
\begin{center}
\includegraphics[width = 1\linewidth]{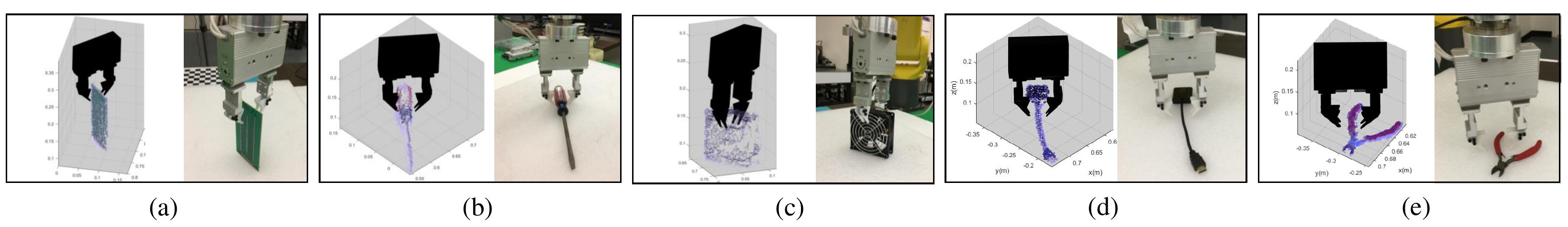}
\caption{T2 Experiment: The grasp examples taught by demonstration.}
\label{fig: T2 train grasp}
\end{center}
\end{figure*}

\begin{figure*}[t]
\begin{center}
\includegraphics[width = 1\linewidth]{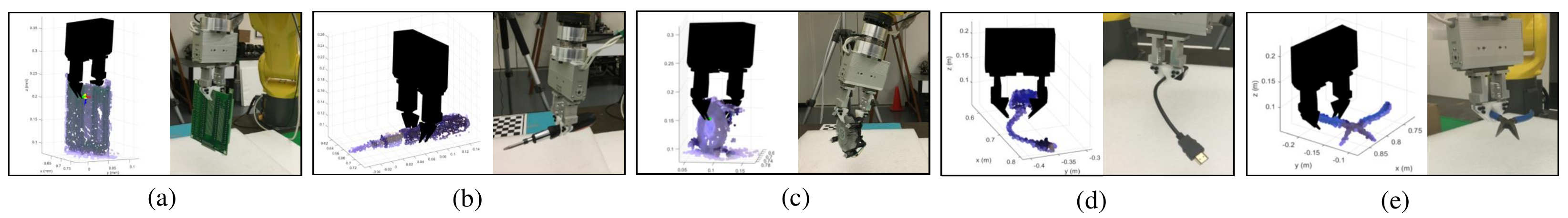}
\caption{T2 Experiment: The grasp pose transferred by CPD.}
\label{fig: T2 test grasp}
\end{center}
\end{figure*}

\subsection{Validation of the Grasp Transferring\label{sec: validation of T2}}
In order to verify the proposed grasping approach in \Cref{sec: T2}, a series of experiments are conducted to grasp various objects that used in the desktop assembly task.

The point clouds are retrieved from the dual Ensenso stereo cameras. By applying the snapshot of the empty workspace as a filter mask, the point clouds of objects are extracted from the background. Then by running the density-based spatial clustering application with noise (DBSCAN) algorithm~\cite{ester1996density}, the point clouds can be separated to several clusters to represent different objects. A voxel grid filter with step size \SI{5}{\milli\meter} is implemented to uniformly downsample the point clouds.

Five categories of objects, including PCB boards, screwdrivers, cooling fans, cable adapters, and pliers, are tested in the experiment (\cref{fig: T2 train grasp}). Note that neither CAD models nor mesh files were used in this work. For each category, a specific source object is selected, and the human operator teaches the preferred grasp poses on it through kinesthetic teaching. The point cloud of the object and the demonstrated grasp poses are recorded as training database.

At the test stage, target objects with different sizes and configurations across all the categories are randomly placed in the workspace. For example, multiple types of PCB boards, screwdrivers and cooling fans are tested for grasping. The pliers are either open or closed. The cable adapter is twisted to various shapes. \Cref{fig: T2 test grasp} shows the grasp transferring results on the target objects. Although the shapes and configurations of target objects are different to the ones of the source object, they share the similar structures. Therefore, the grasp poses on the source object could be transferred to reasonable locations on the target objects. The grasp poses taught by kinesthetic teaching had the intuition from human such as the task specific consideration and fairly good grasping quality, and CPD transferred the insight to the target objects. Therefore, the test can be successful in most of the cases.

\begin{figure*}[htbp]
\begin{center}
\subfloat[Robot in the Idle Mode.\label{fig: sc motion real}]{
\includegraphics[width=6.5cm]{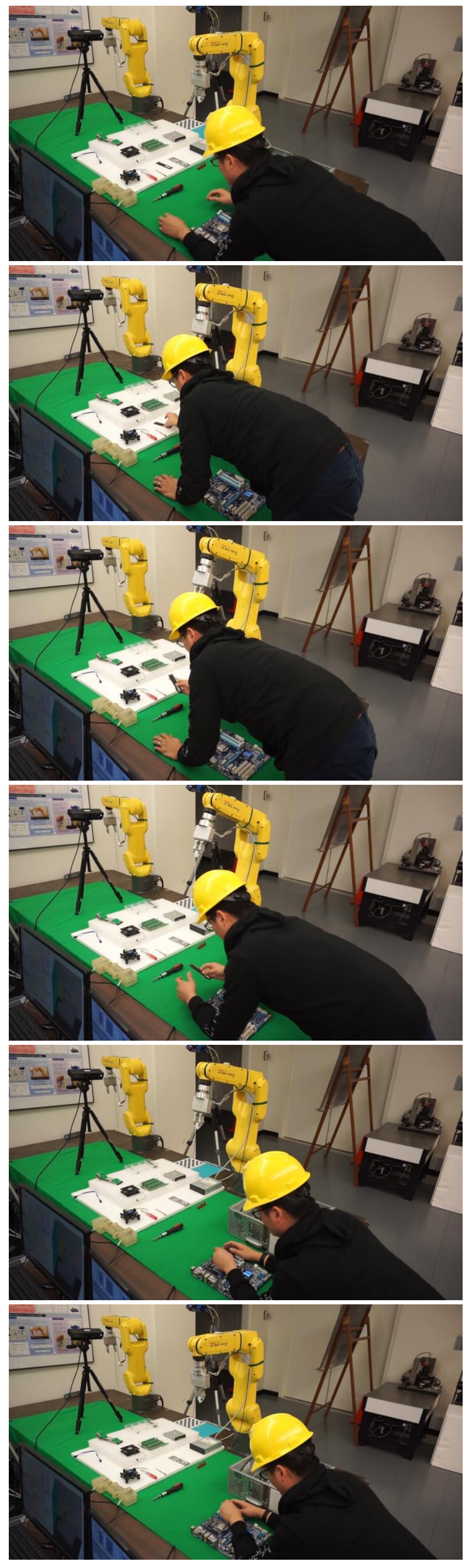}}
\subfloat[Human-Robot Collaboration.\label{fig: rsis motion real}]{
\includegraphics[width=6.5cm]{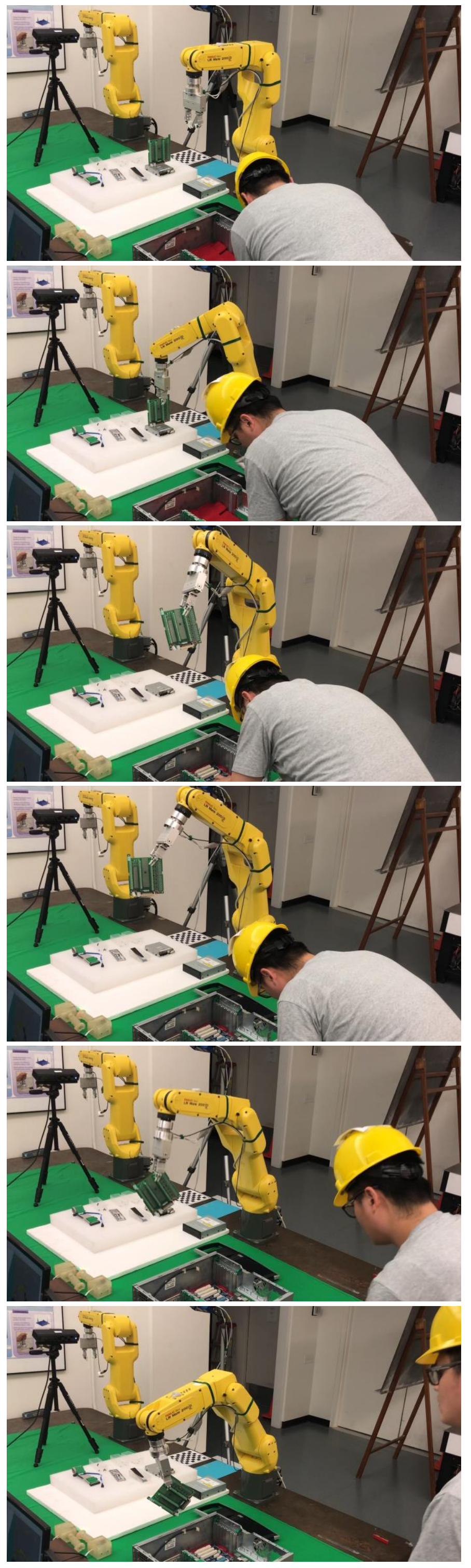}}
\caption{T3 Experiment.}
\end{center}
\end{figure*}

\subsection{Performance of the Safety Controller in the Idle Mode\label{sec: sc experiment}}
\begin{figure*}[t]
\begin{center}
\subfloat[Robot in the Idle Mode.\label{fig: sc motion}]{
\includegraphics[width=7cm]{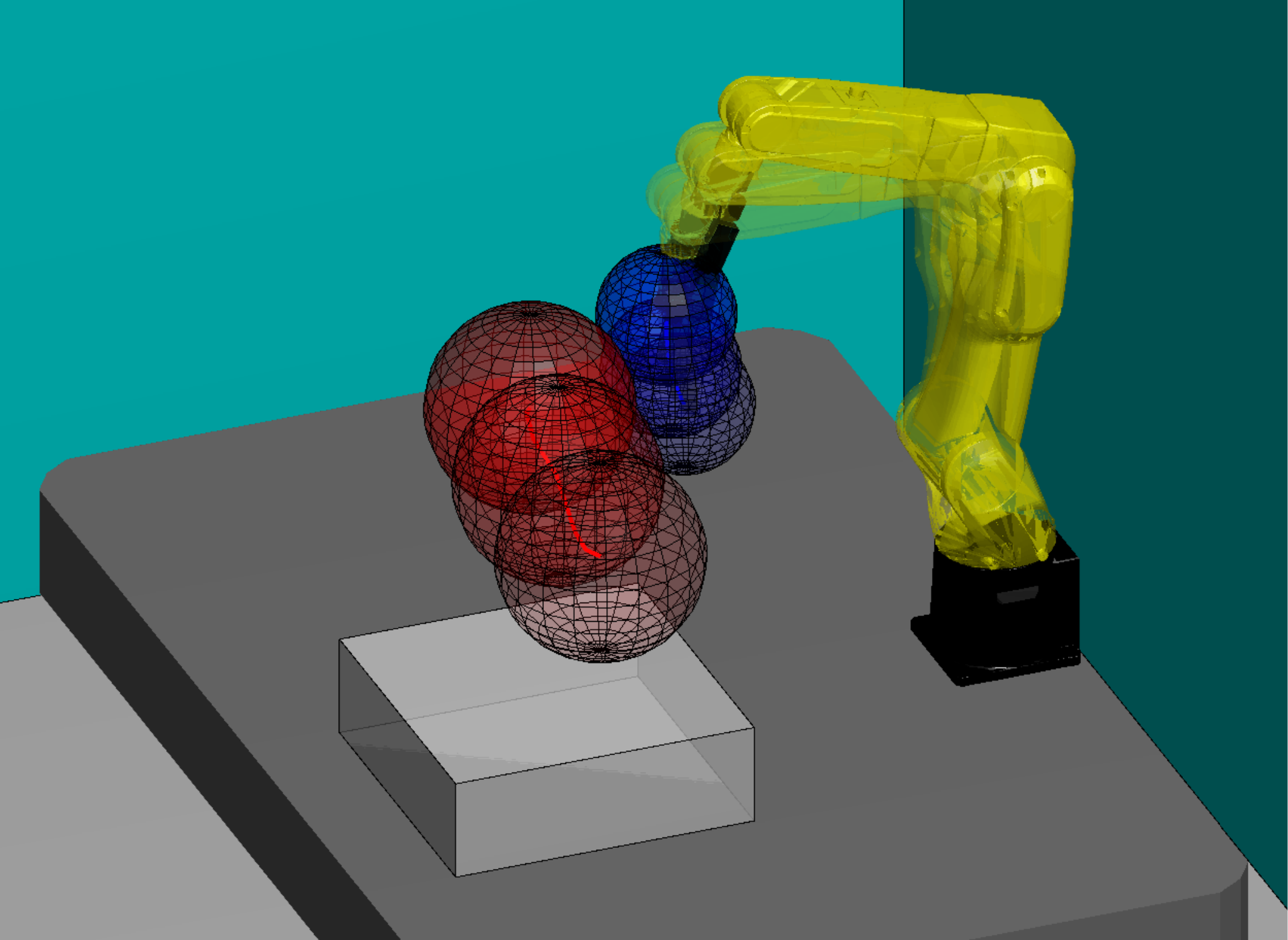}}~~
\subfloat[Human-Robot Collaboration.\label{fig: rsis motion}]{
\includegraphics[width=7cm]{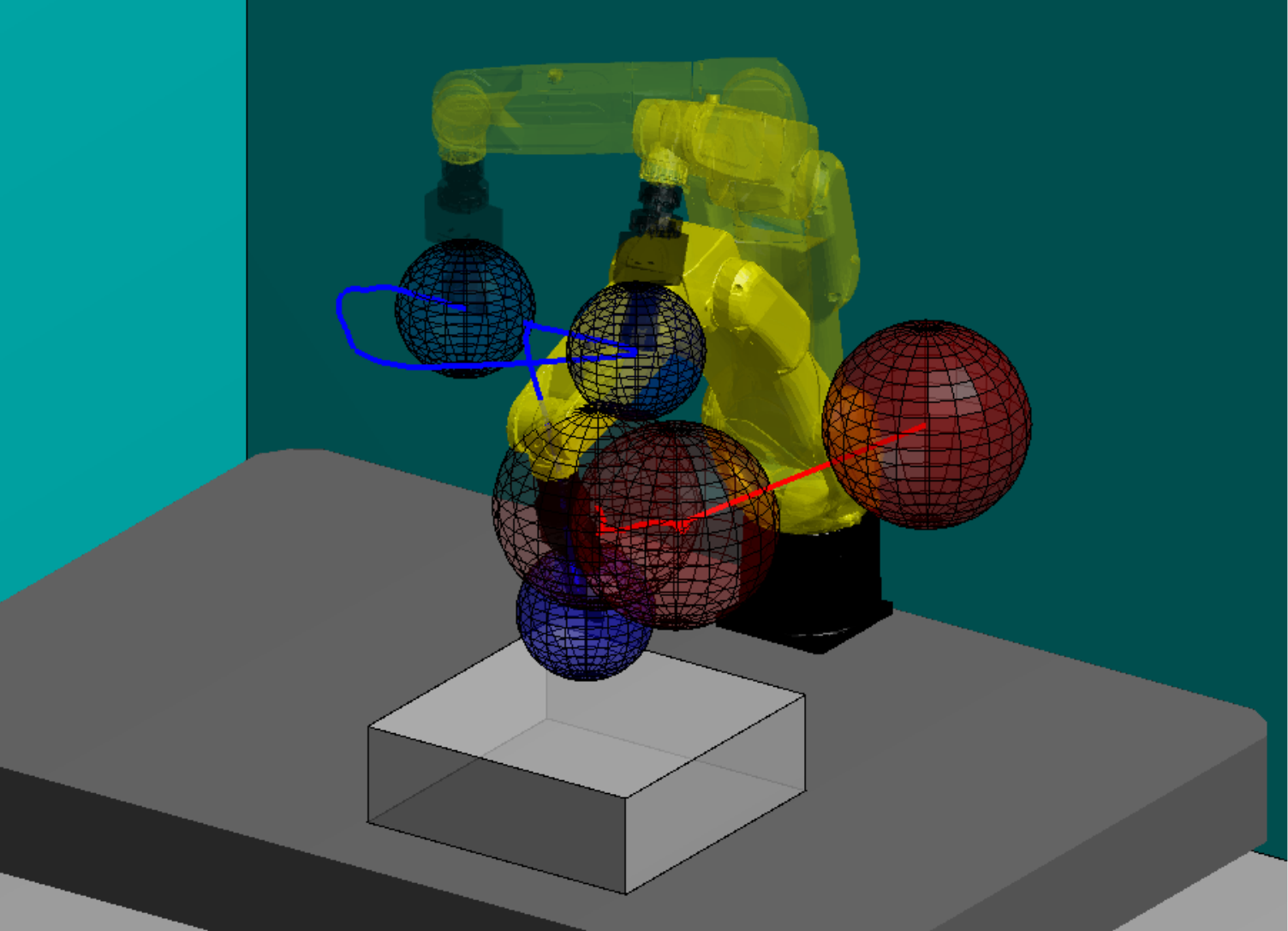}}
\caption{View from the computation model.}
\end{center}
\end{figure*}

\begin{figure}[htbp]
\subfloat[The command from the safety controller.\label{fig: sc only a}]{
\input{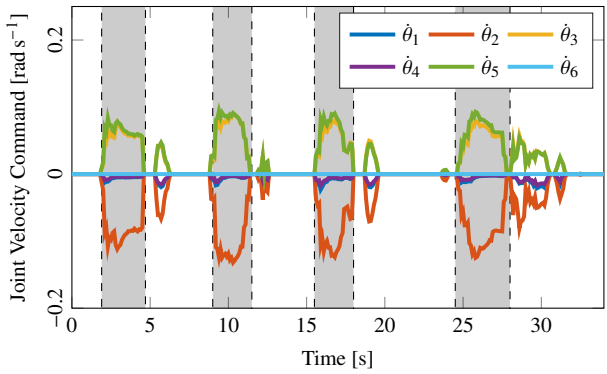}}\\
\subfloat[The joint velocity command applied to the robot.]{
\input{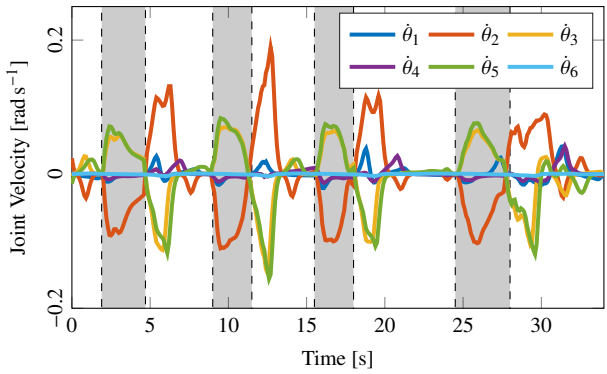}}\\
\subfloat[The Cartesian position of the control point.\label{fig: sc only c}]{
\input{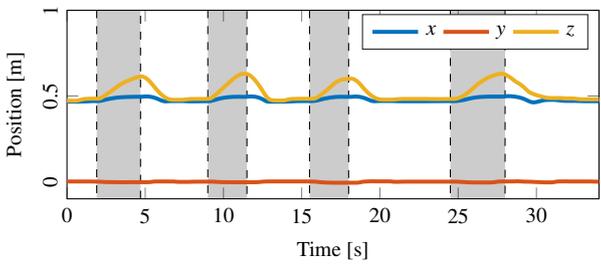}}\\
\subfloat[The minimum distance.\label{fig: sc only d}]{
\input{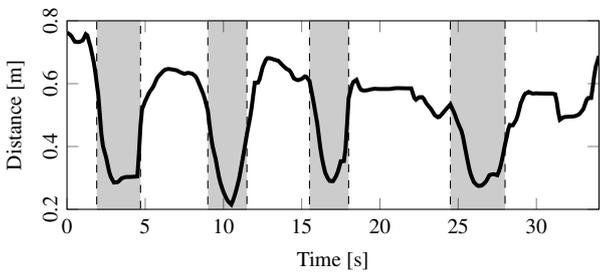}}
\caption{The performance of the safety controller in the idle mode.}
\label{fig: sc only}
\end{figure}

When the robot is in the idle mode, i.e., staying in the neutral position, it can still respond to potential dangers as illustrated in \cref{fig: sc motion real}, \cref{fig: sc motion}, and \cref{fig: sc only}. In this experiment, we are using a simplified environment monitoring module. Markers are placed on the human's helmet such that the robot can track its position. In addition, a safety distance margin of \SI{20}{\centi\meter} is required. Human motion is predicted using a constant speed model, which assumes that the human moves at the same speed in the near future. The Kinect runs at \SI{10}{\hertz}, while the safety controller runs at \SI{1}{\kilo\hertz}. Uncertainties are computed using a predefined maximum acceleration.

\Cref{fig: sc motion real} is a series of pictures taken during the experiment. The robot was in the idle mode, while the human was working on an assembly task. In the second figure, the human reached out to pick a workpiece on the other side of the table. The human did not notice the potential collision with the end effector of the robot arm. As the robot has been actively monitoring the human movement, it moved up to give way to the human. Notice that the upward movement of the robot was most efficient given the prediction of the human movement. After the human got the workpiece, he went back to the sit position. Then the robot went back to its neutral position.

In \cref{fig: sc motion}, the red sphere represents the location of the helmet or the location of the human head. The blue sphere represents the distance margin that we enforce for the critical point. In this case, the critical point is the robot end point. The transparency of the objects (red sphere, blue sphere, and the robot arm) corresponds to different time steps, the lighter the earlier in time. The three configurations correspond to the first three figures in \cref{fig: sc motion real}. The helmet is moving towards the robot arm. To stay safe, the robot arm moves up to avoid collision. 

\Cref{fig: sc only} shows the command from the safety controller, the joint velocity command sent to the robot (which includes the command from both the safety controller and the efficiency controller), the Cartesian position of the control point (in this case, the robot end point), and the minimum distance profile between the human and the robot. The shaded areas in the time axis correspond to the moment that the safety controller is in effect due to collision avoidance. The nonzero commands from the safety controller outside those shaded areas in \cref{fig: sc only a} are due to the velocity regulation, instead of collision avoidance. There are four shaded areas. \Cref{fig: sc motion} corresponds to the second shared area. The collision avoidance strategy adopted by the robot is similar in the four scenarios, that is to move the end effector up, as shown in \cref{fig: sc only c}. The minimum distance between the human and the robot is always kept greater than $\SI{0.2}{\meter}$ as shown in \cref{fig: sc only d}.

\subsection{Performance in Human-Robot Collaborative Assembly\label{sec: validation all}}
\begin{figure}[htbp]
\subfloat[The command from the safety controller.\label{fig: sc ec a}]{
\input{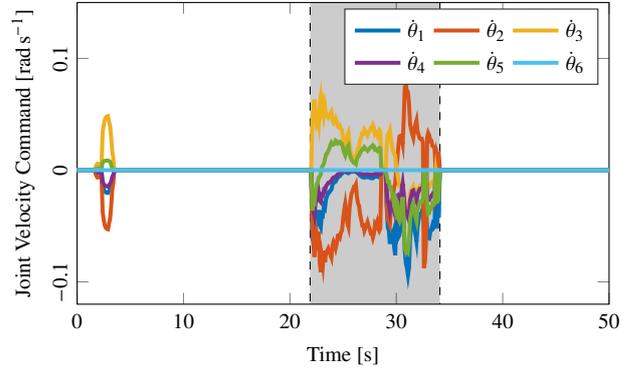}}\\
\subfloat[The joint velocity command applied to the robot.\label{fig: sc ec b}]{
\input{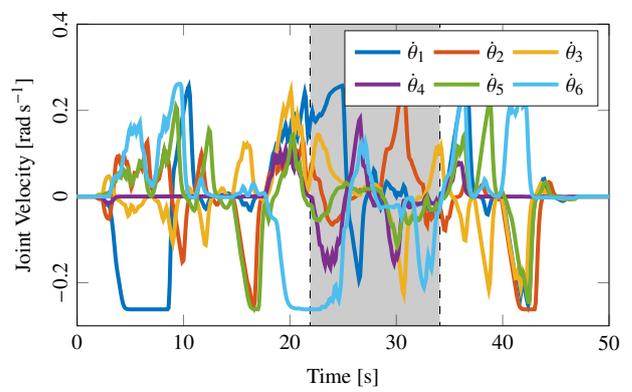}}\\
\subfloat[The Cartesian position of the control point.\label{fig: sc ec c}]{
\input{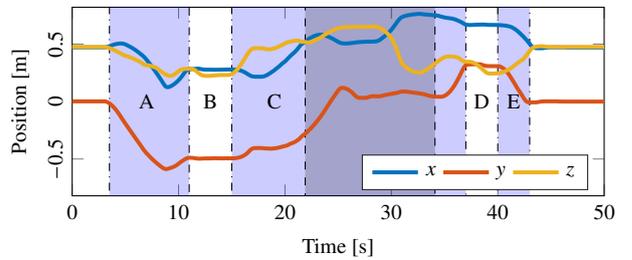}
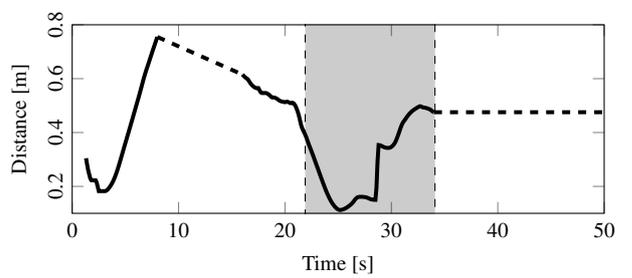}\\
\subfloat[The minimum distance.\label{fig: sc ec d}]{
%
%
\definecolor{mycolor1}{rgb}{0.00000,0.44700,0.74100}%
\begin{tikzpicture}

\begin{axis}[%
width=7cm,
height=2.5cm,
at={(0cm,0cm)},
scale only axis,
unbounded coords=jump,
xmin=0,
xmax=50,
xlabel={Time [\si{\second}]},
yticklabel style={rotate=90},
font=\footnotesize,
ymin=0.1,
ymax=0.8,
ylabel={Distance [\si{\meter}]},
]
\addplot [dashed, fill=black, fill opacity=0.2, forget plot]coordinates {
            (21.9, 1) 
            (21.9, -1)
            (34.1, -1)
            (34.1, 1)};

\addplot [color=black, line width=1.5pt, forget plot]
  table[row sep=crcr]{%
0.001	nan\\
1.201	nan\\
1.301	0.304540335461574\\
1.401	0.2816236453661\\
1.501	0.260210638839487\\
1.601	0.244358669814502\\
1.701	0.229625406105406\\
1.801	0.222715516563591\\
1.901	0.222715609960205\\
2.001	0.222715516563591\\
2.101	0.222715609960205\\
2.201	0.222715609960205\\
2.301	0.21446870209271\\
2.401	0.198233957260355\\
2.501	0.182381141426872\\
2.601	0.182381209382934\\
2.701	0.182381209382934\\
2.801	0.182381209382934\\
2.901	0.182381228151639\\
3.001	0.182381382833036\\
3.101	0.182460467368261\\
3.201	0.183858874505431\\
3.301	0.187034342843338\\
3.401	0.191334642120719\\
3.501	0.196464426737499\\
3.601	0.202119492596088\\
3.701	0.208541467520922\\
3.801	0.215756630697424\\
3.901	0.224120887123453\\
4.001	0.233411031971532\\
4.101	0.243860060647872\\
4.201	0.254938342153179\\
4.301	0.267169602685283\\
4.401	0.27991027923974\\
4.501	0.293186633587205\\
4.601	0.306779038946212\\
4.701	0.320266460415983\\
4.801	0.334037236444086\\
4.901	0.348009131869747\\
5.001	0.361367519570048\\
5.101	0.374760156364167\\
5.201	0.388207607841423\\
5.301	0.401441778341879\\
5.401	0.415250050620909\\
5.501	0.428974130846974\\
5.601	0.442411307492309\\
5.701	0.456021755055256\\
5.801	0.469949332523922\\
5.901	0.483671675639076\\
6.001	0.497543851827817\\
6.101	0.511344495133969\\
6.201	0.524854845277314\\
6.301	0.53900670136735\\
6.401	0.553418914706498\\
6.501	0.567870343458346\\
6.601	0.582755368162179\\
6.701	0.59759123369825\\
6.801	0.611181763044532\\
6.901	0.624116002596196\\
7.001	0.636476740868666\\
7.101	0.648074073420949\\
7.201	0.660009320429203\\
7.301	0.672208990902669\\
7.401	0.68424347606354\\
7.501	0.696375661065118\\
7.601	0.708436448255882\\
7.701	0.719813900656357\\
7.801	0.731375466829056\\
7.901	0.743191394762968\\
8.001	0.754888993370174\\
8.101	nan\\
15.901	nan\\
16.001	0.615270496291214\\
16.101	0.610929723187053\\
16.201	0.606942294398498\\
16.301	0.603535539825443\\
16.401	0.600891636948508\\
16.501	0.598022743556549\\
16.601	0.592015797449471\\
16.701	0.585986310739177\\
16.801	0.580409643171319\\
16.901	0.575675343330136\\
17.001	0.572051912884687\\
17.101	0.568995055072363\\
17.201	0.566325914759535\\
17.301	0.56503822380181\\
17.401	0.564996513601286\\
17.501	0.564993697734838\\
17.601	0.557581873298061\\
17.701	0.550356035888242\\
17.801	0.546827927967287\\
17.901	0.547010136813455\\
18.001	0.546328759156658\\
18.101	0.546958169019035\\
18.201	0.546758663178016\\
18.301	0.545368173999916\\
18.401	0.542938724534167\\
18.501	0.540344507481987\\
18.601	0.536720654267373\\
18.701	0.532241741987825\\
18.801	0.530110111307874\\
18.901	0.530310931296442\\
19.001	0.52983625022969\\
19.101	0.528371867662865\\
19.201	0.526553143489287\\
19.301	0.523774636206441\\
19.401	0.520949837739485\\
19.501	0.51781827461556\\
19.601	0.516340714619059\\
19.701	0.515203422756253\\
19.801	0.514002441720715\\
19.901	0.513391495005326\\
20.001	0.51257357845545\\
20.101	0.513680191405428\\
20.201	0.514486926066016\\
20.301	0.513380223899364\\
20.401	0.511189992851893\\
20.501	0.50912991182497\\
20.601	0.509142072286616\\
20.701	0.510689473081959\\
20.801	0.508948279322302\\
20.901	0.503774682737233\\
21.001	0.497603810180016\\
21.101	0.487655159825625\\
21.201	0.476909843399617\\
21.301	0.463402234466462\\
21.401	0.446794659097909\\
21.501	0.429589236583492\\
21.601	0.417412180297322\\
21.701	0.408439985428231\\
21.801	0.400142042611967\\
21.901	0.391704032417531\\
22.001	0.382304090513633\\
22.101	0.37099930531513\\
22.201	0.360073092637725\\
22.301	0.349438986707483\\
22.401	0.339548433553926\\
22.501	0.329008879720263\\
22.601	0.318698084080081\\
22.701	0.30786153848047\\
22.801	0.297062459290209\\
22.901	0.28600306749706\\
23.001	0.275338964581836\\
23.101	0.26452838576578\\
23.201	0.253658386116365\\
23.301	0.242487417831202\\
23.401	0.23146910297247\\
23.501	0.220862592904133\\
23.601	0.210203312933047\\
23.701	0.199681684251982\\
23.801	0.189560990195391\\
23.901	0.179746249899947\\
24.001	0.170769450351879\\
24.101	0.162341870594197\\
24.201	0.154125060681843\\
24.301	0.14678912410856\\
24.401	0.140091972239752\\
24.501	0.134171402019546\\
24.601	0.129029904594224\\
24.701	0.124098058770494\\
24.801	0.120190673618811\\
24.901	0.116874463095783\\
25.001	0.114140118028277\\
25.101	0.112613501827982\\
25.201	0.112168171163935\\
25.301	0.112878445460154\\
25.401	0.114196311198692\\
25.501	0.115380351834652\\
25.601	0.117107532537682\\
25.701	0.119067294703917\\
25.801	0.121497294155994\\
25.901	0.123896397447676\\
26.001	0.12683062034717\\
26.101	0.129863531236537\\
26.201	0.133212879432509\\
26.301	0.136965178683213\\
26.401	0.141507960882639\\
26.501	0.146699343950964\\
26.601	0.152058106675481\\
26.701	0.15596821849209\\
26.801	0.159277835648711\\
26.901	0.16045055562612\\
27.001	0.161227542326326\\
27.101	0.161336011101412\\
27.201	0.16142616367419\\
27.301	0.161076685674258\\
27.401	0.160698594477031\\
27.501	0.160199156181802\\
27.601	0.159580821256966\\
27.701	0.158402697481209\\
27.801	0.156731324245349\\
27.901	0.155044077246923\\
28.001	0.153114102711631\\
28.101	0.151817906464109\\
28.201	0.151090188331384\\
28.301	0.150569471945783\\
28.401	0.150177218968968\\
28.501	0.150027770908233\\
28.601	0.196084239133404\\
28.701	0.293420940650774\\
28.801	0.35350045425836\\
28.901	0.352931897293082\\
29.001	0.351958609049549\\
29.101	0.349714841312237\\
29.201	0.347704978434585\\
29.301	0.345827645592907\\
29.401	0.34439982644367\\
29.501	0.343463220539235\\
29.601	0.342882054689303\\
29.701	0.342782080405675\\
29.801	0.343197888774788\\
29.901	0.344260643050376\\
30.001	0.34679543313524\\
30.101	0.350139639100005\\
30.201	0.354118059060745\\
30.301	0.359789423181434\\
30.401	0.367331910404997\\
30.501	0.376518670198308\\
30.601	0.386708787899887\\
30.701	0.397920652947304\\
30.801	0.409538776954657\\
30.901	0.420908496738401\\
31.001	0.430472805152981\\
31.101	0.437862827628369\\
31.201	0.443434722890049\\
31.301	0.448595847100798\\
31.401	0.454427729468967\\
31.501	0.460033905342694\\
31.601	0.465105861569476\\
31.701	0.469829935728529\\
31.801	0.473929115823533\\
31.901	0.47768107436362\\
32.001	0.481141035583895\\
32.101	0.483998652306936\\
32.201	0.486675768184725\\
32.301	0.489187582077515\\
32.401	0.491646638052623\\
32.501	0.494113043782364\\
32.601	0.496218086135243\\
32.701	0.497557963317566\\
32.801	0.496543054113305\\
32.901	0.494527846087998\\
33.001	0.49276294499214\\
33.101	0.492551358279238\\
33.201	0.491655604476344\\
33.301	0.49071989474927\\
33.401	0.48930066305901\\
33.501	0.487815459770269\\
33.601	0.486206266724384\\
33.701	0.484213648002456\\
33.801	0.480317480818916\\
33.901	0.477387159189172\\
34.001	0.475412242167835\\
34.101	nan\\
50.001	nan\\
};

\addplot [color=black, dashed, line width=1.5pt, forget plot]
  table[row sep=crcr]{%
8.001	0.754888993370174\\
16.001	0.615270496291214\\
  };

\addplot [color=black, dashed, line width=1.5pt, forget plot]
  table[row sep=crcr]{%
34.001	0.475412242167835\\
50		0.475412242167835\\
  };

\end{axis}
\end{tikzpicture}%
}
\caption{The performance of SERoCS during collaboration.}
\label{fig: sc ec}
\end{figure}

The performance of SERoCS is also evaluated in a human-robot collaborative assembly task. In this task, the human was working with the cable assembly for inside the desktop. Then the robot inferred that he needed to insert the motion board. As the motion board was out of the reach from the human, the robot then picked the motherboard and handed it to the human. The performance of the robot is illustrated in \cref{fig: rsis motion real}, \cref{fig: rsis motion}, and \cref{fig: sc ec}. In this experiment, we are still using the simplified environment monitoring module discussed earlier. While the safety controller runs at the same sampling rate as in the previous experiment, the efficiency controller runs at \SI{5}{\hertz}.

\Cref{fig: rsis motion real} is a series of pictures taken during the experiment. At the beginning, the human was doing assembly inside the desktop, while the robot decided to reach to the motherboard. The robot made the decision through human plan inference in T1. It grasped the motherboard using the skill learned in T2. While the robot was approaching the motherboard, the distance between the human and the robot was above threshold. Hence the safety controller was silent. The joint velocity command was generated by the efficiency controller, which performed online motion planning from the current position to the grasp position specified in T2. After grasping the motherboard, the robot then carried the motion board to its slot for assembly. However, as the human was too close, the robot could not directly deliver the motherboard. The efficiency controller generated a detour in order to place the workpiece from the right hand side of the human worker, which was shown in \cref{fig: rsis motion}. However, it was still not safe as the human worker was moving around. Thus, the safety controller pushed the robot arm away from the human worker. After the human finished his task inside the desktop and stayed away from the desktop, the robot inserted the motherboard using the skill learned in T2.

\Cref{fig: rsis motion} illustrates the configurations in the computation model. The context behind the geometric objects is the same as explained in \cref{sec: sc experiment}. The three configurations correspond to the third, the fourth, and the last figures in \cref{fig: rsis motion real}.

\Cref{fig: sc ec} shows the command from the safety controller, the joint velocity command sent to the robot (which includes the command from both the safety controller and the efficiency controller), the Cartesian position of the control point (in this case, the robot end point), and the minimum distance profile between the human and the robot. The shaded areas in the time axis correspond to the moment that the safety controller is in effect due to collision avoidance. The safety controller for collision avoidance was triggered only once. As the robot was finishing certain tasks, the joint velocity contained much richer spectrums as shown in \cref{fig: sc ec b}. The task phase of the robot can be interpreted from the location of the end effector as shown in \cref{fig: sc ec c}. The end effector both started and ended at the neutral position. Robot approached the motherboard in phase A, grasped the motherboard in phase B, moved the motherboard to the desired location (above the slot) in phase C, placed the motion board in phase D, and returned to the neutral position in phase E. Due to occlusion, there were moments that the robot lost track of the human as shown in \cref{fig: sc ec d}. In the short term (less than $\SI{1}{\second}$), the uncertainty induced by occlusion can be compensated in Task~1. However, it may put the human subject in great danger when the robot loses track of the human in a long time horizon. Avoidance of occlusion and compensation of the uncertainty induced by occlusion in the long term will be studied in the future.

\section{Conclusion\label{sec: conclusion}}
This paper discussed a set of design principles of the safe and efficient robot collaboration system (SERoCS) for the next generation co-robots, which consisted of robust cognition algorithms for environment monitoring, optimal task planning algorithms for safe human-robot collaborations, and safe motion planning and control algorithms for safe human-robot interactions.  As demonstrated by the experiment, the proposed SERoCS addressed the design challenges and significantly expanded the skill sets of the co-robots to allow them to work safely and efficiently with their human counterparts. The development of SERoCS will create a significant advancement toward adoption of co-robots in various industries. In the future, we will apply SERoCS to diverse industrial tasks in addition to the laptop assembly task demonstrated in this paper.

\section*{Acknowledgement}
The authors would like to thank Jessica Leu for her help in the experiment.



\bibliographystyle{elsarticle-num} 
\bibliography{nsf-nri}

\begin{thebibliography}{10}
\expandafter\ifx\csname url\endcsname\relax
  \def\url#1{\texttt{#1}}\fi
\expandafter\ifx\csname urlprefix\endcsname\relax\def\urlprefix{URL }\fi
\expandafter\ifx\csname href\endcsname\relax
  \def\href#1#2{#2} \def\path#1{#1}\fi

\bibitem{charalambous2013human}
G.~Charalambous, Human-automation collaboration in manufacturing: Identifying
  key implementation factors, in: Proceedings of the 2013 International
  Conference on Ergonomics \& Human Factors, CRC Press, 2013, p.~59.

\bibitem{koeppe2005robot}
R.~Koeppe, D.~Engelhardt, A.~Hagenauer, P.~Heiligensetzer, B.~Kneifel,
  A.~Knipfer, K.~Stoddard, Robot-robot and human-robot cooperation in
  commercial robotics applications, in: P.~Dario, R.~Chatila (Eds.), Robotics
  Research, Vol.~15 of Springer Tracts in Advanced Robotics, Springer Berlin
  Heidelberg, 2005, pp. 202--216.

\bibitem{kruger2009cooperation}
J.~Kr{\"u}ger, T.~Lien, A.~Verl, Cooperation of human and machines in assembly
  lines, CIRP Annals-Manufacturing Technology 58~(2) (2009) 628--646.

\bibitem{schmidt2008kobot}
U.~Schmidt, R.~Konzack, {KOBOT} - cooperative robot systems for a versatile
  production, in: Proceedings of the 2008 IEEE International Conference on
  Distributed Human-Machine Systems, IEEE, 2008, pp. 503--507.

\bibitem{pine1999mass}
B.~J. Pine, Mass customization: the new frontier in business competition,
  Harvard Business Press, 1999.

\bibitem{hutchinson1982economic}
G.~K. Hutchinson, J.~R. Holland, The economic value of flexible automation,
  Journal of Manufacturing Systems 1~(2) (1982) 215--228.

\bibitem{jovane2003present}
F.~Jovane, Y.~Koren, C.~Bo{\"e}r, Present and future of flexible automation:
  towards new paradigms, CIRP Annals-Manufacturing Technology 52~(2) (2003)
  543--560.

\bibitem{Volkswagen-Cooperative}
J.~Leber, At {V}olkswagen, robots are coming out of their cages (Sep 2013).

\bibitem{Econ-Our-Friends-Electric}
Working with robots: Our friends electric, The Economist.

\bibitem{rey2013cooperation}
G.~Z. Rey, M.~Carvalho, D.~Trentesaux, Cooperation models between humans and
  artificial self-organizing systems: Motivations, issues and perspectives, in:
  Proceedings of the 2013 6th International Symposium on Resilient Control
  Systems (ISRCS), 2013, pp. 156--161.

\bibitem{ulusoy1997genetic}
G.~Ulusoy, F.~Sivrikaya-{\c{S}}erifo{\v g}lu, {\"U}.~Bilge, A genetic algorithm
  approach to the simultaneous scheduling of machines and automated guided
  vehicles, Computers and Operations Research 24~(4) (1997) 335--351.

\bibitem{UR5}
\href{www.universal-robots.com}{{UR}5 from {Universal Robotics}}.
\newline\urlprefix\url{www.universal-robots.com}

\bibitem{fanucCorobot}
\href{robot.fanucamerica.com}{Collaborative industrial robots {CR-35iA} from
  {FANUC} {Corporation} {Japan}}.
\newline\urlprefix\url{robot.fanucamerica.com}

\bibitem{fanucGreenRobot}
M.~Morioka, T.~Iwayama, Y.~Inoue, T.~Yamamoto, Y.~Naito, T.~Sato, S.~Toda,
  S.~Takahashi, The human-collaborative industrial robot -- {D}evelopment of
  `{Green Robot}', FANUC Techinical Review 24~(2) (2016) 20 -- 29.

\bibitem{rethinkrobotics}
\href{www.rethinkrobotics.com}{Baxter from {Rethink Robotics}}.
\newline\urlprefix\url{www.rethinkrobotics.com}

\bibitem{kawada}
A.Saenz, \href{singularityhub.com}{A drop-in solution for replacing human
  labor? {Kawada's NextAge} robot}.
\newline\urlprefix\url{singularityhub.com}

\bibitem{workerbot}
S.~Bouchard, With two arms and a smile, {Pi4 Workerbot} is one happy factory
  bot, IEEE Spectrum.

\bibitem{yamazaki2012home}
K.~Yamazaki, R.~Ueda, S.~Nozawa, M.~Kojima, K.~Okada, K.~Matsumoto,
  M.~Ishikawa, I.~Shimoyama, M.~Inaba, Home-assistant robot for an aging
  society, Proceedings of the IEEE 100~(8) (2012) 2429--2441.

\bibitem{burgard1999experiences}
W.~Burgard, A.~B. Cremers, D.~Fox, D.~H{\"a}hnel, G.~Lakemeyer, D.~Schulz,
  W.~Steiner, S.~Thrun, Experiences with an interactive museum tour-guide
  robot, Artificial Intelligence 114~(1) (1999) 3--55.

\bibitem{thrun1999minerva}
S.~Thrun, M.~Bennewitz, W.~Burgard, A.~B. Cremers, F.~Dellaert, D.~Fox,
  D.~Hahnel, C.~Rosenberg, N.~Roy, J.~Schulte, D.~Schulz, Minerva: A
  second-generation museum tour-guide robot, in: Proceedings of the 1999 IEEE
  International Conference on Robotics and Automation (ICRA), Vol.~3, IEEE,
  1999.

\bibitem{pineau2003towards}
J.~Pineau, M.~Montemerlo, M.~Pollack, N.~Roy, S.~Thrun, Towards robotic
  assistants in nursing homes: Challenges and results, Robotics and Autonomous
  Systems 42~(3) (2003) 271--281.

\bibitem{harper2010towards}
C.~Harper, G.~Virk, Towards the development of international safety standards
  for human robot interaction, International Journal of Social Robotics 2~(3)
  (2010) 229--234.

\bibitem{PHRIENDS}
\href{www.phriends.eu}{{PHRIENDS}: Physical human-robot interaction:
  dependability and safety}.
\newline\urlprefix\url{www.phriends.eu}

\bibitem{ROSETTA}
\href{www.fp7rosetta.org}{{ROSETTA}: Robot control for skilled execution of
  tasks in natural interaction with humans based on autonomy, cumulative
  knowledge and learning}.
\newline\urlprefix\url{www.fp7rosetta.org}

\bibitem{SAPHARI}
\href{www.saphari.eu}{{SAPHARI}: Safe and autonomous physical human-aware robot
  interaction}.
\newline\urlprefix\url{www.saphari.eu}

\bibitem{anandan2014major}
T.~M. Anandan, \href{www.robotics.org}{Major robot {OEM}s fast-tracking
  cobots}.
\newline\urlprefix\url{www.robotics.org}

\bibitem{Tadele2014safety}
T.~S. Tadele, T.~J.~d. Vries, S.~Stramigioli, The safety of domestic robots: a
  survey of various safety-related publications, IEEE Robotics and Automation
  Magazine (2014) 134--142.

\bibitem{hirzinger2001new}
G.~Hirzinger, A.~Albu-Schaffer, M.~Hahnle, I.~Schaefer, N.~Sporer, On a new
  generation of torque controlled light-weight robots, in: Proceedings of the
  2001 IEEE International Conference on Robotics and Automation (ICRA), Vol.~4,
  IEEE, 2001, pp. 3356--3363.

\bibitem{zinn2004new}
M.~Zinn, B.~Roth, O.~Khatib, J.~K. Salisbury, A new actuation approach for
  human friendly robot design, The International Journal of Robotics Research
  23~(4-5) (2004) 379--398.

\bibitem{jafari2010novel}
A.~Jafari, N.~G. Tsagarakis, B.~Vanderborght, D.~G. Caldwell, A novel actuator
  with adjustable stiffness (awas), in: Proceedings of the 2010 IEEE/RSJ
  International Conference on Intelligent Robots and Systems (IROS), IEEE,
  2010, pp. 4201--4206.

\bibitem{english1999implementation}
C.~English, D.~Russell, Implementation of variable joint stiffness through
  antagonistic actuation using rolamite springs, Mechanism and Machine Theory
  34~(1) (1999) 27--40.

\bibitem{albu2007unified}
A.~Albu-Sch{\"a}ffer, C.~Ott, G.~Hirzinger, A unified passivity-based control
  framework for position, torque and impedance control of flexible joint
  robots, The International Journal of Robotics Research 26~(1) (2007) 23--39.

\bibitem{luo2011adaptive}
R.~C. Luo, H.~B. Huang, C.~Yi, Y.~W. Perng, Adaptive impedance control for safe
  robot manipulator, in: Proceedings of the 2011 9th World Congress on
  Intelligent Control and Automation (WCICA), IEEE, 2011, pp. 1146--1151.

\bibitem{hogan1984impedance}
N.~Hogan, Impedance control: An approach to manipulation, in: Proceedings of
  the 1984 American Control Conference (ACC), IEEE, 1984, pp. 304--313.

\bibitem{tan2009human}
J.~T.~C. Tan, F.~Duan, Y.~Zhang, K.~Watanabe, R.~Kato, T.~Arai, Human-robot
  collaboration in cellular manufacturing: Design and development, in:
  Proceedings of the 2009 IEEE/RSJ International Conference on Intelligent
  Robots and Systems (IROS), 2009, pp. 29--34.

\bibitem{krizhevsky2012imagenet}
A.~Krizhevsky, I.~Sutskever, G.~E. Hinton, Imagenet classification with deep
  convolutional neural networks, in: Advances in Neural Information Processing
  Systems, 2012, pp. 1097--1105.

\bibitem{ravichandar2017human}
H.~C. Ravichandar, A.~P. Dani, Human intention inference using
  expectation-maximization algorithm with online model learning, IEEE
  Transactions on Automation Science and Engineering 14~(2) (2017) 855--868.

\bibitem{liu2015safe}
C.~Liu, M.~Tomizuka, Safe exploration: Addressing various uncertainty levels in
  human robot interactions, in: Proceedings of the 2015 American Control
  Conference (ACC), 2015, pp. 465 -- 470.

\bibitem{myronenko2010point}
A.~Myronenko, X.~Song, Point set registration: Coherent point drift, IEEE
  Transactions on Pattern Analysis and Machine Intelligence 32~(12) (2010)
  2262--2275.

\bibitem{girosi1995regularization}
F.~Girosi, M.~Jones, T.~Poggio, Regularization theory and neural networks
  architectures, Neural computation 7~(2) (1995) 219--269.

\bibitem{dempster1977maximum}
A.~P. Dempster, N.~M. Laird, D.~B. Rubin, Maximum likelihood from incomplete
  data via the em algorithm, Journal of the royal statistical society. Series B
  (methodological) (1977) 1--38.

\bibitem{myronenko2007non}
A.~Myronenko, X.~Song, M.~A. Carreira-Perpin{\'a}n, Non-rigid point set
  registration: Coherent point drift, in: Advances in Neural Information
  Processing Systems, 2007, pp. 1009--1016.

\bibitem{abraham1978foundations}
R.~Abraham, J.~E. Marsden, J.~E. Marsden, Foundations of mechanics, Vol.~36,
  Benjamin/Cummings Publishing Company Reading, Massachusetts, 1978.

\bibitem{liu2017real}
C.~Liu, M.~Tomizuka, Real time trajectory optimization for nonlinear robotic
  systems: Relaxation and convexification, System \& Control Letters 108 (2017)
  56 -- 63.

\bibitem{savkin2015safe}
A.~V. Savkin, A.~S. Matveev, M.~Hoy, C.~Wang, Safe Robot Navigation Among
  Moving and Steady Obstacles, Butterworth-Heinemann, 2015.

\bibitem{boggs1995sequential}
P.~T. Boggs, J.~W. Tolle, Sequential quadratic programming, Acta numerica 4
  (1995) 1--51.

\bibitem{liu2017sicon}
C.~Liu, C.-Y. Lin, M.~Tomizuka, The convex feasible set algorithm for real time
  optimization in motion planning, SIAM Journal on Control and Optimization.

\bibitem{liu2017convex}
C.~Liu, C.-Y. Lin, Y.~Wang, M.~Tomizuka, Convex feasible set algorithm for
  constrained trajectory smoothing, in: Proceedings of the American Control
  Conference (ACC), 2017, pp. 4177--4182.

\bibitem{liu2014control}
C.~Liu, M.~Tomizuka, Control in a safe set: Addressing safety in human robot
  interactions, in: Proceedings of the ASME 2014 Dynamic Systems and Control
  Conference (DSCC), ASME, 2014, p. V003T42A003.

\bibitem{ester1996density}
M.~Ester, H.-P. Kriegel, J.~Sander, X.~Xu, A density-based algorithm for
  discovering clusters in large spatial databases with noise., in: Kdd,
  Vol.~96, 1996, pp. 226--231.

\end{thebibliography}


%
%
%
\end{document}